\documentclass{article}
\usepackage[utf8]{inputenc}
\usepackage[numbers]{natbib}
\usepackage{manfnt} 

\usepackage{amsthm}

\newtheorem{theorem}{Theorem}
\newtheorem{lemma}{Lemma}
\newtheorem{proposition}{Proposition}

\newtheorem{definition}{Definition}
\usepackage[top=0.8in, bottom=1in, left=1.25in, right=1.25in]{geometry}
\usepackage{scrextend}
\usepackage[utf8]{inputenc} 
\usepackage[T1]{fontenc}    
\usepackage[english]{babel}
\usepackage{lmodern}
\usepackage[bitstream-charter]{mathdesign}
\usepackage[table]{xcolor}
 \usepackage{hyperref}
 \definecolor{burgundy}{rgb}{0.5, 0.0, 0.13}
\definecolor{camel}{rgb}{0.76, 0.6, 0.42}
\definecolor{chamoisee}{rgb}{0.63, 0.47, 0.35}
\definecolor{grey1}{RGB}{128,128,128}
 \hypersetup{colorlinks=true,linkcolor=burgundy,citecolor=chamoisee,urlcolor=burgundy,linktoc=page}
\usepackage{url}            
\usepackage{booktabs}       
\usepackage{nicefrac}       
\usepackage{microtype}      
\usepackage{dsfont}         
\renewcommand{\epsilon}{\varepsilon}
\renewcommand{\phi}{\varphi}

\title{\vspace*{0cm}\bf Markov random geometric graph,\\ MRGG: A growth model for temporal dynamic networks\thanks{This work was supported by grants from Région Ile-de-France.}}

\title{Markov random geometric graph,\\ MRGG: A growth model for temporal dynamic networks}

\author{%
  Quentin Duchemin\thanks{This work was supported by grants from Région Ile-de-France.} \\
 LAMA, Univ Gustave Eiffel, CNRS, Marne-la-Vallée, France.\\
  \texttt{quentin.duchemin@univ-eiffel.fr} \\
  $\And$\\
  Yohann De Castro\\
  Institut Camille Jordan, École Centrale de Lyon, Lyon, France \\
  \texttt{yohann.de-castro@ec-lyon.fr} \\
}

\date{}

\usepackage{amsmath}
\usepackage{mathtools}

\DeclarePairedDelimiter\floor{\lfloor}{\rfloor}
\usepackage{diagbox}
\usepackage{subcaption}

\usepackage{algorithm,algorithmic,refcount}
\usepackage{wrapfig}
\usepackage{tikz}
\usepackage{makecell}
\newcommand{\alglinelabel}{%
  \addtocounter{ALC@line}{-1}
  \refstepcounter{ALC@line}
  \label
}
\usepackage{tikz}
\usetikzlibrary{%
  arrows,%
  calc,%
  shapes.geometric,%
  shapes.misc,%
  shapes.symbols,%
  shapes.arrows,%
  automata,%
  through,%
  positioning,%
  scopes,%
  decorations.shapes,%
  decorations.text,%
  decorations.pathmorphing,%
  shadows}
\usetikzlibrary{positioning}
\usepackage{multirow}
\usepackage{makecell}
\usepackage{mdframed}

\usepackage{savesym}
\savesymbol{checkmark}
\usepackage{dingbat}
\usepackage{wasysym}
\makeatletter
\g@addto@macro\appendix{%
  \addtocontents{toc}{\protect\setcounter{tocdepth}{1}}%
}
\makeatother

\usepackage{enumitem}

\begin{document}

\maketitle

\begin{abstract}
 We introduce Markov Random Geometric Graphs (MRGGs), a growth model for temporal dynamic networks. It is based on a Markovian latent space dynamic: consecutive latent points are sampled on the Euclidean Sphere using an unknown Markov kernel; and two nodes are connected with a probability depending on a unknown function of their latent geodesic distance. 

More precisely, at each stamp-time $k$ we add a latent point $X_k$ sampled by jumping from the previous one $X_{k-1}$ in a direction chosen uniformly $Y_k$ and with a length~$r_k$ drawn from an unknown distribution called the \textit{latitude function}. The connection probabilities between each pair of nodes are equal to the \textit{envelope function} of the distance between these two latent points. We provide theoretical guarantees for the non-parametric estimation of the latitude and the envelope functions.

We propose an efficient algorithm that achieves those non-parametric estimation tasks based on an ad-hoc Hierarchical Agglomerative Clustering approach. As a by product, we show how MRGGs can be used to detect dependence structure in growing graphs and to solve link prediction problems.
\end{abstract}

\section{Introduction}
\label{sec:intro}

In Random Geometric Graphs (RGG), nodes are sampled independently in latent space $\mathds{R}^d$. Two nodes are connected if their distance is smaller than a threshold. A thorough probabilistic study of RGGs can be found in~\cite{P03}. RGGs have been widely studied recently due to their ability to provide a powerful modeling tool for networks with spatial structure. We can mention applications in bioinformatics~\cite{HRP08} or analysis of social media~\cite{HRH02}. One main feature is to uncover hidden representation of nodes using latent space and to model interactions by relative positions between latent points.

Furthermore, nodes interactions may evolve with time. In some applications, this evolution is given by the arrival of new nodes as in online collection growth~\cite{online_collection}, online social network growth~\cite{backstrom2006group,jin2001structure}, or outbreak modeling~\cite{ugander2012structural} for instance. The network is growing as more nodes are entering. Other time evolution modelings have been studied, we refer to~\cite{rossetti2018community} for a review. 

A natural extension of RGG consists in accounting this time evolution. In~\cite{DMP08}, the expected length of connectivity and dis-connectivity periods of the Dynamic Random Geometric Graph is studied: each node choose at random an angle in $[0,2\pi)$ and make a constant step size move in that direction. In~\cite{SS09}, a random walk model for RGG on the hypercube is studied where at each time step a vertex is either appended or deleted from the graph. Their model falls into the class of Geometric Markovian Random Graphs that are generally defined in~\cite{CPMS09}.

As far as we know, there is no extension of RGG to growth model for temporal dynamic networks. For the first time, we consider a Markovian dynamic on the latent space where the new latent point is drawn with respect to the latest latent point and some Markov kernel to be estimated.

\paragraph{Estimation of graphon in RGGs: the Euclidean sphere case}
Random graphs with latent space can be defined using a {\it graphon}, cf.~\cite{L12}. A graphon is a kernel function that defines edge distribution. In~\cite{TS13}, Tang and al. prove that spectral method can recover the matrix formed by graphon evaluated at latent points up to an orthogonal transformation, assuming that graphon is a positive definite kernel~(PSD). Going further, algorithms have been designed to estimate graphons, as in~\cite{KTV17} which provide sharp rates for the Stochastic Block Model (SBM). Recently, the paper~\cite{CL18} provides a non-parametric algorithm to estimate RGGs on Euclidean spheres, without PSD assumption.

We present here RGG on Euclidean sphere. Given $n$ points $X_1, X_2, \dots, X_n$ on the Euclidean sphere~$\mathds{S}^{d-1}$, we set an edge between nodes $i$ and $j$ (where $i,j \in [n]$, $i \neq j$) with independent probability~$\mathbf{p}(\langle X_i,X_j \rangle)$. The unknown function $\mathbf{p}:[-1,1] \to [0,1]$ is called the \textit{envelope function}. This RGG is a graphon model with a symmetric kernel~$W$ given by $W(x,y)=\mathbf{p}(\langle x,y \rangle)$. Once the latent points are given, independently draw the random undirected adjacency matrix $A$ by 
\[
A_{i,j}\sim  \mathrm{Ber}(\mathbf{p}(\langle X_i,X_j \rangle))\,,\quad i<j
\]
with Bernoulli r.v.\! drawn independently (set zero on the diagonal and complete by symmetry), and set 
\begin{equation}
\label{e:enveloppe}
   T_n:=\frac{1}{n}\left(\mathbf{p}(\langle X_i,X_j \rangle)\right)_{i,j \in [n]}\quad \text{ and }\quad \widehat{T}_n:= \frac{1}{n}A, 
\end{equation}
We do not observe the latent points and we have to estimate the envelope $\mathbf{p}$ from $A$ only. A standard strategy is to remark that~$\widehat T_n$ is a random perturbation of $T_n$ and to dig into $T_n$ to uncover $\mathbf{p}$.

One important feature of this model is that the interactions between nodes is depicted by a simple object: the envelope function $\mathbf{p}$. The envelope summarises how individuals connect each others given their latent positions. Standard examples~\cite{B16} are given by $\mathbf{p}_\tau(t)=\mathds 1_{\{t\geq \tau\}}$ where one connects two points as soon as their geodesic distance is below some threshold. The non-parametric estimation of~$\mathbf{p}$ is given by~\cite{CL18} where the authors assume that latent points $X_i$ are independently and uniformly distributed on the sphere, which will not be the case in this work.

\paragraph{A new growth model: the latent Markovian dynamic}
\label{sec:dynamic}

\begin{figure}
\vskip 0.2in
\begin{center}
\includegraphics[width=\linewidth]{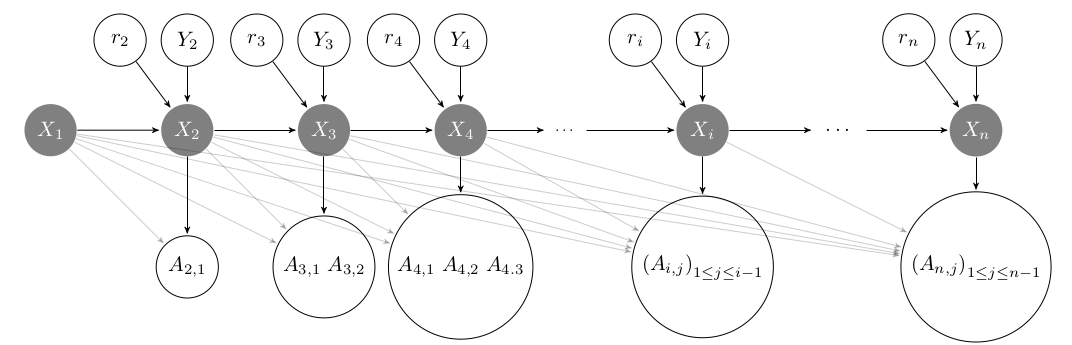}
\caption[Graphical model presenting the Markov Random Geometric Graph.]{Graphical model of the MRGG model: Markovian dynamics on Euclidean sphere where we jump from $X_k$ onto $X_{k+1}$. The $Y_k$ encodes direction of jump while $r_k$ encodes its distance, see~\eqref{e:jump}.}
\label{fig:model}
\end{center}
\vskip -0.2in
\end{figure}

Consider RGGs where latent points are sampled with Markovian jumps, the Graphical Model under consideration can be found in Figure~\ref{fig:model}. Namely, we sample~$n$ points $X_1, X_2, \dots, X_n$ on the Euclidean sphere $\mathds{S}^{d-1}$ using a Markovian dynamic. We start by sampling randomly $X_1$ on $\mathds{S}^{d-1}$. Then, for any $i \in \{2, \dots,n\}$, we sample
\begin{itemize}
\item a unit vector $Y_i \in \mathds{S}^{d-1}$ uniformly, orthogonal to $X_{i-1}$.
\item a real $r_i \in [-1,1]$ encoding the distance between $X_{i-1}$ and $X_i$, see~\eqref{e:geo}. $r_i$ is sampled from a distribution $f_{\mathcal{L}}:[-1,1] \to [0,1]$, called the \textit{latitude function}. 
\end{itemize}
then $X_i$ is defined by 
\begin{equation*}
\label{e:jump}
X_i = r_i \times X_{i-1} + \sqrt{1-r_i^2}\times  Y_i\,.
\end{equation*}
This dynamic can be pictured as follows. Consider that $X_{i-1}$ is the north pole, then chose uniformly a direction (i.e., a longitude) and, in a independent manner, randomly move along the latitudes (the longitude being fixed by the previous step). The geodesic distance $\gamma_i$ drawn on the latitudes satisfies 
\begin{equation}
\label{e:geo}
\gamma_i=\arccos(r_i)\,,
\end{equation}
where random variable $r_i=\langle X_i, X_{i-1}\rangle$ has density~$f_{\mathcal{L}}(r_i)$. The resulting model will be referred to as the Markov Random Geometric Graph (MRGG) and is described with Figure~\ref{fig:model}.

\begin{figure}
\vskip 0.2in
\begin{center}
 \centering
    \begin{subfigure}[b]{0.45\textwidth}
        \centering
        \includegraphics[width=\textwidth]{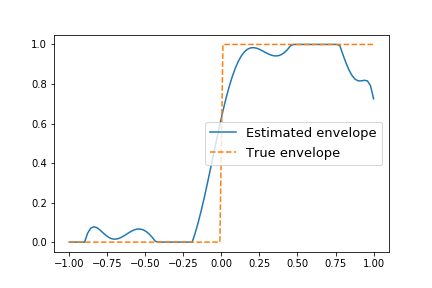}
        \caption{{\small Envelope function}}    
    \end{subfigure}
    \hfill
    \begin{subfigure}[b]{0.45\textwidth}
        \centering
        \includegraphics[width=\textwidth]{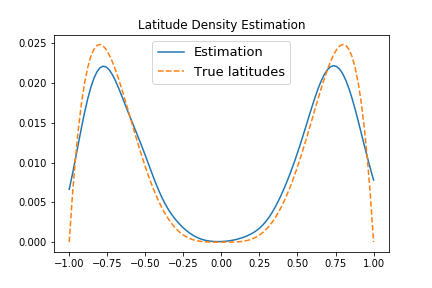}
        \caption{{\small Latitude function}}    
    \end{subfigure}
\caption[First visualization of the non-parametric estimation of envelope and latitude functions obtained with our algorithms.]{Non-parametric estimation of envelope and latitude functions using algorithms of Sections~\ref{section:envelope} and \ref{section:latitude}. We built a graph of $1500$ nodes sampled on the sphere~$\mathds{S}^{2}$ and using envelope $\mathbf{p}^{(1)}$ and latitude $f_{\mathcal L}^{(1)}$ (dot orange curves) defined in Section \ref{experiments} by Eq.\eqref{eq:simu-env-lat}. The estimated envelope is thresholded to get a function in $[0,1]$ and the estimated latitude function is normalized with integral $1$ (plain blue lines).}
\label{mix-heav}
\end{center}
\vskip -0.2in
\end{figure}

\paragraph{Temporal Dynamic Networks: MRGG estimation strategy.}  

Seldom growth models exist for temporal dynamic network modeling, see~\cite{rossetti2018community} for a review. In our model, we add one node at a time making a Markovian jump from the previous latent position. It results in 
\[
\text{the observation of }(A_{i,j})_{1\leq j\leq i-1}\text{ at time }T=i\,,
\]
as pictured in Figure~\ref{fig:model}. Namely, we observe how a new node connects to the previous ones. For such dynamic, we aim at estimating the model, namely envelope $\mathbf p$ and respectively latitude~$f_{\mathcal L}$. These functions capture in a simple function on~$\Omega=[-1,1]$ the range of interaction of nodes (represented by $\mathbf p$) and respectively the dynamic of the jumps in latent space (represented by~$f_{\mathcal L}$), where, in abscissa $\Omega$, values $r=\langle X_i,X_j\rangle$ near $1$ corresponds to close point $X_i\simeq X_j$ while values close to~$-1$ corresponds to antipodal points $X_i\simeq -X_j$. These functions may be non-parametric. 
  
\begin{figure}[!ht]
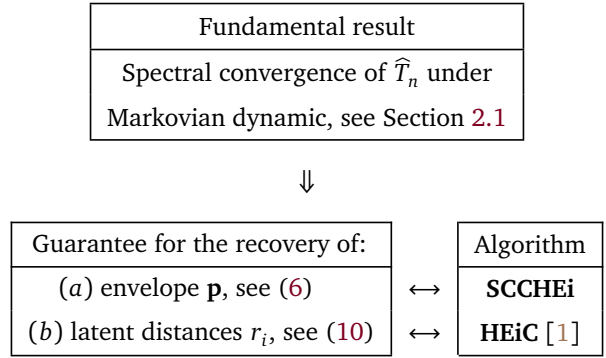

\centering
\renewcommand{\arraystretch}{1.4}
\begin{tabular}{|c|}\hline
Fundamental result \\\hline
Spectral convergence of $\widehat T_n$ under\\
Markovian dynamic, see Section~\ref{sec:integral}\\\hline
\end{tabular}
$$\Downarrow$$
\begin{tabular}{|l|c|c|}\cline{1-1}\cline{3-3}
 \makecell{Guarantee for the recovery of:} &  &  Algorithm
\\ \cline{1-1}\cline{3-3}
\makecell{$(a)$ envelope $\mathbf{p}$, see~\eqref{hatp}} & $\leftrightarrow$  & \textbf{SCCHEi}
\\
 \makecell{$(b)$ latent distances $r_i$, see~\eqref{hatG}} & $\leftrightarrow$  &  \textbf{HEiC}~\cite{AC19}
\\ \cline{1-1}\cline{3-3}
\end{tabular}
\caption{Presentation of our method to recover the envelope and the latitude functions.}
\label{approach}
\end{figure}

{\it From snapshots of the graph at different time steps, can we recover envelope and latitude functions?} We prove that it is possible under mild conditions on the Markovian dynamic of the latent points and our approach is summed up with Figure~\ref{approach}.

Define $\lambda(T_n):=(\lambda_1, \dots,\lambda_n)$ and resp. $\lambda(\widehat{T}_n):=(\hat{\lambda}_1, \dots,\hat{\lambda}_n)$ the spectrum of $T_n$ and resp. $\widehat T_n$, see~\eqref{e:enveloppe}. Building clusters from $\lambda(\widehat{T}_n)$, Algorithm~\ref{SCCHEi} (SCCHEi) estimates the spectrum of envelope~$\mathbf p$ while Algorithm~\ref{HEiC}~\cite{AC19} (HEiC, cf. Section~\ref{sec:reminder_heic}) extracts $d$ eigenvectors of $\widehat{T}_n$ to uncover the Gram matrix of the latent positions. Both can then be used to estimate the unknown functions of our model (cf. Figure~\ref{mix-heav}).

\paragraph{Previous works.}
The latent space approach to model dynamics of network has already been studied in a large span of recent works. Most of them focus on block models with dynamic generalizations covering discrete dynamic evolution via hidden
Markov models (cf.~\cite{matias17}) or continuous time analysis via extended Kalman filter (cf.~\cite{xu14}). \cite{yang18b} and \cite{durante2014bayesian} use a Gamma Markov process allowing to model evolving mixed membership in graphs using respectively the Bernoulli Poisson link function and the logistic function to generate edges from the latent space representation. While the above mentioned papers consider community based random graphs with fixed size where edges and communities change through time, we focus on growing RGGs on Euclidean sphere where new nodes are added along time.\\      Non-parametric estimation of RGGs on $\mathds S^{d-1}$ has been investigated in~\cite{CL18} with i.i.d.\! latent points. Estimation of latent point relative distances with HEiC Algorithm has been introduced in~\cite{AC19} under i.i.d.\! latent points assumption. Phase transitions on the detection of geometry in RGGs (against Erd\"os Rényi alternatives) has been investigated in~\cite{B16}. 

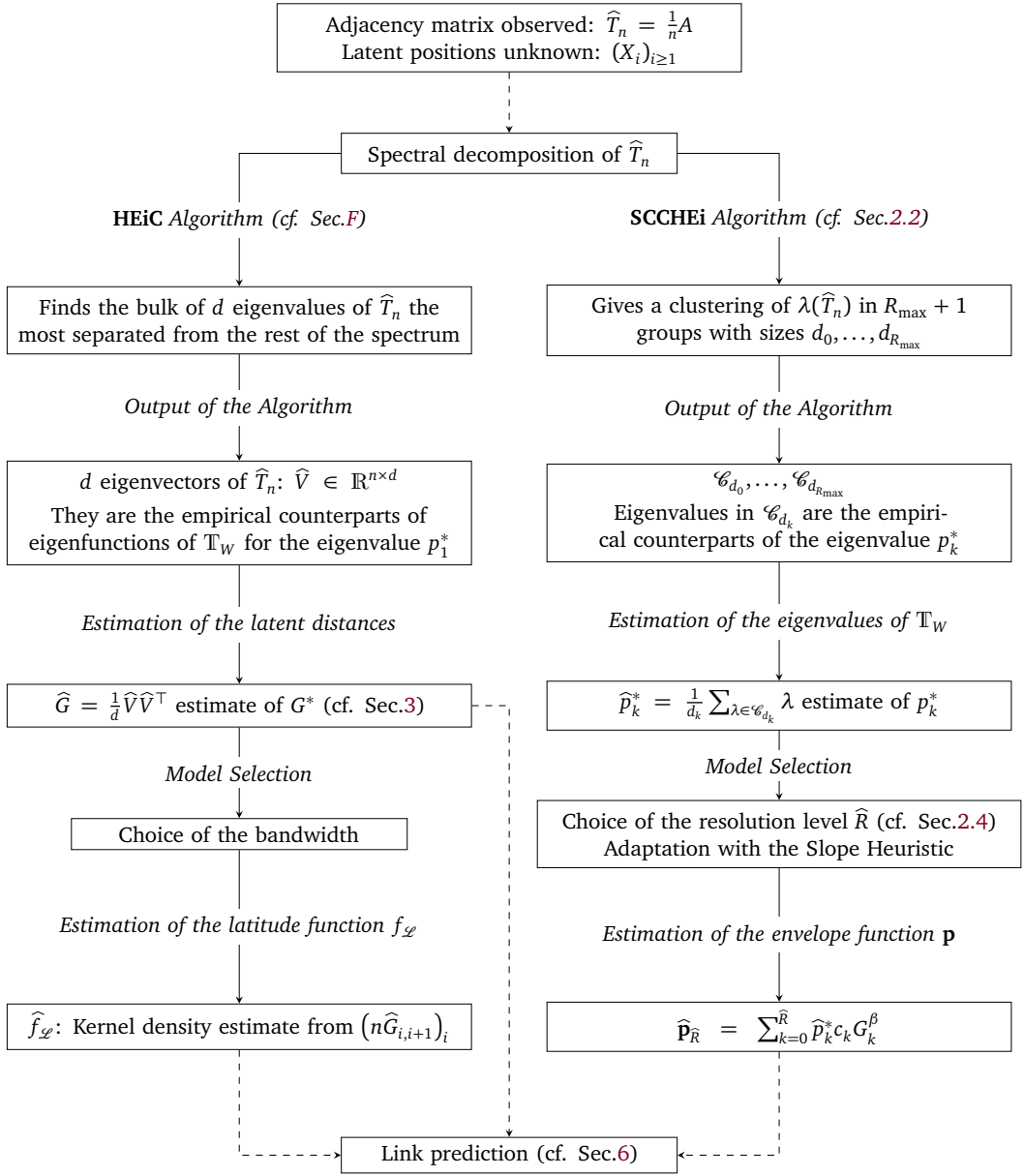
\begin{figure}[!ht]
\centering
\resizebox{.92\textwidth}{!}{
\begin{tikzpicture}[>=stealth,yscale=2,xscale=1.4]  
\node(PB) at (0,0.2)[rectangle,draw,text width=7cm,text centered] {Adjacency matrix observed: $\widehat T_n=\frac{1}{n}A$\\Latent positions unknown: $(X_i)_{i\geq1}$};
\node(eigs) at (0,-0.7)[rectangle,draw,text width=5cm,text centered] {Spectral decomposition of $\widehat T_n$};
\node(SCCHEI0) at (3,-2)[rectangle,draw,text width=7cm,text centered] {Gives a clustering of $\lambda(\widehat T_n)$ in $R_{\max}+1$ groups with sizes $d_0,\dots,d_{R_{\max}}$};
\node(HEI0) at (-3,-2)[rectangle,draw,text width=7cm,text centered] {Finds the bulk of $d$ eigenvalues of $\widehat T_n$ the most separated from the rest of the spectrum};
\node(SCCHEI1) at (3,-3.5)[rectangle,draw,text width=7cm,text centered] {$\mathcal C_{d_0}, \dots,\mathcal C_{d_{R_{\max}}}$\\\smallskip Eigenvalues in $\mathcal C_{d_k}$ are the empirical counterparts of the eigenvalue $p^*_k$};
\node(HEI1) at (-3,-3.5)[rectangle,draw,text width=7cm,text centered] {$d$ eigenvectors of $\widehat T_n$: $\widehat V\in \mathds R ^{n\times d}$\\\smallskip  They are the empirical counterparts of \\eigenfunctions of $\mathds T_{W}$ for the eigenvalue $p^*_1$};
\node(SCCHEI2) at (3,-5)[rectangle,draw,text width=7cm,text centered] {$\widehat p^*_k = \frac{1}{d_k}\sum_{\lambda\in \mathcal C_{d_k}}\lambda$ estimate of $p^*_k$};
\node(HEI2) at (-3,-5)[rectangle,draw,text width=7cm,text centered] {$\widehat G=\frac1d \widehat V \widehat V^{\top}$ estimate of $G^*$ (cf. Sec.\ref{section:latitude})};
\node(HEI3) at (-3,-6)[rectangle,draw,text width=5cm,text centered] {Choice of the bandwidth};
\node(SCCHEI3) at (3,-6)[rectangle,draw,text width=7.3cm,text centered] {Choice of the resolution level $\widehat R$ (cf. Sec.\ref{sec:slope-main})\\
Adaptation with the Slope Heuristic };
\node(HEI4) at (-3,-7.5)[rectangle,draw,text width=7cm,text centered] {$\widehat f_{\mathcal L}$: Kernel density estimate from $\big(n \widehat G_{i,i+1}\big)_{i}$};
\node(SCCHEI4) at (3,-7.5)[rectangle,draw,text width=7cm,text centered] {$\widehat {\mathbf p}_{\widehat R}=\sum_{k=0}^{\widehat R} \widehat p^*_k c_k G_k^{\beta}$};
\node(LINK) at (0,-8.5)[rectangle,draw,text width=5cm,text centered] {Link prediction (cf. Sec.\ref{sec:applications})};

\draw[dashed,->] (PB) -- (eigs);
 \draw[->] (eigs) -| (HEI0) node[pos=0.75,fill=white,text width=5.5cm,text centered,xshift=0cm]{\it  {\bf HEiC} Algorithm (cf. Sec.\ref{sec:reminder_heic}) \\
 };
\draw[->] (eigs) -| (SCCHEI0) node[pos=0.75,fill=white,text width=5.5cm,text centered,xshift=0cm]{\it  {\bf SCCHEi} Algorithm (cf. Sec.\ref{subsec:SCCA}) \\
 };
 \draw[->] (HEI0) -- (HEI1) node[pos=0.5,fill=white,text width=5.5cm,text centered,xshift=0cm]{\it  Output of the Algorithm 
 };
\draw[->] (SCCHEI0) -- (SCCHEI1) node[pos=0.5,fill=white,text width=5.5cm,text centered,xshift=0cm]{\it Output of the Algorithm 
 };
 \draw[->] (HEI1) -- (HEI2) node[pos=0.5,fill=white,text width=5.5cm,text centered,xshift=0cm]{\it  Estimation of the latent distances 
 };
\draw[->] (SCCHEI1) -- (SCCHEI2) node[pos=0.5,fill=white,text width=5.5cm,text centered,xshift=0cm]{\it Estimation of the eigenvalues of $\mathds T_W$ 
 };
  \draw[->] (HEI2) -- (HEI3) node[pos=0.5,fill=white,text width=5.5cm,text centered,xshift=0cm]{\it  Model Selection 
 };
\draw[->] (SCCHEI2) -- (SCCHEI3) node[pos=0.5,fill=white,text width=5.5cm,text centered,xshift=0cm]{\it Model Selection
 };
  \draw[->] (HEI3) -- (HEI4) node[pos=0.5,fill=white,text width=6.5cm,text centered,xshift=0cm]{\it  Estimation of the latitude function $f_{\mathcal L}$ };
\draw[->] (SCCHEI3) -- (SCCHEI4) node[pos=0.5,fill=white,text width=5.5cm,text centered,xshift=0cm]{\it Estimation of the envelope function $\mathbf p$ };
\draw[dashed,->] (SCCHEI4) |- (LINK);
\draw[dashed,->] (HEI4) |- (LINK);
\draw[dashed,->] (HEI2) -| (LINK);
\end{tikzpicture}}
\caption{Synthetic presentation of the different estimation procedures.}
\label{fig:MRGG-synthesis}
\end{figure}

For the first time, we introduce latitude function and non-parametric estimations of envelope and latitude using new results on kernel matrices concentration with dependent variables. 
\paragraph{Outline} 
Sections \ref{section:envelope} and \ref{section:latitude} present the estimation method with new theoretical results under Markovian dynamic. These new results are random matrices operator norm control and resp. U-statistics control under Markovian dynamic, presented in the Appendix at Section~\ref{proof-delta2-proba-ope} and resp. Section~\ref{proof-ustat}. The envelope {\it adaptive} estimate is built from a size constrained clustering (Algorithm~\ref{SCCHEi}) tuned by slope heuristic Eq.\eqref{R-kappa}, and the latitude function estimate (cf. Section~\ref{sec:latitude_nutshell}) is derived from estimates of latent distances $r_i$. Our method can handle random graphs with logarithmic growth node degree (i.e., new comer at time $T=n$ connects to $\mathcal O(\log n)$ previous nodes), referred to as {\it relatively sparse} models, see Section~\ref{sec:relatively_sparse}. Sections \ref{experiments} and \ref{sec:applications} investigate synthetic data experiments. We propose heuristics to solve link prediction problems and to test for a Markovian dynamic. In a last section (Section~\ref{sec:discussion}), we dig deeper into the analysis of our methods by studying their behaviour under model mispecification or under slow mixing conditions. We conclude by presenting final remarks and future research directions.
 At the end of Section~\ref{sec:discussion}, we provide with Figure~\ref{fig:MRGG-synthesis} a synthetic presentation of the estimation methods of this paper.

\paragraph{Notations.}
Consider a dimension $d \geq 3$. Denote by $\| \cdot \|_2$ (resp. $\langle \cdot , \cdot \rangle$) the Euclidean norm (resp. inner product) on $\mathds{R}^{d}$. Consider the $d$-dimensional sphere $\mathds{S}^{d-1}:=\{ x \in \mathds{R}^d\;:\; \|x\|_2=1 \}$ and denote by $\pi$ the uniform probability measure on $\mathds{S}^{d-1}$. For any matrix $M=(m_{i,j})_{i,j}\in\mathds R^{D_1\times D_2}$, we define $\|M\|_F^2:=\sum_{i=1}^{D_1}\sum_{j=1}^{D_2} |m_{i,j}|^2 $ and the operator norm of $M$ as $\|M\|:=\sup_{x\in \mathds S^{D_2-1}} \|Mx\|_2.$ For two real valued sequences $(u_n)_{n \in \mathds{N}}$ and~$(v_n)_{n \in \mathds{N}}$, denote $u_n \underset{n \to \infty}{=} \mathcal{O}(v_n)$ if there exist $k_1 >0 $ and $n_0 \in \mathds{N}$ such that 
$\forall n >n_0$, $|u_n| \leq k_1 |v_n|$. For any $x,y \in \mathds R$, $x\wedge y:= \min(x,y)$ and $x\vee y:=\max(x,y).$ Given two sequences $x,y$ of reals--completing finite sequences by zeros--such that $\sum_i x_i^2+y_i^2<\infty$, we define the $\ell_2$ rearrangement distance $\delta_2(x,y)$ as 
\[\delta_2^2(x,y):= \inf_{\sigma \in \mathfrak{S}}\sum_i (x_i-y_{\sigma(i)})^2\,,
\]
where $\mathfrak{S}$ is the set of permutations with finite support. This pseudo-distance is useful to compare two spectra.

\clearpage

\section{Nonparametric estimation of the envelope function}
\label{section:envelope}\label{model}

One can associate with $W(x,y)=\mathbf{p}(\langle x,y \rangle)$ the integral operator \[\mathds{T}_W : L^2(\mathds{S}^{d-1})\to L^2(\mathds{S}^{d-1}),\] such that for any $ g \in L^2(\mathds{S}^{d-1}),$
\begin{equation} \notag \forall x \in \mathds{S}^{d-1}, \quad (\mathds{T}_Wg)(x)=\int_{\mathds{S}^{d-1}} g(y)\mathbf{p}(\langle x,y \rangle) \pi(dy),\end{equation}
where $\pi$ is the uniform probability measure on $\mathds{S}^{d-1}$. The operator $\mathds{T}_W$ is Hilbert-Schmidt and it has a countable number of bounded eigenvalues $\lambda_k^*$ with zero as only accumulation point. The eigenfunctions of~$\mathds{T}_W$ have the remarkable property that they do not depend on $\mathbf p$ (cf.~\cite{Xu} Lemma 1.2.3): they are given by the real Spherical Harmonics. We denote $\mathcal{H}_l$ the space of real Spherical Harmonics of degree $l$ with dimension~$d_l$ and with orthonormal basis $(Y_{l,j})_{j \in [d_l]}$ where
\[ d_l:= \mathrm{dim}(\mathcal{H}_l) =  \left\{
    \begin{array}{ll}
        1 & \mbox{if } l=0 \\
        d & \mbox{if } l=1 \\
        \binom{l+d-1}{l}-\binom{l+d-3}{l-2} & \mathrm{otherwise}.
    \end{array}
\right.\]
We end up with the following spectral decomposition
\begin{equation}
    \label{reconstruction-p0}
    \mathbf{p}(\langle x,y \rangle)= \sum_{l\geq 0} p^*_l \sum_{1 \leq j\leq d_l} Y_{l,j}(x)Y_{l,j}(y)=\sum_{k\geq 0}p^*_kc_kG_k^{\beta}(\langle x,y \rangle)\,,
\end{equation}
where $\lambda(\mathds{T}_W)= \{p_0^*,p_1^*, \dots ,p_1^*, \dots, p_l^*, \dots ,p_l^*, \dots \}$ meaning that each eigenvalue $p^*_l$ has multiplicity $d_l$; and $G_k^{\beta}$ is the Gegenbauer polynomial of degree $k$ with parameter $\beta:=\frac{d-2}{2}$ and $c_k:= \frac{2k+d-2}{d-2}$ (cf. Appendix \ref{apdx:Harmonic-Analysis}). Since $\mathbf p$ is bounded, one has $\mathbf{p}\in L^2((-1,1),w_{\beta})$ where the weight function $w_{\beta}$ is defined by $w_{\beta}(t):=(1-t^2)^{\beta-\frac12}$ and
\[L^2((-1,1),w_{\beta}):=\big\{g:[-1,1] \to \mathds R \; \big|\; \|g\|_2^2:=\int_{-1}^1 |g(t)|^2 w_{\beta}(t)dt<+\infty\big\}.\]$\mathbf{p}$ can be decomposed as $\mathbf{p}\equiv \sum_{k\geq 0}p^*_kc_kG_k^{\beta}$ and the Gegenbauer polynomials $G_k^{\beta}$ are an orthogonal basis of $L^2((-1,1),w_{\beta})$.

We finally introduce for any resolution level $R \in \mathds N$ the truncated graphon $W_R$ which is obtained from $W$ by keeping only the $\widetilde{R}$ first eigenvalues, that is
\begin{equation*}\label{eq:WR-def}\forall x,y \in \mathds{S}^{d-1}, \quad W_R(x,y): = \sum_{k=0}^R p^*_k \sum_{l=1}^{d_k} Y_{k,l}(x)Y_{k,l}(y).\end{equation*} Similarly, we denote for all $t\in [0,1]$, $\mathbf{p}_R(t) = \sum_{k=0}^Rp^*_kc_kG_k^{\beta}(t)$.

\paragraph{Weighted Sobolev space} The space $Z^s_{w_{\beta}}((-1,1))$ with regularity $s >0$ is defined as the set of functions $g=\sum_{k \geq0} g^*_k c_k G_k^{\beta}\in L^2((-1,1),w_{\beta})$ such that
\[\| g\|_{Z^s_{w_{\beta}}((-1,1))}^*:= \left[ \sum_{l=0}^{\infty}d_l |g_l^*|^2\left(1+ (l(l+2\beta))^s\right) \right]^{1/2}<\infty.\]

\clearpage
\subsection{Integral operator spectrum estimation with dependent variables}
\label{sec:integral}

One key result is a new control of $U$-statistics with latent Markov variables (cf. Section~\ref{proof-ustat}) and it makes use of a Talagrand's concentration inequality for Markov chains. This article follows the hypotheses made on the Markov chain $(X_i)_{i\geq1}$ by~\cite{duchemin20}. Namely, we work under the following assumption.
\smallskip

\noindent \textbf{Assumption A} \label{assumptionA} The latitude function $f_{\mathcal{L}}$ is such that $\|f_{\mathcal{L}}\|_{\infty} < \infty$ and makes the chain $(X_i)_{i\geq 1}$ uniformly ergodic.
\smallskip

Under \hyperref[assumptionA]{Assumption A}, we prove in Section~\ref{apdx:properties-MC} that the unique stationary distribution of the Markov chain~$(X_i)_{i \geq1}$ is the uniform probability measure on $\mathds S^{d-1}$ denoted $\pi$.
Theorem \ref{thm:delta2-proba-ope} is a theoretical guarantee for a random matrix approximation of the spectrum of integral operator with {\bf dependent} latent variables. Theorem \ref{delta2-proba-ope} in Section~\ref{proof-delta2-proba-ope} gives explicitly the constants hidden in the big O below which depend on the absolute spectral gap of the Markov chain $(X_i)_{i \geq 1}$ (cf. Definition~\ref{apdx:spectralgap}).  
\begin{theorem}
\label{thm:delta2-proba-ope} 
We consider that \hyperref[assumptionA]{Assumption A} holds and we assume the envelope $\mathbf p$ has regularity $s>0$. Then, it holds 
\begin{align}\notag &\mathds{E}\left[\delta_2^2(\lambda(\mathds{T}_W),\lambda(T_n)) \right]=\quad \mathcal{O}\left(\left[ \frac{n}{\log^2(n)} \right]^{- \frac{2s}{2s+d-1}}\right).\end{align}
Using this preliminary result and the near optimal error bound for the operator norm of random matrices from~\cite{bandeira2016} we obtain
\begin{align}\notag &\mathds{E}\left[ \delta_2^2(\lambda(\mathds{T}_W),\lambda^{R_{opt}}(\widehat{T}_n)) \right]=\quad \mathcal{O}\left(\left[ \frac{n}{\log^2(n)} \right]^{- \frac{2s}{2s+d-1}}\right),\end{align}
with $\lambda^{R_{opt}}(\widehat{T}_n)=(\hat{\lambda}_1, \dots,\hat{\lambda}_{\widetilde R_{opt}}, 0,0,\dots)$ and $R_{opt}=\floor{\left(n/\log^2(n)\right)^{\frac{1}{2s+d-1}}}$. $\hat{\lambda}_1, \dots,\hat{\lambda}_n$ are the eigenvalues of $\widehat{T}_n$ sorted in decreasing order of magnitude. 
\end{theorem}

\textbf{Remark.} In Theorem~\ref{thm:delta2-proba-ope} and Theorem~\ref{gram}, note that we recover, up to a $\log$ factor, the {\it minimax rate of non-parametric estimation} of $s$-regular functions on a space of (Riemannian) dimension $d-1$. Even with i.i.d. latent variables, it is still an open question to know if this rate is the minimax rate of non-parametric estimation of RGGs.

Eq.\eqref{reconstruction-p0} shows that one could use an approximation of $(p^*_k)_{k \geq 1}$ to estimate the envelope~$\mathbf{p}$ and Theorem~\ref{thm:delta2-proba-ope} states we can recover $(p^*_k)_{k \geq 1}$ up to a permutation. In most cases, the problem of finding such a permutation is NP-hard and we introduce in the next section an efficient algorithm to fix this issue.

\subsection{Size Constrained Clustering Algorithm}
\label{subsec:SCCA}

Note the spectrum of $\mathds{T}_W$ is given by $(p_l^*)_{l\geq 0}$ where $p_l^*$ has multiplicity $d_l$. In order to recover envelope $\mathbf p$, we build clusters from eigenvalues of $\widehat T_n$ while respecting the dimension $d_l$ of each eigen-space of $\mathds{T}_W$. In~\cite{CL18}, an algorithm is proposed testing all permutations of $\{0, \dots, R\}$ for a given maximal resolution $R$. To bypass the high computational cost of such approach, we propose an efficient method based on the tree built from {\it Hierarchical Agglomerative Clustering}~(HAC). In the following, for any $\nu_1, \dots ,\nu_n \in \mathds{R}$, we denote by $\mathrm{HAC}(\{\nu_1, , \dots, \nu_n\}, d_c)$ the tree built by a HAC on the real values $\nu_1, \dots ,\nu_n $ using the complete linkage function $d_c$ defined by  $\forall A,B \subset \mathds{R}$, $d_c(A,B) = \max_{a\in A} \max_{b \in B} \| a-b \|_2$. Algorithm \ref{SCCHEi} describes our approach.

\begin{algorithm}
{\small\textbf{Data: } Resolution $R$, matrix $\widehat T_n=\frac1n A$, dimensions $(d_k)_{k=0}^R$.

\begin{algorithmic}[1]
\STATE Let $\hat{\lambda}_1, \dots , \hat{\lambda}_n$ be the eigenvalues of $\widehat{T}_n$ sorted in decreasing order of magnitude. 
\STATE Set $\mathcal{P}:=\{ \hat{\lambda}_1, \dots , \hat{\lambda}_{\widetilde{R}}\}$ and $dims=[d_0,d_1, \dots,d_R]$.
    \WHILE{All eigenvalues in $\mathcal{P}$ are not clustered} \alglinelabel{line:goto}
    \STATE $tree \leftarrow $ HAC(nonclustered eigenvalues in $\mathcal{P}$, $d_c$) \;
        \FOR{ $d \in dims$ }
            \STATE Search for a cluster of size $d$ in $tree$ as close as possible to the root.
            \STATE \textbf{if} such a cluster $\mathcal{C}_d$ exists \textbf{then} Update$(dims,tree,\mathcal{C}_d,d)$. 
        \ENDFOR
        \FOR{$d \in dims$}
            \STATE Search for the group $\mathcal{C}$ in $tree$ with a size larger than $d$ and as close as possible to $d$.
            \STATE \textbf{if} such a group exists \textbf{then} Update$(dims,tree,\mathcal{C},d)$ \textbf{else} Go to line \ref{line:goto}.
        \ENDFOR
    \ENDWHILE
 	\end{algorithmic}
 	\textbf{Return:} $\mathcal{C}_{d_0},\dots,\mathcal{C}_{d_R},\{\hat{\lambda}_{\widetilde{R}+1}, \dots , \hat{\lambda}_n \}$ }
 \caption{Size Constrained Clustering for Harmonic Eigenvalues (SCCHEi).}
 \label{SCCHEi}
\end{algorithm}
\begin{algorithm}
\begin{algorithmic}[1]
 	\STATE Save the subset $\mathcal{C}_d$ consisting of the $d$ eigenvalues in $\mathcal{C}$ with the largest absolute values. \STATE Delete from $tree$ all occurrences to eigenvalues in $\mathcal{C}_d$ and delete $d$ from $dims$.
\end{algorithmic}
\caption{Update($dims,tree,\mathcal{C},d)$.}
\end{algorithm}

Given some resolution level $R\in \mathds N$, our estimator $\widehat {\mathbf{p}}_R$ of the envelope function $\mathbf{p}$ is obtained from the clustering of the eigenvalues obtained by the SCCHEi algorithm as follows
\begin{equation}\label{eq:hatpR}\widehat {\mathbf{p}}_R:t\mapsto \sum_{k=0}^R\widehat p_k c_kG_k^{\beta}(t) \quad \text{where} \quad \forall k \in \{0,\dots,R\}, \quad  \widehat p_k:= \frac{1}{d_k} \sum_{\lambda \in \mathcal C_{d_k}} \lambda.\end{equation}

\subsection{Theoretical guarantees}

\label{sec:theo-guarantees}

Let us recall that for any resolution level $R\geq 0$, 
\[\lambda(\mathds T_{W_{R}}) = (\lambda^*_1, \dots,\lambda^*_{\widetilde R},0,0,\dots) \; \text{and}\; \lambda^{R}(\widehat{T}_n)=(\hat \lambda_1, \dots,\hat \lambda_{\widetilde R}, 0,0,\dots)\]
where $\hat \lambda_1, \dots,\hat \lambda_n$ are the eigenvalues of $\widehat{T}_n$ sorted in decreasing order of magnitude. We order the eigenvalues $\hat \lambda_1, \dots,\hat \lambda_{\widetilde R}$ and in the following we consider that $\lambda^{R}(\widehat{T}_n)_1\geq\dots\geq\lambda^{R}(\widehat{T}_n)_{\widetilde R}$.

\begin{theorem} \label{thm:SCCHEi}
Let us consider some resolution level $R\in \mathds N$. We keep the assumptions of Theorem~1. We recall that we consider $\lambda^{R}(\widehat{T}_n)_1\geq\dots\geq\lambda^{R}(\widehat{T}_n)_{\widetilde R}$.

Then for $n$ large enough, the clusters $\mathcal C_{d_0}, \dots, \mathcal C_{d_R}$ obtained from the SCCHEi algorithm satisfy
\begin{equation*} \label{eq:thm-SCCHEi}\delta_2^2(\lambda(\mathds{T}_{W_{R}}),\lambda^{R}(\widehat{T}_n)) = \sum_{k=0}^R \sum_{\hat \lambda\in \mathcal C_{d_k}} (\hat \lambda - p^*_k)^2.
\end{equation*}
\end{theorem}

\begin{proof}[Proof of Theorem~\ref{thm:SCCHEi}]
Let us denote 
\[\Delta^G = \min_{0 \leq k\neq l\leq R,\; p^*_k \neq p^*_l} \; |p^*_k - p^*_l| \wedge \min_{ 0\leq k\leq R,\;  p^*_k \neq 0} \; |p^*_k| >0.\]
For any $g \in (0, \frac{\Delta^G}{4}),$ the proof of Theorem~\ref{thm:delta2-proba-ope} (cf. Section~\ref{proof-delta2-proba-ope}) ensures that for $n$ large enough it holds
\begin{equation} \label{delta2-control}\delta_2^2(\lambda(\mathds{T}_{W_{R}}),\lambda^{R}(\widehat{T}_n))\leq g^2 .\end{equation}
Let us recall that
\begin{align*}\delta_2^2(\lambda(\mathds{T}_{W_{R}}),\lambda^{R}(\widehat{T}_n))&=\inf_{\sigma \in \mathfrak{S}} \sum_{i\geq 1} \left(\lambda(\mathds{T}_{W_{R}})_{\sigma(i)} -\lambda^{R}(\widehat{T}_n)_i \right)^2 \label{eq:delta2-hac}.
\end{align*}
The proof of Theorem~\ref{thm:SCCHEi} relies on the following two Lemmas. The proofs of these Lemmas are postponed to Section~\ref{apdx:lemmas-SCCHEi}.
\clearpage

\begin{lemma}\label{lemma1:SCCHEi}
We keep the assumptions of Theorem~\ref{thm:SCCHEi}. Then, for $n$ large enough for Eq.\eqref{delta2-control} to hold, one can choose a permutation $\sigma^*$ such that 
\begin{itemize}
\item $\sigma^*(\{1, \dots, \widetilde R\})=\{1, \dots, \widetilde R\}$.
\item $\delta_2^2(\lambda(\mathds{T}_{W_{R}}),\lambda^{R}(\widehat{T}_n)) = \sum_{i=1}^{\widetilde R} (\lambda(\mathds{T}_{W_{R}})_{\sigma^*(i)}-  \lambda^{R}(\widehat{T}_n)_i)^2$.
\end{itemize}
Moreover, the function $f^*$ given by 
\begin{align*}
f^*:\{1,\dots, \widetilde R\} &\rightarrow \{p^*_k,\; 0\leq k \leq R\}\\
i &\mapsto \lambda(\mathds{T}_{W_{R}})_{\sigma^*(i)},
\end{align*}
is non-increasing.
\end{lemma}

\begin{lemma}\label{lemma2:SCCHEi}
We keep the assumptions and notations of Lemma~\ref{lemma1:SCCHEi}.
A clustering $\left( \widehat {\mathcal C}_{d_k} \right)_{0\leq k\leq R}$ at depth $R$ in the tree of the HAC algorithm applied to $\mathcal P:=\{\lambda^{R}(\widehat{T}_n)_{1},\dots ,\lambda^{R}(\widehat{T}_n)_{\widetilde R}\}$ is said to be of type $(\mathcal S)$ if it satisfies:
\begin{align*}\widehat {\mathcal C}_{d_0}\subset & \{\lambda^{R}(\widehat{T}_n)_{i} \; |\; 1\leq i\leq \widetilde R, \;  f^*(i)=p^*_0\}, \quad |\widehat {\mathcal C}_{d_0}|=d_0,\\
\widehat {\mathcal C}_{d_1}\subset & \{\lambda^{R}(\widehat{T}_n)_{i} \; |\; 1\leq i\leq \widetilde R, \;  f^*(i)=p^*_1\}, \quad |\widehat {\mathcal C}_{d_1}|=d_1,\\
&\dots\\
\widehat {\mathcal C}_{d_{R}}\subset & \{\lambda^{R}(\widehat{T}_n)_{i} \; |\; 1\leq i\leq \widetilde R, \;  f^*(i)=p^*_{R}\},  \quad |\widehat {\mathcal C}_{d_R}|=d_R.
\end{align*}
Then 
the HAC algorithm with complete linkage applied to $\mathcal P$ reaches (after $\widetilde R - R-1$ iterations) a state $\left( \widehat {\mathcal C}_{d_k} \right)_{0\leq k\leq R}$ of type $(\mathcal S)$. As a consequence, the SCCHEi algorithm returns the clusters $\mathcal C_{d_0}=\widehat {\mathcal C}_{d_0}, \dots, \mathcal C_{d_{R}}=\widehat {\mathcal C}_{d_{R}}$.
\end{lemma}

Theorem~\ref{thm:SCCHEi} directly follows from the conclusion of Lemma~\ref{lemma2:SCCHEi} since we get that
\begin{align*} \sum_{k=0}^R \sum_{\hat \lambda\in \mathcal C_{d_k}} (\hat \lambda - p^*_k)^2&=\sum_{i=1}^{\widetilde R} (\lambda^{R}(\widehat{T}_n)_i - f^*(i))^2=\sum_{i=1}^{\widetilde R} (\lambda^{R}(\widehat{T}_n)_i - \lambda(\mathds{T}_{W_{R}})_{\sigma^*(i)})^2\\&=\delta_2^2(\lambda(\mathds{T}_{W_{R}}),\lambda^{R}(\widehat{T}_n)) ,
\end{align*}
where the first equality comes from the conclusion of Lemma~\ref{lemma2:SCCHEi}, the second one comes from the definition of $f^*$ from Lemma~\ref{lemma1:SCCHEi} and the last one comes from the choice of $\sigma^*$ from Lemma~\ref{lemma1:SCCHEi}.
\end{proof}

\bigskip

Theorem~\ref{thm:SCCHEi} ensures that under appropriate conditions, the SCCHEi leads to a clustering of the eigenvalues of the adjacency matrix that achieves the $\delta_2$ distance between $\lambda(\mathds{T}_{W_{R}})$ and $\lambda^{R}(\widehat{T}_n)$. Nevertheless, this is not a sufficient condition to ensure that the $L^2$ error between the true envelope function and our plug-in estimator (cf. Eq.\eqref{eq:hatpR}) goes to $0$ has $n \to +\infty.$ This is due to identifiability issues coming from the $\delta_2$ metric. This was already mentioned in \cite[Section 3.6]{CL18}, where the authors present the following example.
Consider the case $d = 3,$ which implies $\beta = 1/2$, $d_k = 2k + 1$, $c_k = 2k + 1$. For $\mu > 0$, let
\begin{align*}
 \mathbf{p}_a &= \frac12 c_0 G_0^{\beta}+\mu c_1 G_1^{\beta}+0\times c_2 G_2^{\beta}+0\times c_3 G_3^{\beta}+\mu c_4G_4^{\beta}\\
 \mathbf{p}_b &= \frac12 c_0 G_0^{\beta}+0\times c_1 G_1^{\beta}+\mu c_2 G_2^{\beta}+\mu c_3 G_3^{\beta}+0\times c_4G_4^{\beta}
\end{align*}
Then the associated spectrum are
\begin{align*}
\lambda_a^* = (1/2, \underbrace{\mu, \mu, \mu}_{3}, \underbrace{0, 0, 0, 0, 0}_{5},\underbrace{0, 0, 0, 0, 0, 0, 0}_{7}, \underbrace{\mu, \mu, \mu, \mu, \mu, \mu, \mu, \mu, \mu}_{9})\\
\lambda_b^* =  (1/2, \underbrace{0, 0, 0}_{3}, \underbrace{\mu, \mu, \mu, \mu, \mu}_{5},\underbrace{\mu, \mu, \mu, \mu, \mu,\mu,\mu}_{7}, \underbrace{0,0,0,0,0,0,0,0,0}_{9})
\end{align*}
which are indistinguishable in $\delta_2$ metric, although $\|\mathbf{p}_a - \mathbf{p}_b\|_2 = \mu
\sqrt{24}$. 

Nevertheless, we can obtain a theoretical guarantee on the $L^2$ error between the true envelope function and our plug-in estimate using Theorem~\ref{thm:SCCHEi} if we consider additional conditions on the eigenvalues $(p^*_k)_{k\geq0}$.

\begin{theorem} \label{thm:poly}
Assume that the envelope function $\mathbf{p}$ is polynomial of degree $D\in \mathds N$, i.e., $p^*_k = 0$ for any $k> D$ and $p^*_D\neq0$. Assume also that all nonzeros $p_k^*$ for $k \in \{0, \dots , D\}$ are distinct and that $R\geq D$. Then for $n$ large enough it holds with probability at least $1-n^{-8}$,
\[\| \widehat {\mathbf{p}}_R - \mathbf{p}\|_2^2 \leq c\frac{\widetilde R}{n}  \ln(n),\]
where $c>0$ is a universal numerical constant.
\end{theorem}
\noindent{\bf Remarks.}
\begin{itemize}
\item The question of whether the problem of estimating $ \mathbf{p} $ is NP-hard was still completely open. Theorem~\ref{thm:poly} brings a first partial answer to this question by showing that $\mathbf p$ can be estimated in polynomial time in the case where $\mathbf{p}$ is a polynomial with all non-zero eigenvalues distinct.
\item The proof of Theorem~\ref{thm:poly} is strictly analogous to the one of \cite[Proposition 9]{CL18}. In a nutshell, considering that the envelope function $\mathbf{p}$ is a polynomial with all non-zeros eigenvalues $p^*_k$ distinct ensures that (since $R\geq D$) \[\delta_2^2(\lambda(\mathds{T}_{W_R}),\lambda^R(\widehat{T}_n))=\delta_2^2(\lambda(\mathds{T}_{W}),\lambda^R(\widehat{T}_n)),\]which coincides with the $L^2$ norm of the difference between $\mathbf{p}$ and its estimate \[\widehat {\mathbf{p}}_{opt,R}:=\sum_{k=0}^R \widehat p_{opt,k} c_k G_k^{\beta} \quad \text{with} \quad \widehat p_{opt,k}:=\frac{1}{d_k} \sum_{ i \in (\sigma^*)^{-1}([\widetilde k +1, \widetilde{k+1}])} \lambda^{R}(\widehat{T}_n)_i,\] where $\sigma^*$ is a permutation as defined in Lemma~\ref{lemma1:SCCHEi}. Since we proved that for $n$ large enough, the clusters returned by the SCCHEi algorithm correspond to an allocation given by $f^*$, we deduce that the $L^2$ norm between $\mathbf{p}$ and our plug-in estimate $\widehat {\mathbf{p}}_R$ is equal to the $\delta_2$ distance between spectra. The result then comes directly using Theorem~\ref{thm:delta2-proba-ope}.
\end{itemize}

\subsection{Adaptation: Slope heuristic as model selection of Resolution}
\label{sec:slope-main}

A data-driven choice of model size $R$ can be done by {\it slope heuristic}, see~\cite{arlot2019} for a nice review. One main idea of slope heuristic is to penalize the empirical risk by $\kappa\,\mathrm{pen}(\widetilde R)$ and to calibrate $\kappa>0$. If the sequence $(\mathrm{pen}(\widetilde R))_{\widetilde R}$ is equivalent to the sequence of variances of the population risk of empirical risk minimizer~(ERM) as model size $\widetilde R$ grows, then, penalizing the empirical risk (as done in Eq.\eqref{R-kappa}), one may ultimately uncover an empirical version of the $U$-shaped curve of the population risk. Hence, minimizing it, one builds a model size $\hat R$ that balances between bias ({\it under-fitting} regime) and variance ({\it over-fitting} regime). First, note that empirical risk is given by the intra-class variance below.
\begin{definition} \label{def:intraclass}
For any output $(\mathcal{C}_{d_0}, \dots ,\mathcal{C}_{d_R},\Lambda)$ of the Algorithm SCCHEi, the thresholded intra-class variance is defined by  \begin{equation}\notag\mathcal{I}_R:= \frac{1}{n}\left[ \sum_{k=0}^R \sum_{\lambda \in \mathcal{C}_{d_k} }\left( \lambda- \frac{1}{d_k}\sum_{\lambda' \in \mathcal{C}_{d_k} }\lambda' \right)^2 + \sum_{\lambda \in \Lambda } \lambda^2 \right]\,,\end{equation}
and the estimations $(\hat{p}_k)_{k \geq 0}$ of the eigenvalues $(p^*_k)_{k \geq 0}$ is given by 
\begin{equation}\label{hatp}\forall k \in \mathds{N},\quad \hat{p}_k = \left\{
    \begin{array}{ll}
        \frac{1}{d_k} \sum_{ \lambda \in \mathcal{C}_{d_k}} \lambda  & \mbox{if } k \in \{0,\dots,\hat{R} \}\\
        0 & \mbox{otherwise.}
    \end{array}
\right.  \end{equation}
\end{definition}

Second, as underlined in the proof of Theorem~\ref{thm:delta2-proba-ope} (see Theorem \ref{delta2-proba-ope} in Section~\ref{proof-delta2-proba-ope}), the estimator's variance of our estimator scales linearly in $\widetilde R$. 

Hence, we apply Algorithm SCCHEi for $R$ varying from $0$ to $R_{\max}$ (with $R_{\max}:= \max \{R \geq0\;:\; \widetilde{R} \leq n\}$) to compute the \textit{thresholded intra-class variance} $\mathcal{I}_R$ (see Definition~\ref{def:intraclass}) and given some $\kappa>0$, we select 
\begin{equation} 
\label{R-kappa} 
R(\kappa) 
\in \underset{R \in \{0,\dots,R_{max}\}}{\arg\min} 
\Big\{\mathcal I_R + \kappa  \frac{\widetilde R}n\Big\}\,.
\end{equation} 
The hyper-parameter $\kappa$ controlling the bias-variance trade-off is set to $2\kappa_0$ where $\kappa_0$ is the value of $\kappa>0$ leading to the ‘‘{\it largest jump}'' of the function $\kappa \mapsto R(\kappa)$. Once $\hat{R}:=R(2\kappa_0)$ has been computed, we approximate the envelope function $\mathbf{p}$ using Eq.\eqref{hatp} (see Eq.\eqref{reconstruction-p} for the closed form). We denote this estimator~$\widehat {\mathbf p}$ and with the notations of Eq.\eqref{eq:hatpR} it holds~$\widehat {\mathbf p} = \widehat {\mathbf p}_{\widehat R}$. In Appendix~\ref{apdx:slope-real-data}, we describe this slope heuristic on a concrete example and our results can be reproduced using the notebook 
\href{https://github.com/quentin-duchemin/Markovian-random-geometric-graph}{\textit{Experiments}}\footnote{\label{git}\url{https://github.com/quentin-duchemin/Markovian-random-geometric-graph}} in the Appendix.

\section{Nonparametric estimation of the latitude function}
\label{section:latitude}

\subsection{Our approach to estimate the latitude function in a nutshell}
\label{sec:latitude_nutshell}

In Theorem~\ref{gram} (see below), we show that we are able to estimate consistently the pairwise distances encoded by the Gram matrix $G^*$ where \begin{equation}
\notag G^* := \frac{1}{n} \left( \langle X_i,X_j \rangle \right)_{i,j \in [n]}. 
\end{equation} 
Taking the diagonal just above the main diagonal (referred to as \textit{superdiagonal}) of $\widehat{G}$ - an estimate of the matrix $G$ to be specified - we get estimates of the i.i.d. random variables $\left(\langle X_i,X_{i-1} \rangle\right)_{2\leq i\leq n}=\left(r_i\right)_{2\leq i\leq n}$ sampled from $f_{\mathcal{L}}$. Using~$\left(\hat{r}_i\right)_{2\leq i\leq n}$ the superdiagonal  of $n\widehat{G}$, we can build a kernel density estimator of the latitude function $f_{\mathcal{L}}$. In the following, we describe the algorithm used to build our estimator $\hat{G}$ with theoretical guarantees.

\subsection{Spectral gap condition and Gram matrix estimation}

The Gegenbauer polynomial of degree one is defined by $G_1^{\beta}(t) = 2\beta t, \; \forall t \in [-1,1].$ As a consequence, using the \textit{addition theorem} (cf.~\citep[Lem.1.2.3 and Thm.1.2.6]{Xu}), the Gram matrix $G^*$
is related to the Gegenbauer polynomial of degree one. More precisely, for any $i,j \in [n]$ it holds 
\begin{equation}
G^*_{i,j} = \frac{1}{2\beta n} G_1^{\beta}(\langle X_i,X_j \rangle) = \frac{1}{n d} \sum_{k=1}^d Y_{1,k}(X_i) Y_{1,k}(X_j).\label{gram2gegen}
\end{equation}
 Denoting for all $k \in [d]$ $v^*_k:=\frac{1}{\sqrt{n}}\left(Y_{1,k}(X_1), \dots , Y_{1,k}(X_n)\right) \in \mathds R^n$, and $V^* = (v_1^*,\dots,v^*_d) \in \mathds{R}^{n \times d}$, Eq.\eqref{gram2gegen} becomes
\begin{equation}\notag G^*:= \frac{1}{d}V^*(V^*)^{\top} .\end{equation}
We will prove that for $n$ large enough there exists a matrix $\widehat{V} \in \mathds{R}^{n\times d}$ where each column is an eigenvector of $\widehat{T}_n$, such that $\widehat{G}:=\frac{1}{d} \widehat{V} \widehat{V}^{\top}$ approximates $G^*$ well, in the sense that the Frobenius norm $\|G^*-\widehat{G}\|_F$ converges to $0$. To choose the $d$ eigenvectors of the matrix $\widehat{T}_n$ that we will use to build the matrix $\widehat{V}$, we need the following spectral gap condition \begin{equation} \Delta^*:= \min_{k \in \mathds{N}, \; k\neq 1} |p^*_1-p^*_k|>0.\label{gap-condition}\end{equation}
This condition will allow us to apply Davis-Kahan type inequalities.

Now, thanks to Theorem \ref{thm:delta2-proba-ope}, we know that the spectrum of the matrix~$\widehat{T}_n$ converges towards the spectrum of the integral operator~$\mathds{T}_W$. Then, based on Eq.\eqref{gram2gegen}, one can naturally think that extracting the $d$ eigenvectors of the matrix~$\widehat{T}_n$ related with the eigenvalues that converge towards $p^*_1$, we can approximate the Gram matrix $G^*$ of the latent positions. Theorem~\ref{gram} proves that the latter intuition is true with high probability under the spectral gap condition~\eqref{gap-condition}. The algorithm HEiC~\cite{AC19} (cf. Section~\ref{sec:reminder_heic} for a presentation) aims at identifying the above mentioned $d$ eigenvectors of the matrix~$\widehat{T}_n$ to build our estimate of the Gram matrix~$G^*$. 

\begin{theorem}
\label{gram}
We consider that \hyperref[assumptionA]{Assumption A} holds, we assume $\Delta^*>0$, and we assume that graphon $W$ has regularity $s>0$. We denote $\widehat{V} \in \mathds{R}^{n \times d}$ the $d$ eigenvectors of the matrix $\widehat{T}_n$ associated with the eigenvalues returned by the algorithm HEiC and we define $\widehat{G}:= \frac{1}{d}\widehat{V}\widehat{V}^{\top}$. Then for $n$ large enough and for some constant $D>0$, it holds with probability at least $1-5/n^2$,
\begin{equation}
\label{hatG}
\|G^*-\widehat{G}\|_F \leq D \left(  \frac{n}{\log^2(n)} \right)^{\frac{-s}{2s+d-1}}.
\end{equation}
\end{theorem}

Based on Theorem~\ref{gram}, we propose a kernel density approach to estimate the latitude function $ f_{\mathcal{L}}$ based on the super-diagonal of the matrix $\widehat G$, namely $\left( \hat r_i:=n \widehat G_{i-1,i} \right)_{i\in \{2,\dots,n\}}$. In the following, we denote $\hat f_{\mathcal{L}}$ this estimator.

\section{Relatively Sparse Regime}
\label{sec:relatively_sparse}
Although we deal so far with the so-called {\it dense} regime (i.e. when the expected number of neighbors of each node scales linearly with $n$), our results may be generalized to the {\it relatively sparse} model connecting nodes $i$ and $j$ with probability $W(X_i,X_j)=\zeta_n \mathbf p(\langle X_i,X_j \rangle)$ where $\zeta_n \in (0,1]$ satisfies 
\\$\underset{n}{\lim \inf} \; \zeta_n n /\log n \geq Z $ for some universal constant $Z>0$. 

\begin{figure}[!ht]
\centering
\includegraphics[width=\linewidth]{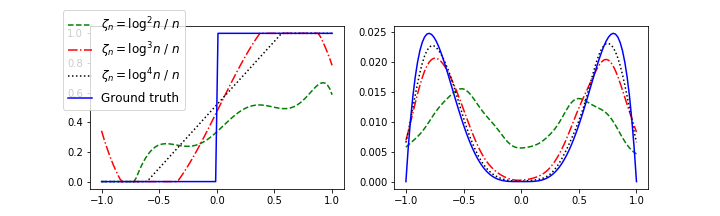}
\vskip -0.1in
\caption[Non-parametric estimation of the envelope and the latitude functions in the relatively sparse regime.]{Results of our algorithms for graph of size $2 000$ with functions $\mathbf p^{(1)}$ and $f_{\mathcal L}^{(1)}$ of Eq.\eqref{eq:simu-env-lat} and sparsity parameter $\zeta_n = \log^k n / n$, $k\in \{2, 3, 4\} $. }
\label{fig:rela-sparse}
\end{figure}

In the relatively sparse model, one can show following the proof of Theorem \ref{thm:delta2-proba-ope} that the resolution should be chosen as $\widehat{R} = \left(\frac{n \zeta_n}{1+\zeta_n \log^2 n}\right)^{\frac{1}{2s+d-1}}$. Specifying that $\lambda^*=(p_0^*,p_1^*,\dots,p^*_1, p^*_2,\dots)$ and $\widehat{T}_n=A/n$, Theorem \ref{thm:delta2-proba-ope} becomes for a graphon with regularity $s>0$ \begin{equation} \notag\mathds{E}\left[\delta_2^2\left(\lambda^*,\frac{\lambda(\widehat{T}_n)}{\zeta_n}\right) \right]=\mathcal{O}\left(\left(  \frac{n \zeta_n}{1+\zeta_n \log^2 n} \right)^{\frac{-2s}{2s+d-1}}\right).\end{equation}
Figure~\ref{fig:rela-sparse} illustrates the estimation of the latitude and the envelope functions in some relatively sparse regimes.

\clearpage
\section{Experiments}
\label{experiments}

In the following, we test our methods using different envelope and latitude functions. Note that a common choice of connection functions in RGGs are the {\it Rayleigh fading} activation functions which take the form\[\mathcal R_{\zeta,\eta,r}(\rho)= \exp\left[-\zeta \rho^{\eta}\right], \quad \zeta>0, \eta>0.\]
Any Rayleigh function $\mathcal R_{\zeta,\eta}$ corresponds to the following envelope function
\[\mathbf{p}_{\zeta,\eta}:t\mapsto \mathcal R_{\zeta,\eta}(2(1-t)),\]
so that it holds
\[\forall x,y\in \mathds S^{d-1}, \quad \mathbf{p}_{\zeta,\eta}( \langle x, y \rangle) =\mathcal R_{\zeta,\eta}(\|x-y\|_2) .\]
Let us also denote for any $\alpha,\beta>0$ $g(\cdot;\alpha,\beta)$ the density of the beta distribution $\mathcal B(\alpha,\beta)$ with parameters $(\alpha,\beta)$.
In this paper, we will study the numerical results of our methods considering the following envelope and latitude functions
\begin{align*}\notag \mathbf{p}^{(1)}:x&\mapsto \mathds{1}_{x \geq 0},  \; && \mathbf{p}^{(2)}\equiv \mathbf{p}_{0.5,1} \\
f_{\mathcal{L}}^{(1)} :r &\mapsto  \left\{
    \begin{array}{ll}
        \frac12 g(1-r;2,2) & \mbox{if } r \geq 0 \\
         \frac12 g(1+r;2,2) & \mbox{otherwise}
    \end{array}
\right. ,\; &&f_{\mathcal{L}}^{(2)}:r\mapsto \frac12 g\left(\frac{1-r}{2};1,3\right) 
\end{align*}
\begin{align}\notag \text{and} \quad & \mathbf{p}^{(3)}\equiv \mathbf{p}_{0.25,3} \\
&f_{\mathcal{L}}^{(3)}:r\mapsto   \frac12 g\left(\frac{1-r}{2};2,2\right).
\label{eq:simu-env-lat} \end{align}

\begin{figure}[!ht]
\centering
\subfloat[Envelope function]{\includegraphics[width = 3in]{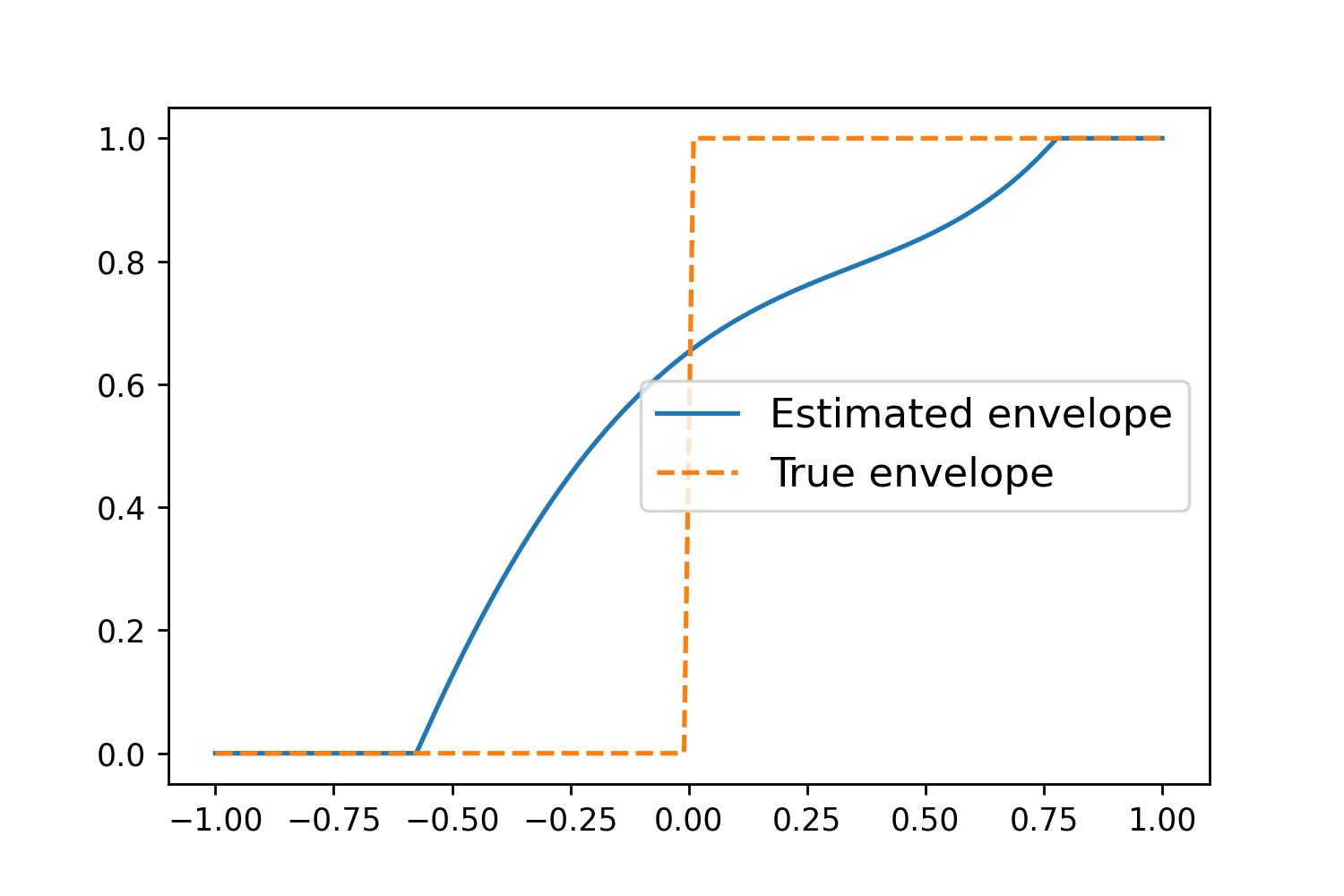}} 
\subfloat[Latitude function]{\includegraphics[width = 3in]{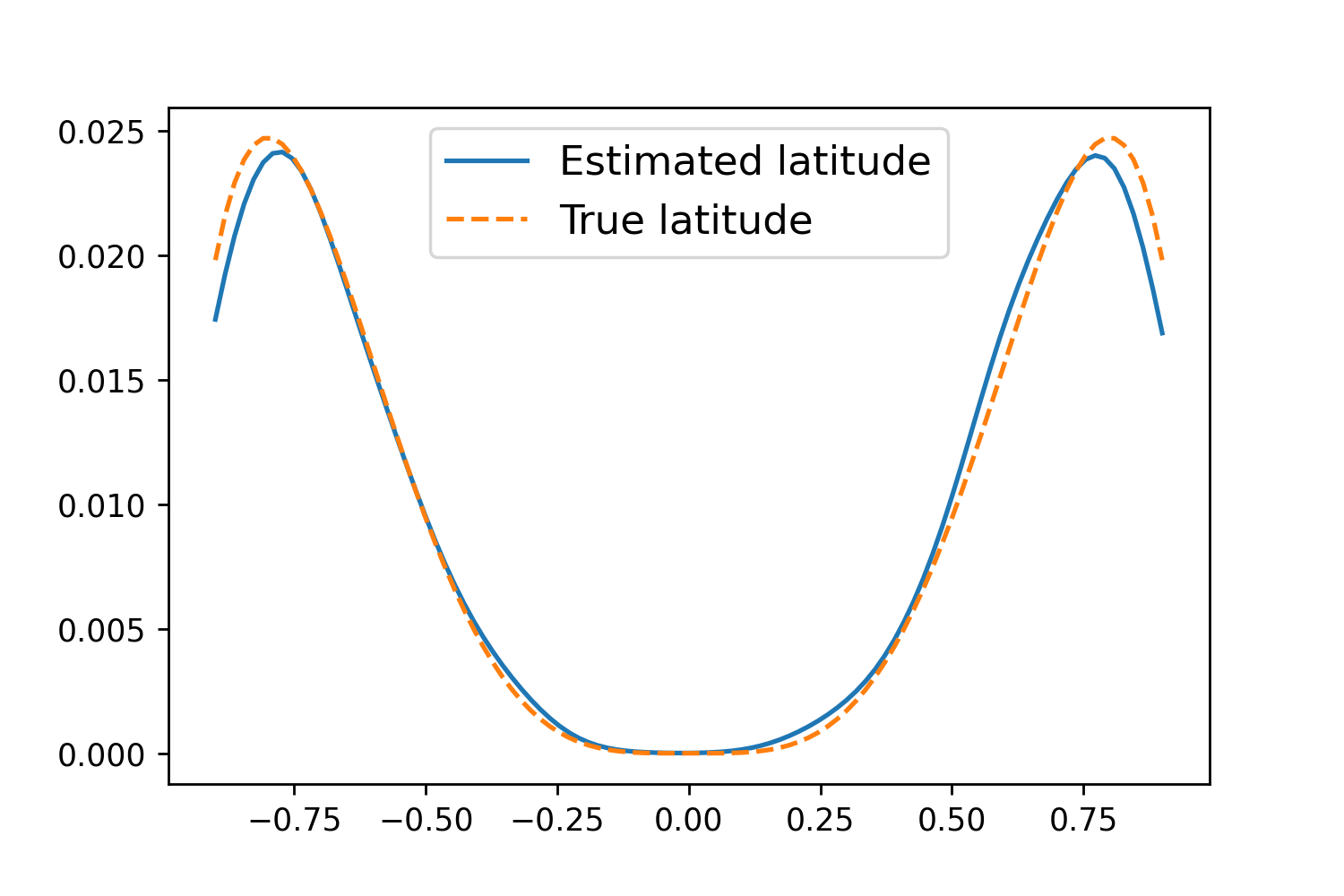}}\\
\subfloat[Eigenvalues envelope]{\includegraphics[width = 3in]{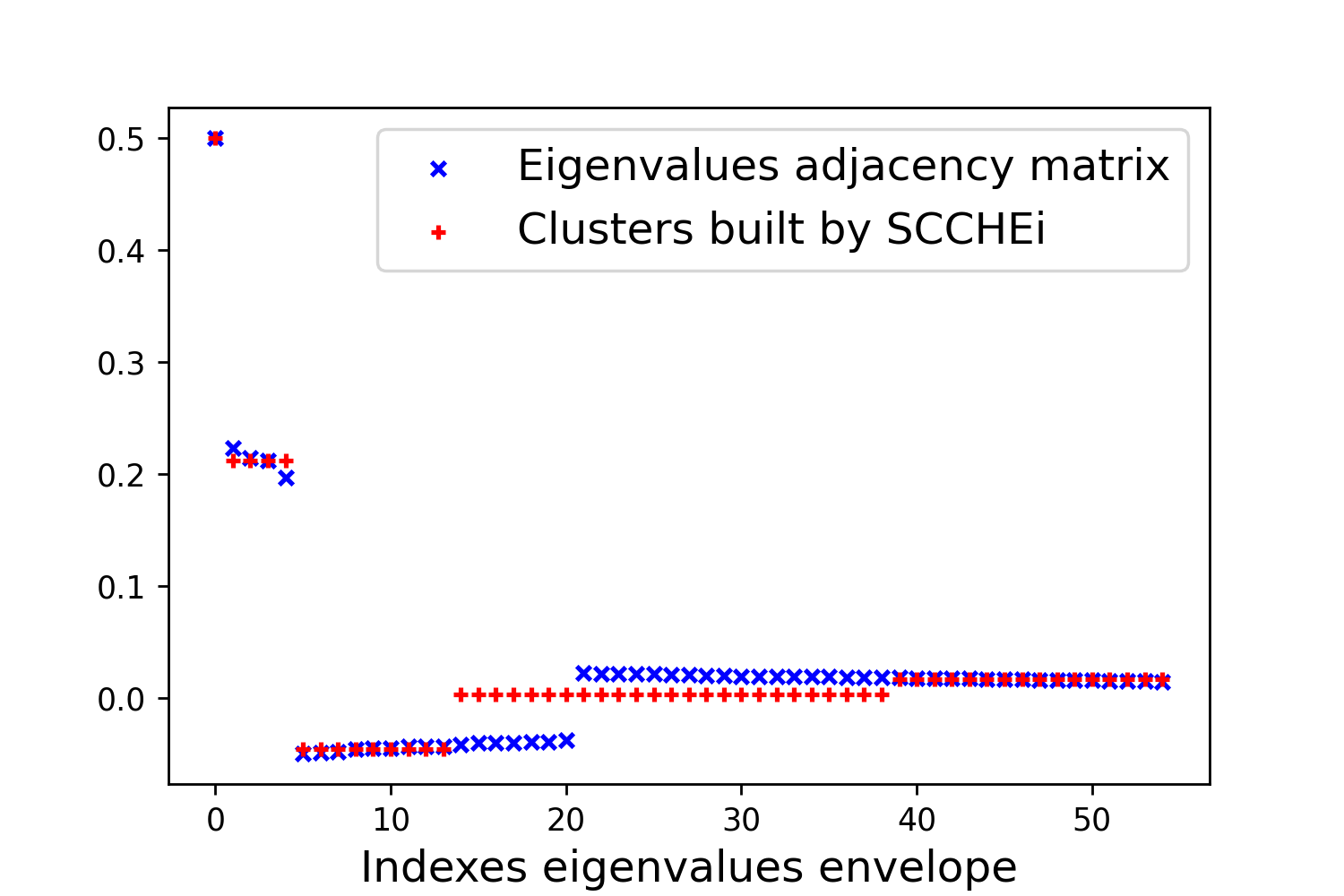}}
\subfloat[$\delta_2$ errors]{\includegraphics[width = 3in]{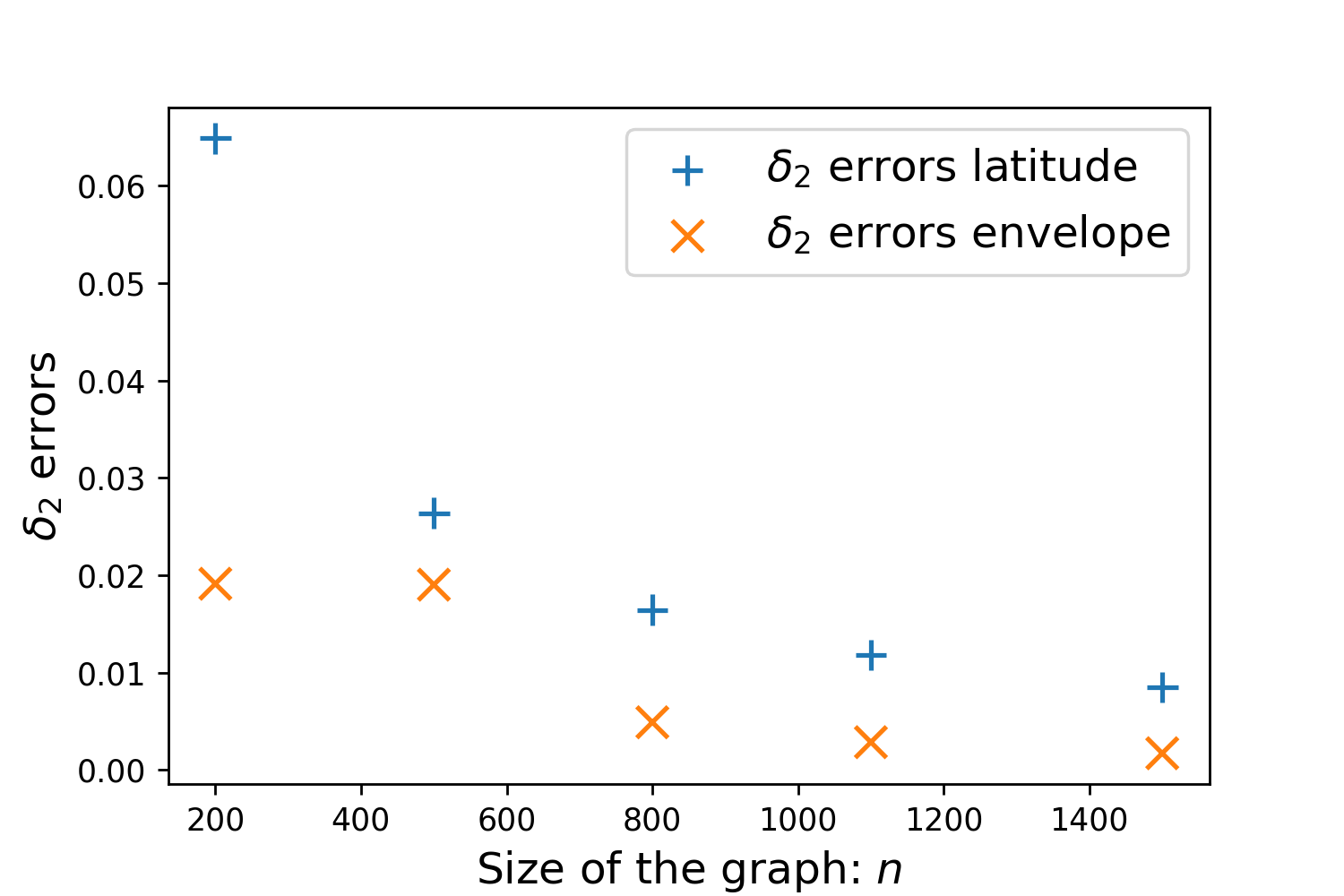}}\\
\subfloat[$L^2$ errors]{\includegraphics[width = 3.1in]{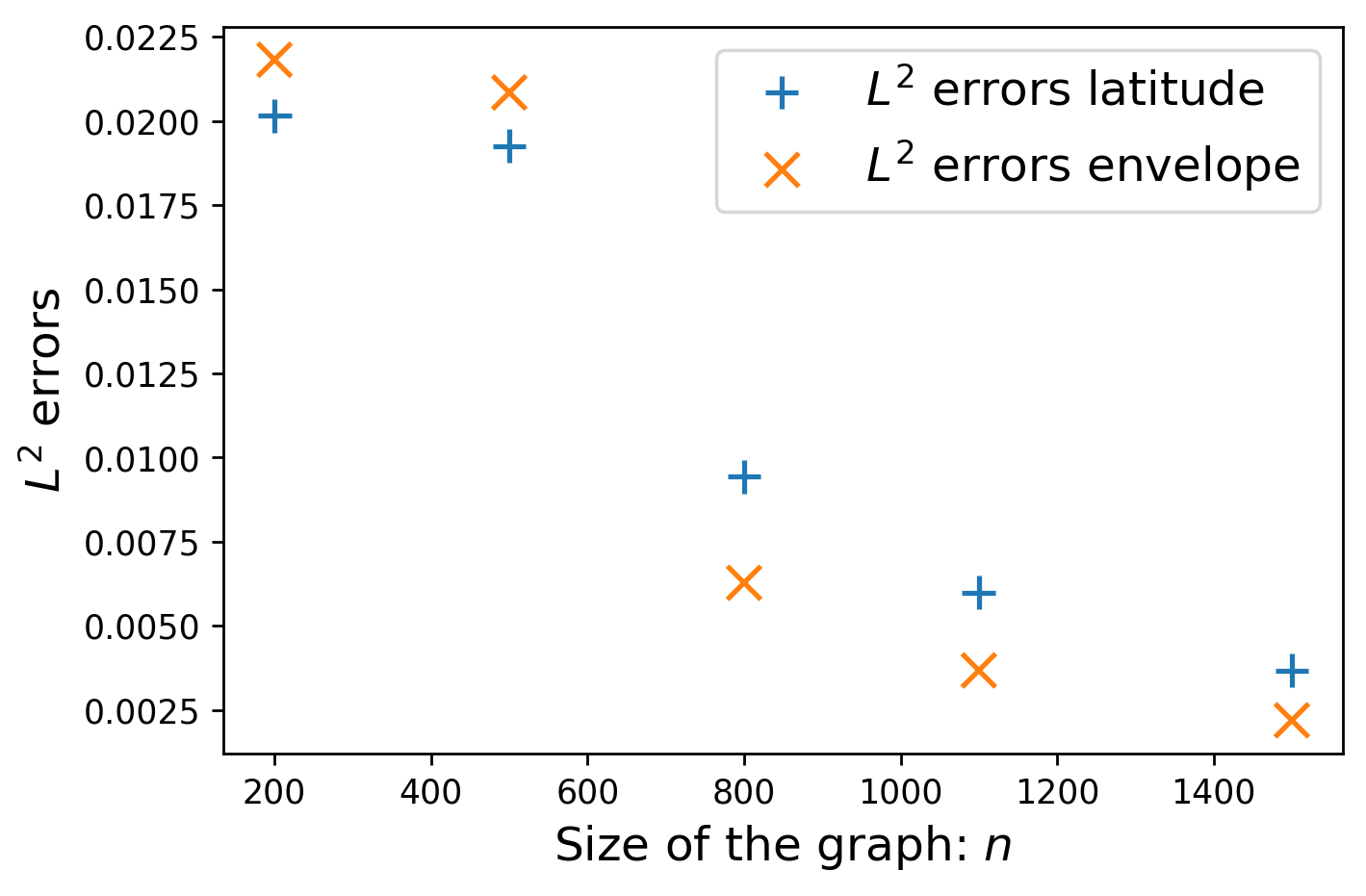}}
\caption[Synthetic presentation of our procedures (1/3).]{Results for $d=4$, the envelope $\mathbf{p}^{(1)}$ and the latitude $f_{ \mathcal L}^{(1)}$ of Eq.\eqref{eq:simu-env-lat}.}
\label{fig:mrgg-expe7}
\end{figure}

\begin{figure}
\centering
\subfloat[Envelope function]{\includegraphics[width = 3in]{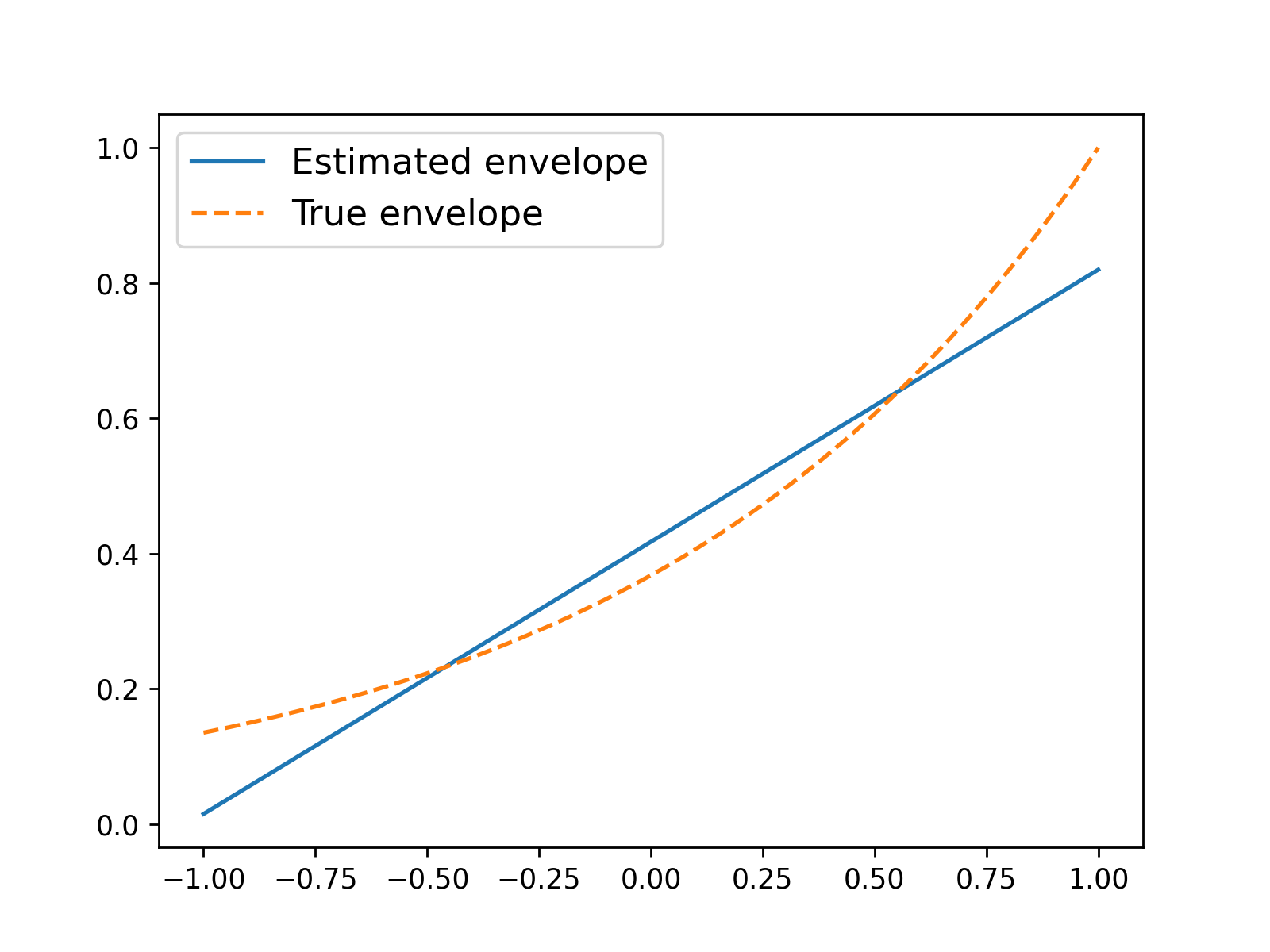}} 
\subfloat[Latitude function]{\includegraphics[width = 3in]{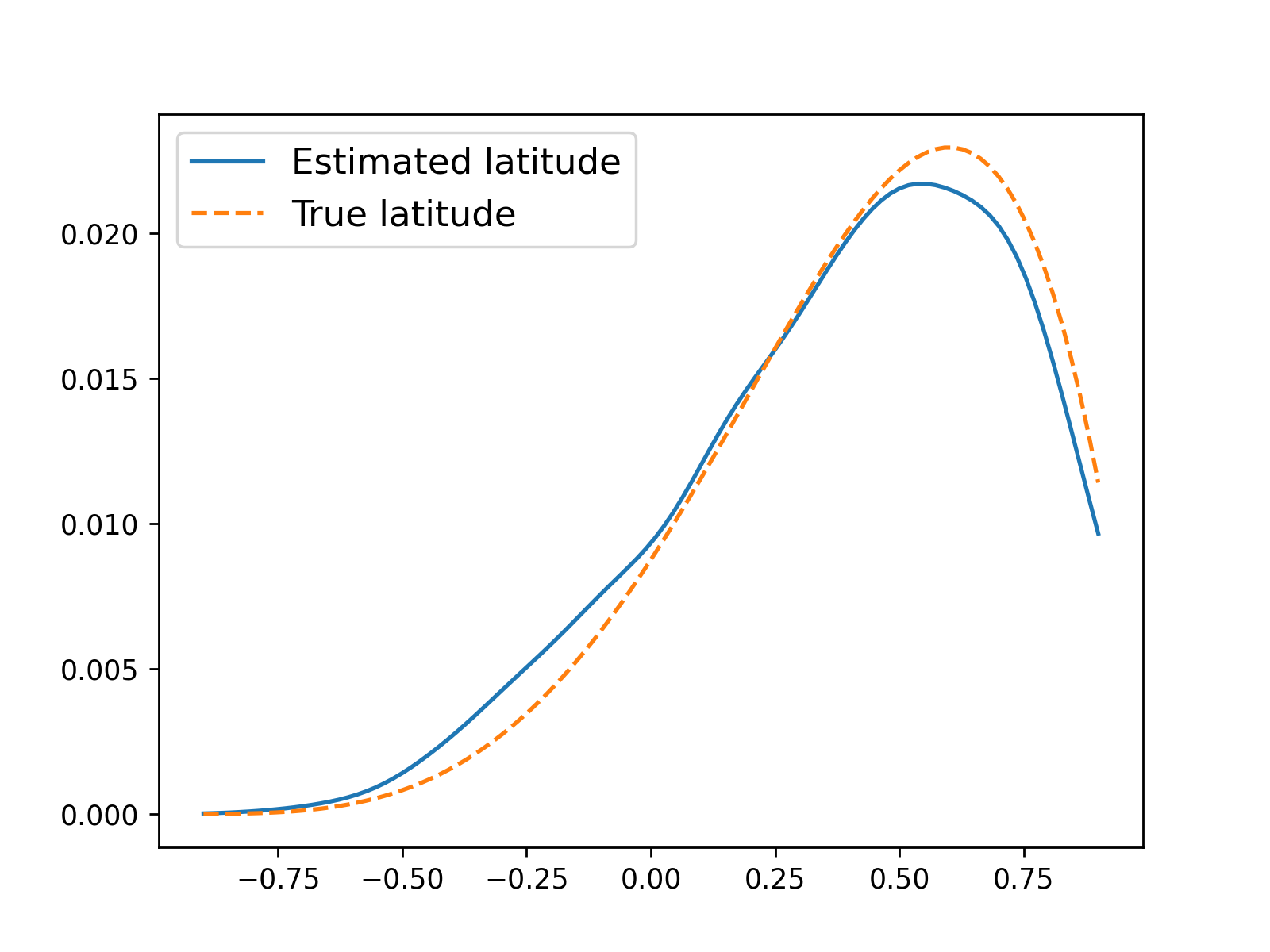}} \\
\subfloat[Eigenvalues envelope]{\includegraphics[width = 2.9in]{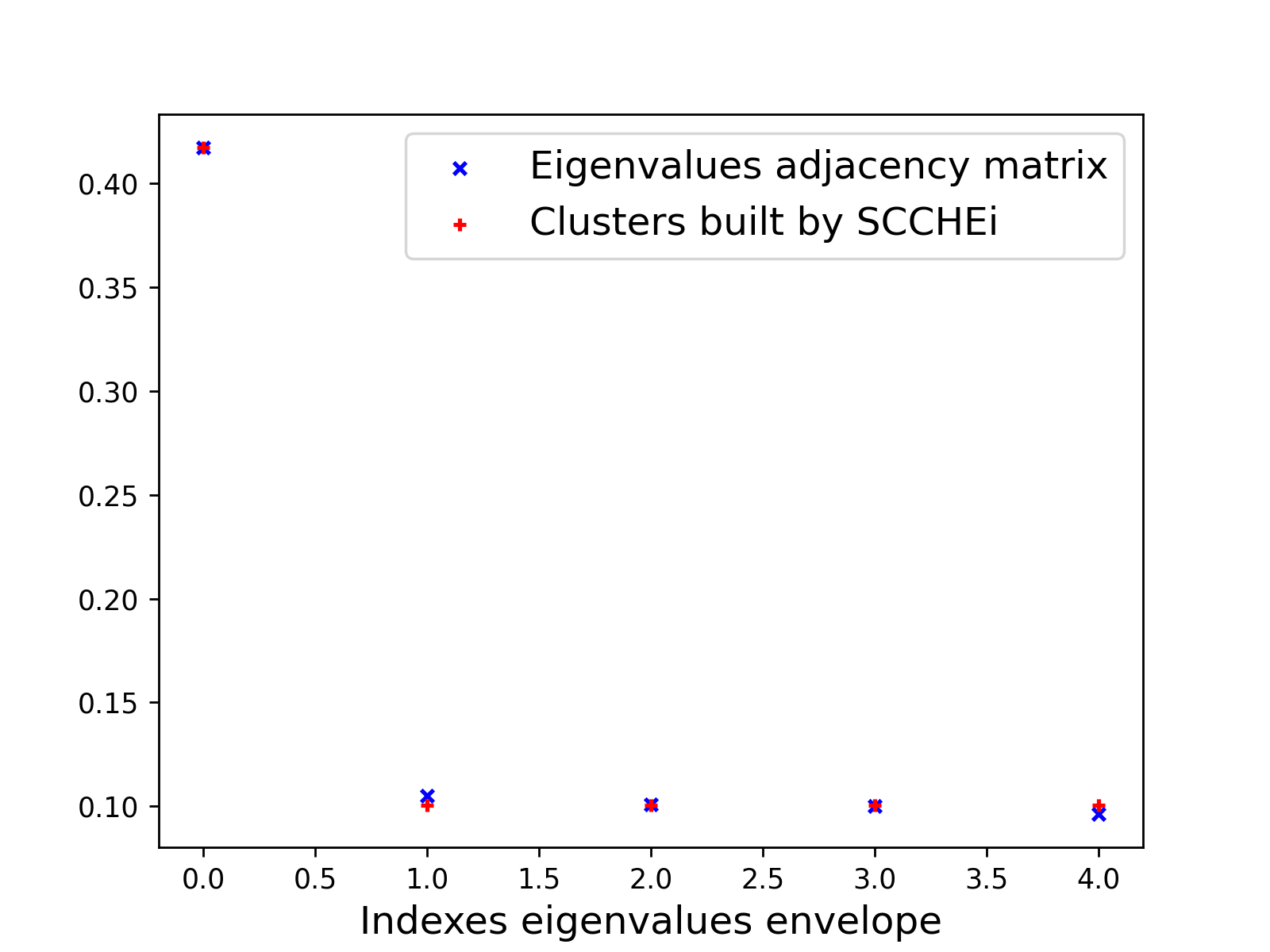}}
\subfloat[$\delta_2$ errors]{\includegraphics[width = 3in]{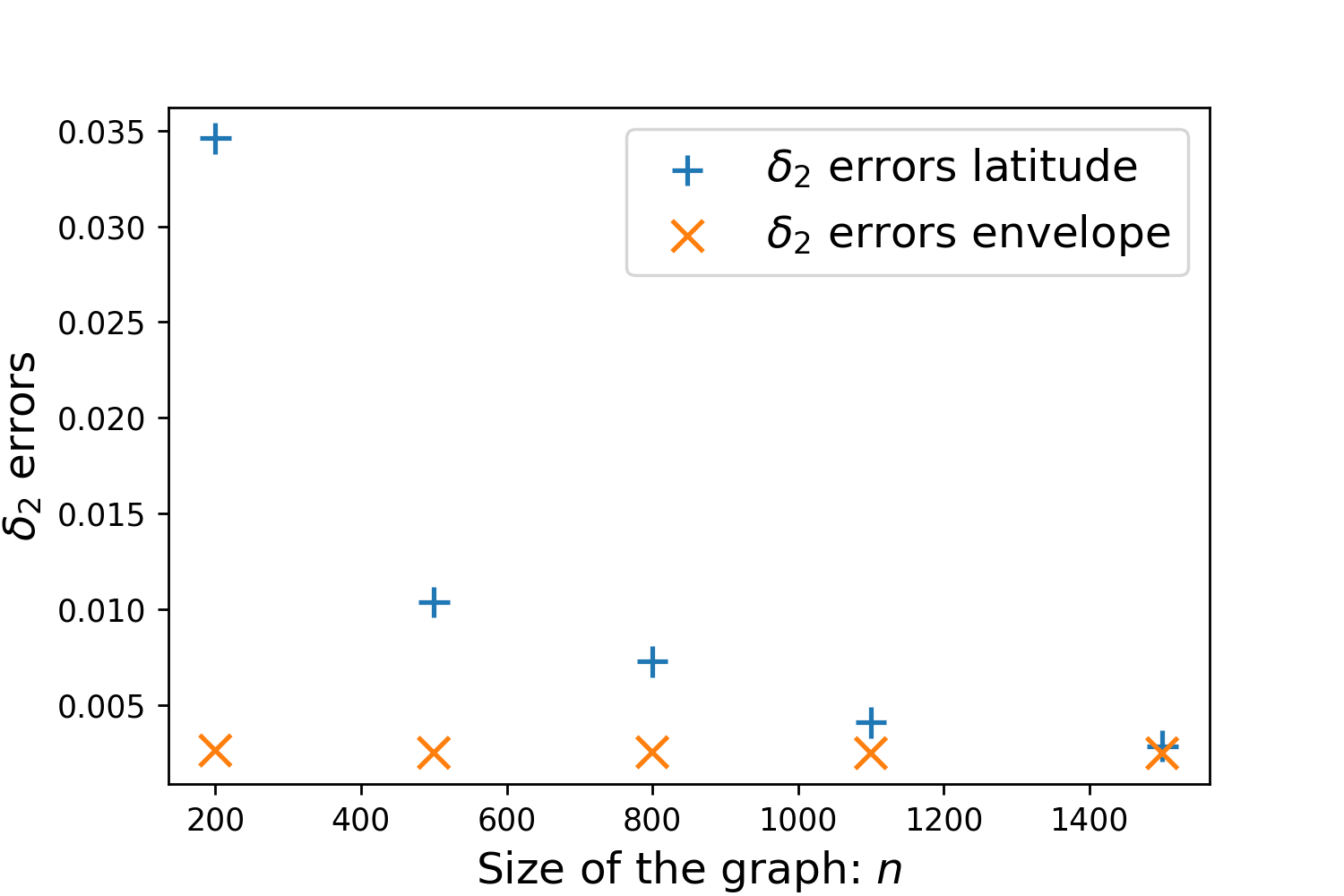}}\\
\subfloat[$L^2$ errors]{\includegraphics[width = 3.1in]{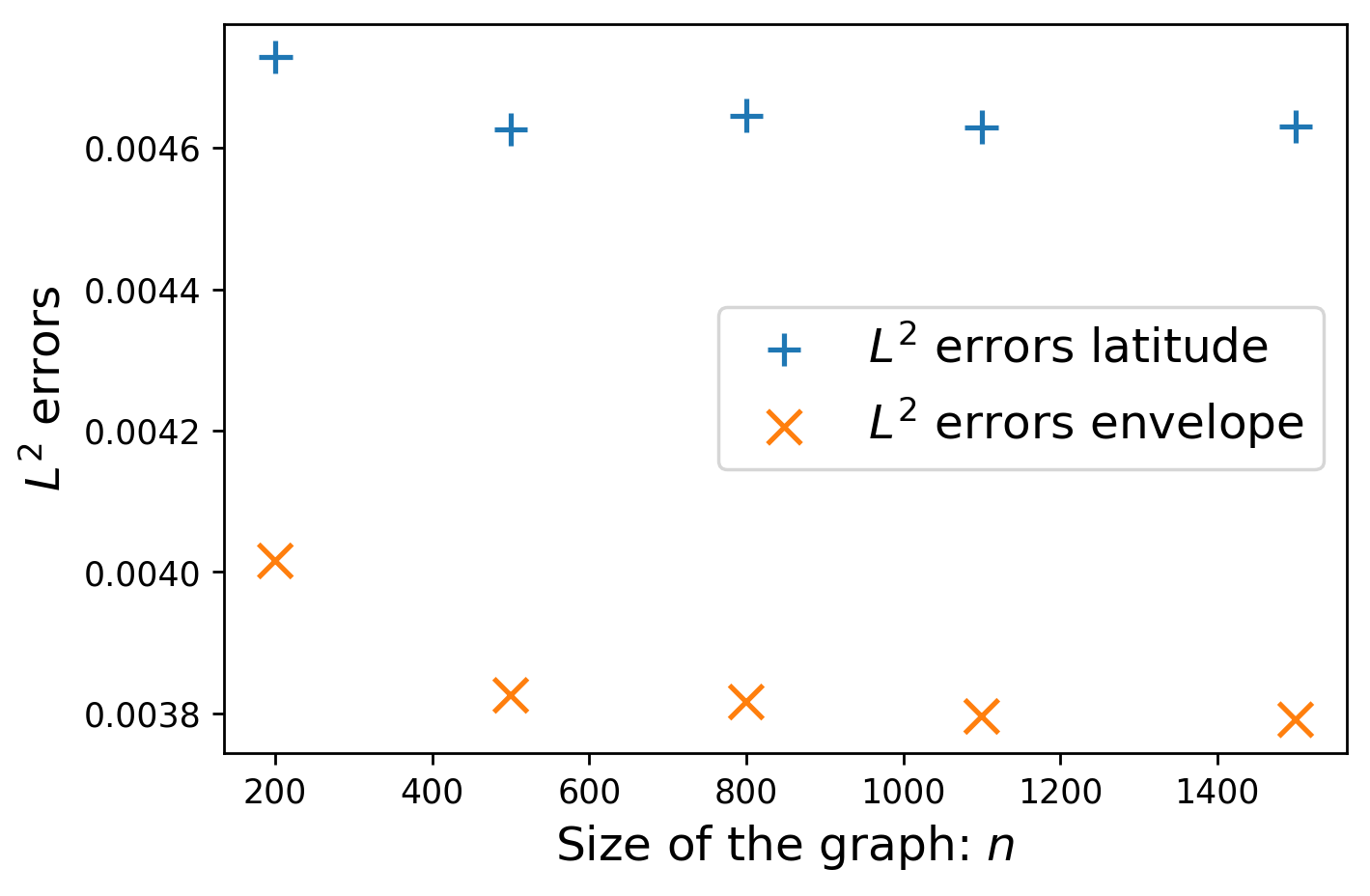}}
\caption[Synthetic presentation of our procedures (2/3).]{Results for $d=4$, the envelope $\mathbf{p}^{(2)}$ and the latitude $f_{ \mathcal L}^{(2)}$ of Eq.\eqref{eq:simu-env-lat}.}
\label{fig:mrgg-expe1}
\end{figure}

\begin{figure}
\centering
\subfloat[Envelope function]{\includegraphics[width = 3in]{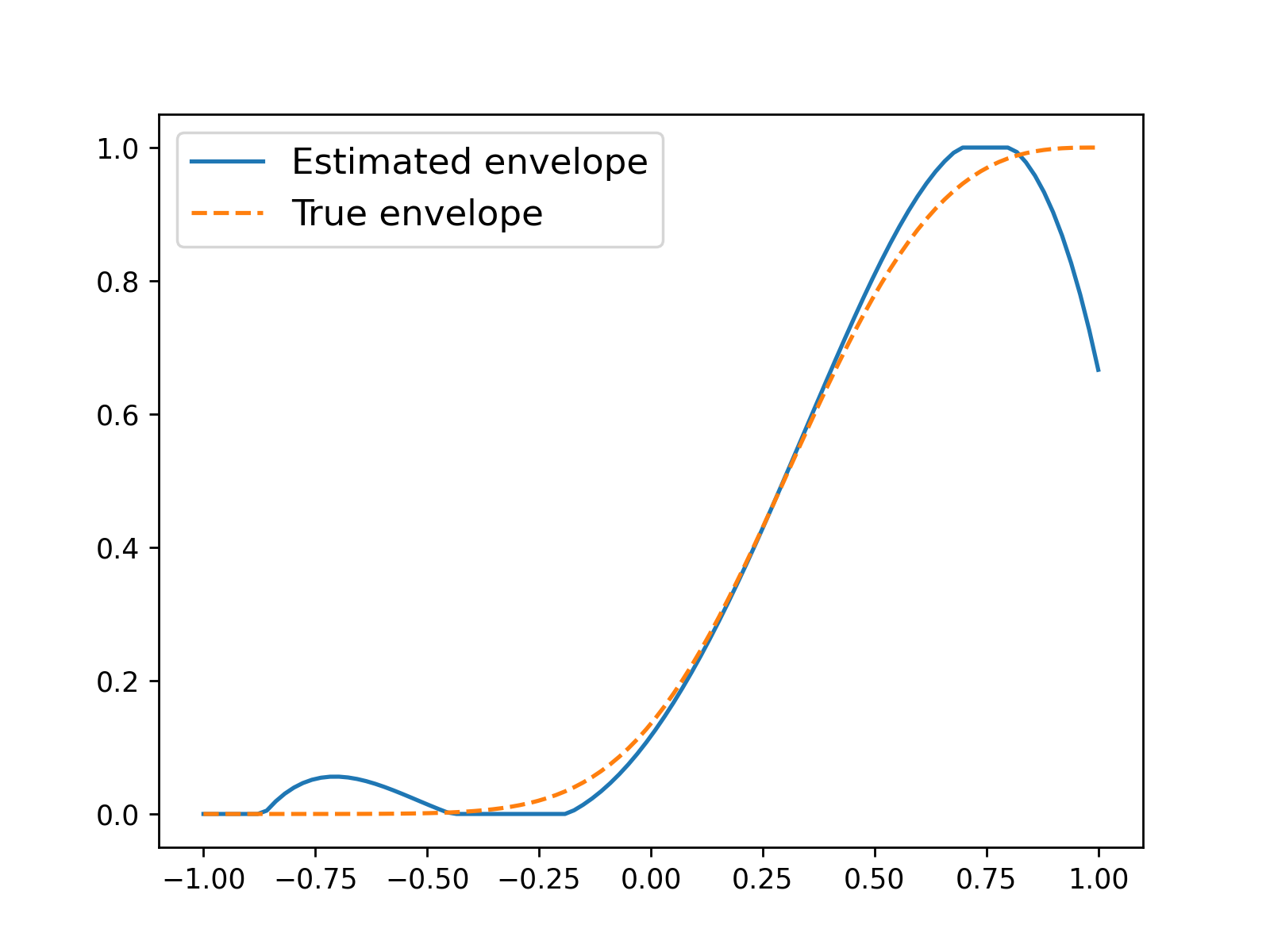}} 
\subfloat[Latitude function]{\includegraphics[width = 3in]{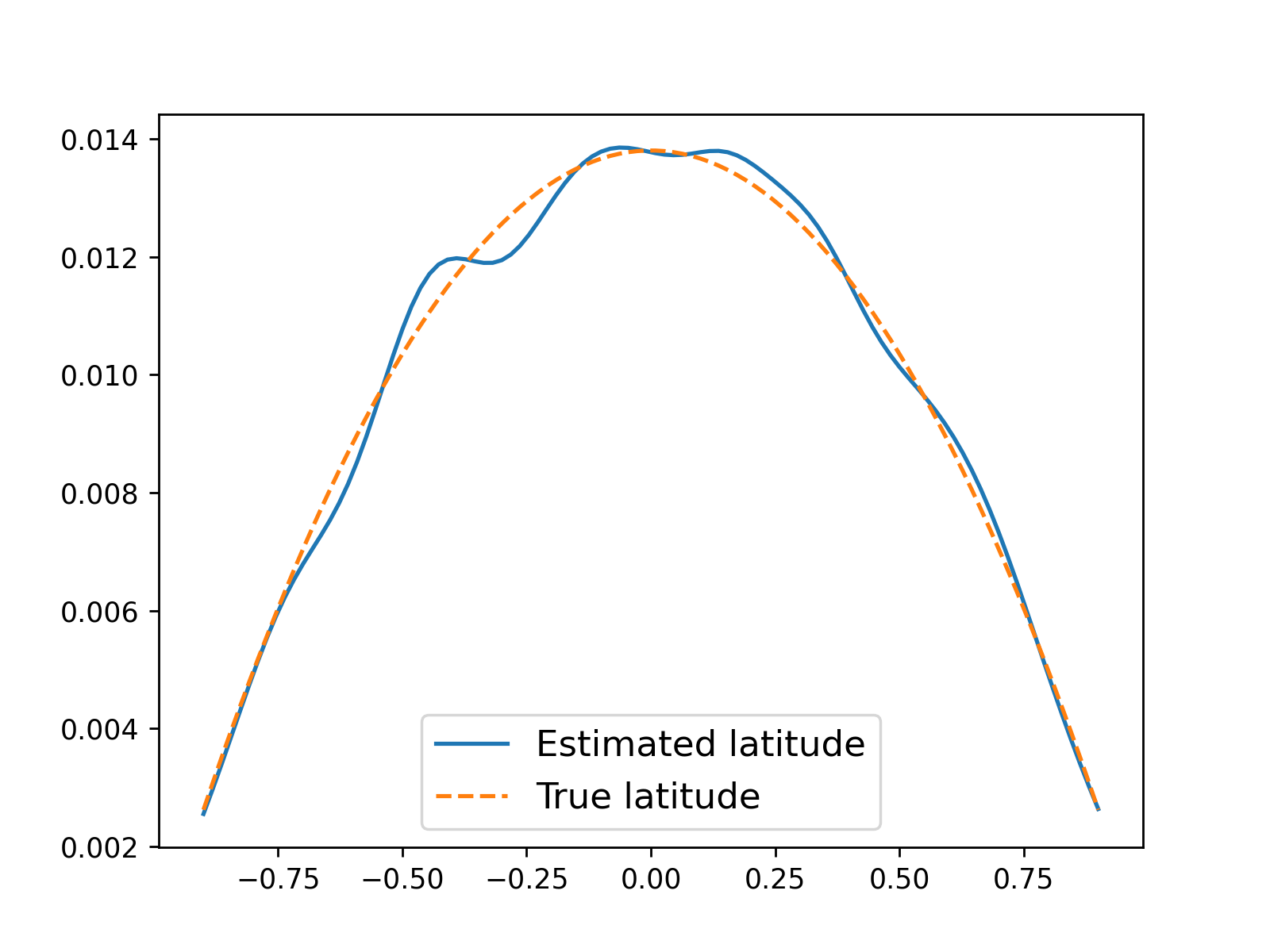}}\\
\subfloat[Eigenvalues envelope]{\includegraphics[width = 2.8in]{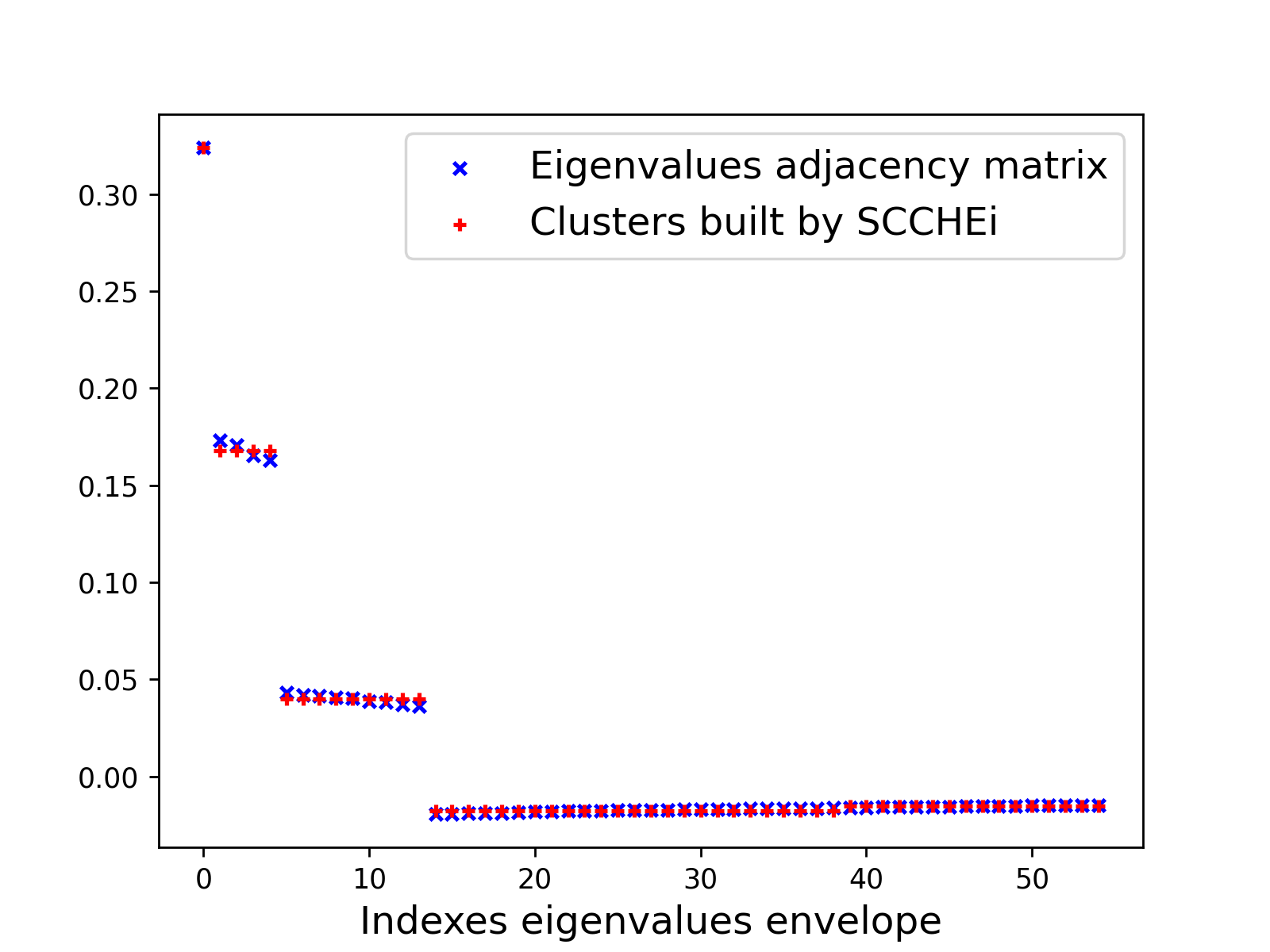}}
\subfloat[$\delta_2$ errors]{\includegraphics[width = 3in]{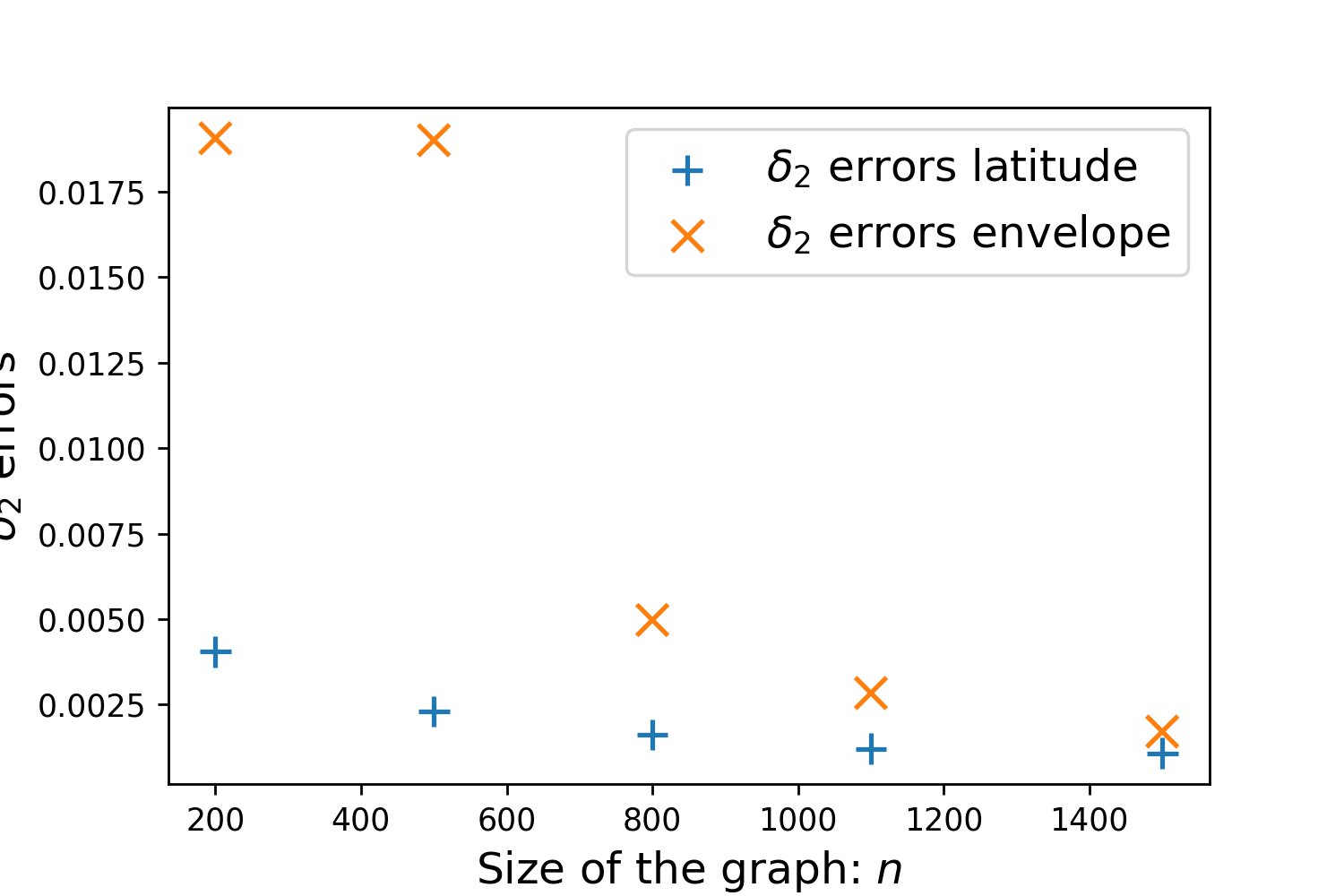}}\\
\subfloat[$L^2$ errors]{\includegraphics[width = 3.1in]{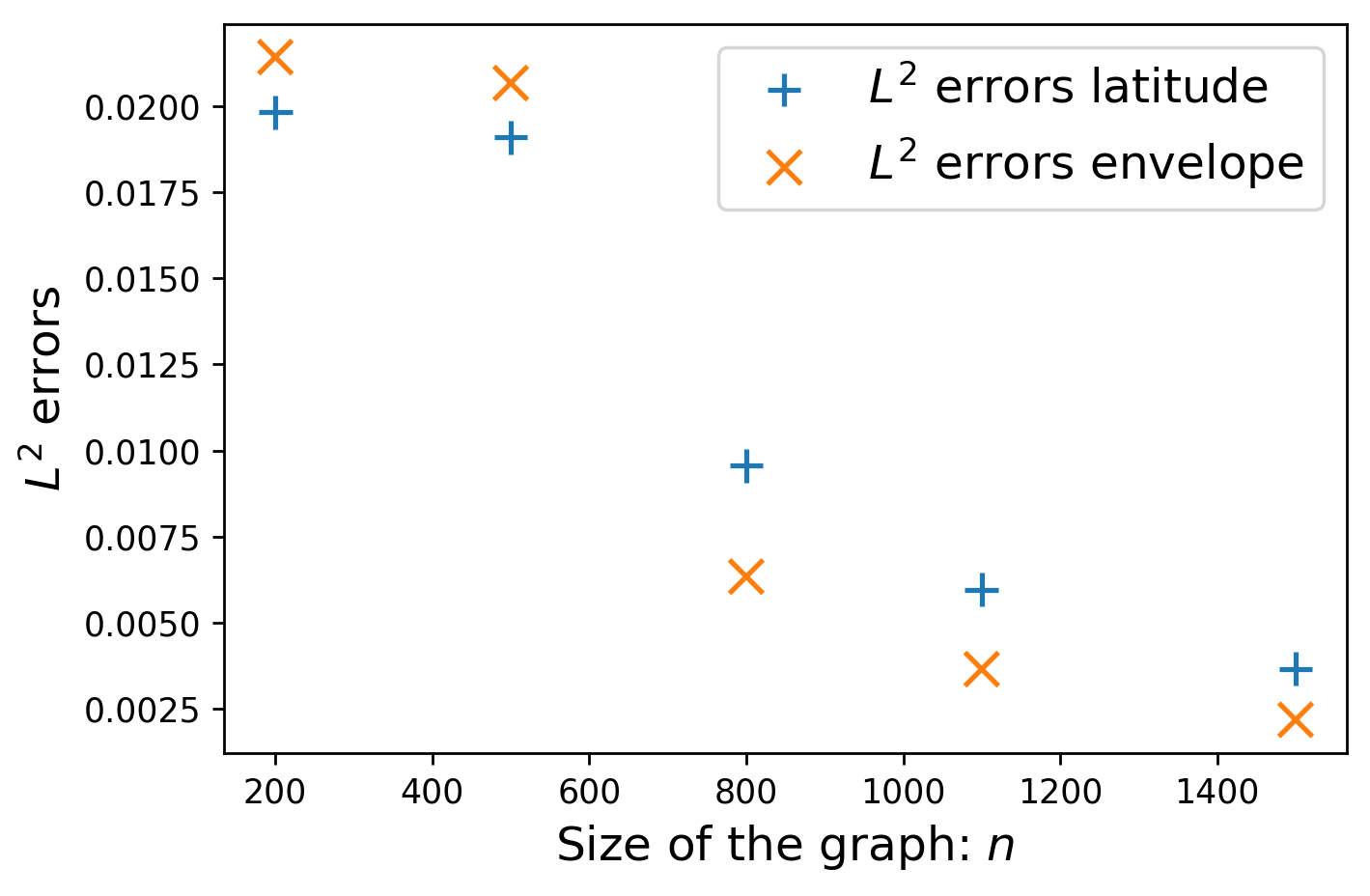}}
\caption[Synthetic presentation of our procedures (3/3).]{Results for $d=4$, the envelope $\mathbf{p}^{(3)}$ and the latitude $f_{ \mathcal L}^{(3)}$ of Eq.\eqref{eq:simu-env-lat}.}\label{fig:mrgg-expe6}
\end{figure}

Note that considering the latitude function $f_{\mathcal L}^{(2)}$ (resp. $f_{\mathcal L}^{(3)}$) is equivalent to consider that one fourth of the Euclidean distance between consecutive latent positions is distributed as $Z \sim \mathcal B(1,3)$ (resp. $Z \sim \mathcal B(2,2)$). With Figures~\ref{fig:mrgg-expe7},~\ref{fig:mrgg-expe1} and~\ref{fig:mrgg-expe6}, we present the results of our experiments for the three different settings described in Eq.\eqref{eq:simu-env-lat}. In each case, we work with a latent dimension $d=4$ and we show:
\begin{enumerate}
\item the estimates of the envelope and latitude functions obtained with our adaptive procedure working the graph of $1 500$ nodes (see Figures $(a)$ and $(b)$).

\smallskip
\item the corresponding clustering obtained by the SCCHEi algorithm for the resolution level $R$ determined by the slope heuristic (see Figures $(c)$).  \vspace{0.1cm} 

Blue crosses represent the $\widetilde{R}$ eigenvalues of $\widehat{T}_n$ with the largest magnitude, which are used to form clusters corresponding to the $R+1$-first spherical harmonic spaces. The red plus are the estimated eigenvalues $(\hat{p}_k)_{0 \leq k \leq R}$ (plotted with multiplicity) defined from the clustering given by our algorithm SCCHEi (see Eq.~\eqref{hatp}). Those results show that SCCHEi achieves a relevant clustering of the eigenvalues of $\widehat{T}_n$ which allows us to recover the envelope function.\smallskip 

\item the errors between the estimated functions and the true ones in $\delta_2$ metric and in $L^2$ norm for different size of graphs (see Figures $(d)$ and $(e)$).
\vspace{0.1cm} 
We notice that a significant decrease of the $\delta_2$ distance between spectra does not necessarily means that the $L^2$ norm between the estimated and the true envelope functions shrinks seriously. We refer in particular to Figures~\ref{fig:mrgg-expe7} and~\ref{fig:mrgg-expe6}. The identifiability issue highlighted in Section~\ref{sec:theo-guarantees} is one of the possible explanations of this phenomenon. Nevertheless, these experiments show that both the $\delta_2$ and $L^2$ errors on our estimate of the envelope or the latitude functions are decreasing as the size of the graph is getting larger. Let us also recall that Theorem~\ref{thm:poly} ensures that the $L^2$ error on our estimate of the envelope function goes to zero as $n$ grows when $\mathbf{p}$ has a finite number of non zeros eigenvalues that are all distinct. 
\end{enumerate}


  \clearpage

 \section{Applications}
 
 \label{sec:applications} 
 
 In this section, we apply the MRGG model to link prediction and hypothesis testing in order to demonstrate the usefulness of our approach as well as the estimation procedure.

 \subsection{Markovian Dynamic Testing}
 \label{subsec:testing}
 
As a first application of our model, we propose a hypothesis test to statistically distinguish between an independent sampling the latent positions and a Markovian dynamic. The null is then set to $\mathbb H_0: $ \textit{nodes are independent and uniformly distributed on the sphere} (i.e., {\it no Markovian dynamic}). Our test is based on estimate $\hat f_{\mathcal L}$ of latitude and thus the null can be rephrased as $\mathbb H_0:\ f_{\mathcal L}=f_{\mathcal L}^0$ where $f_{\mathcal L}^0$ is the latitude of uniform law, dynamic is then i.i.d. dynamic.

\begin{figure}[!ht]
\begin{minipage}[c]{0.52\linewidth}
\centering
\includegraphics[width=70mm]{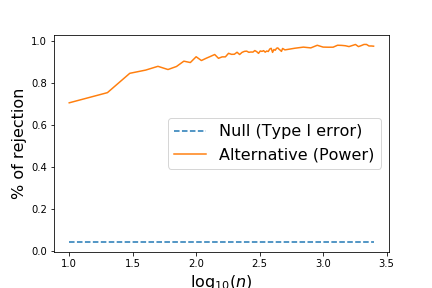} \caption[Markovian dynamic testing in the MRGG model.]{Hypothesis testing.}
\label{fig:hypo-testing}
\end{minipage}
\begin{minipage}[c]{0.45\linewidth}
Figure~\ref{fig:hypo-testing} shows the power of a hypothesis test with level $5\%$ (Type I error). One can use any {\it black-box goodness-of-fit test} comparing $\hat f_{\mathcal L}$ to $f_{\mathcal L}^0$, and we choose $\chi^2$-test discretizing $(-1,1)$ in $70$ regular intervals. 
Rejection region is calibrated (i.e., threshold of the $\chi^2$-test here) by {\it Monte Carlo simulations under the null}. It allows us to control Type I error as depicted by dotted blue line. We choose alternative given by Heaviside envelope $\mathbf{p}^{(1)}$ and latitude $f_{\mathcal L}^{(1)}$ of Eq.\eqref{eq:simu-env-lat}. We run our algorithm to estimate latitude from which we sample a batch to compute the $\chi^2$-test statistic. We see that for graphs of size larger than $1,000$, the rejection rate is almost $1$ under the alternative (Type II error is almost zero), the test is very powerful.
\end{minipage}
\end{figure}

\subsection{Link Prediction}

Suppose that we observe a graph with $n$ nodes. Link prediction is the task that consists in estimating the probability of connection between a given node of the graph and the upcoming node. 

\subsubsection{Bayes Link Prediction}

We propose to show the usefulness of our model solving a link prediction problem. Let us recall that we do not estimate the latent positions but only the {\it pairwise distances} (embedding task is not necessary for our purpose). Denoting by $ \mathrm{proj}_{X_n^{\perp}}(\cdot)$ the orthogonal projection onto the orthogonal complement of $\mathrm{Span(X_n)}$, the decomposition of $\langle X_i,X_{n+1} \rangle$ defined by
\begin{align}&\langle X_i,X_n \rangle \langle X_n , X_{n+1} \rangle\notag \\
&+ \sqrt{1-\langle X_n , X_{n+1} \rangle^2}  
\label{eq:decompo} \sqrt{1-\langle X_i,X_n \rangle^2} \langle \frac{\mathrm{proj}_{X_n^{\perp}}(X_i)}{\|\mathrm{proj}_{X_n^{\perp}}(X_i)\|_2}, Y_{n+1}\rangle ,\end{align} shows that latent distances are enough for link prediction. Indeed, it can be achieved using a {\it forward step} on our Markovian dynamic, giving the posterior probability (cf. Definition~\ref{def:posterior-proba}) $\eta_i(\mathbf D_{1:n})$ defined by
{\small\begin{equation}\label{eq:link} \underset{[-1,1]^2}{\int} \mathbf{p}\left(   \langle X_i,X_n \rangle r + \sqrt{1-r^2} \sqrt{1-\langle X_i,X_n \rangle^2}u\right) f_{\mathcal{L}}(r) w_{\frac{d-3}{2}}(u)\frac{\Gamma(\frac{d-1}{2})}{\Gamma(\frac{d-2}{2})\sqrt \pi}drdu,\end{equation}}
where $w_{\frac{d-3}{2}}(u):=(1-u^2)^{\frac{d-3}{2}-\frac12}$ and where $\Gamma:a \in ]0,+\infty[\mapsto \int_0^{+\infty} t^{a-1}e^{-t}dt$.

 \begin{definition} \label{def:posterior-proba} (Posterior probability function) \\
The posterior probability function $\eta$ is defined for any latent pairwise distances $\mathbf D_{1:n} = (\langle X_i,X_j\rangle )_{1\leq i,j\leq n} \in [-1,1]^{n\times n}$  by
\[\forall i \in [n], \quad  \eta_i(\mathbf D_{1:n}) = \mathds P\left( A_{i,n+1}=1 \;|\; \mathbf D_{1:n}  \right) ,\]
where $A_{i,n+1} \sim  \mathrm{Ber}\left( \mathbf p(\langle X_i,X_{n+1} \rangle) \right)$ is a random variable that equals $1$ if there is an edge between nodes $i$ and $n+1$, and is zero otherwise. 
\end{definition}

We consider a classifier $g$ (cf. Definition~\ref{def:classifier}) and an algorithm that, given some latent pairwise distances $\mathbf D_{1:n}$, estimates $A_{i,n+1}$ by putting an edge between nodes $X_i$ and $X_{n+1}$ if $g_i(\mathbf D_{1:n})$ is $1$. 

\begin{definition} \label{def:classifier}
A \textit{classifier} is a function which associates to any pairwise distances $\mathbf D_{1:n} =(\langle X_i,X_j\rangle)_{1\leq i,j\leq n} $, a label $\left( g_{i}(\mathbf D_{1:n}) \right)_{i \in [n]}  \in \{0,1\}^n$.
\end{definition}

The risk of this algorithm is as in {\it binary classification}, {\small\begin{align}&\mathcal R(g, \mathbf D_{1:n}):= \frac{1}{n} \sum_{i=1}^n \mathds P\left(  g_i(\mathbf D_{1:n}) \neq A_{i,n+1}\; | \; \mathbf D_{1:n}\right)\notag\\
&=\frac{1}{n} \sum_{i=1}^n  \left\{(1-\eta_i(\mathbf D_{1:n})) \mathds 1_{g_i(\mathbf D_{1:n})=1} + \eta_i(\mathbf D_{1:n}) \mathds 1_{g_i(\mathbf D_{1:n})=0} \right\}  , \label{BLP:risk}
\end{align}}where we used the independence between $A_{i,n+1}$ and $ g_{i}(\mathbf D_{1:n}) $ conditionally on $ \varkappa(\mathbf D_{1:n}) $.  Pushing further this analogy, we can define the classification error of some classifier $g$ by $L(g) = \mathds E \left[\mathcal R(g, \mathbf D_{1:n})\right]$. Proposition \ref{prop:bayes-optimal} shows that the Bayes estimator - introduced in Definition \ref{def:bayes-estimator} - is optimal for the risk defined in Eq.\eqref{BLP:risk}. 

 \begin{definition} \label{def:bayes-estimator} (Bayes estimator) \\
We keep the notations of Definition~\ref{def:posterior-proba}.
The Bayes estimator $g^*$ of $\left( A_{i,n+1}\right)_{1\leq i \leq n}$ is defined by
\[ \forall i \in [n], \quad g^*_i(\mathbf D_{1:n}) = \left\{
    \begin{array}{ll}
        1 & \mbox{if } \eta_i(\mathbf D_{1:n}) \geq \frac12 \\
        0 & \mbox{otherwise.}
    \end{array}
\right.
 \] 
\end{definition}

\begin{proposition} \label{prop:bayes-optimal} (Optimality of the Bayes classifier for the risk $\mathcal R$)\\
We keep the notations of Definitions~\ref{def:posterior-proba} and \ref{def:bayes-estimator}. For any classifier $g$, it holds for all $i \in [n]$,
\begin{align*}&\mathds P\left( g_i(\mathbf D_{1:n}) \neq A_{i,n+1}  \; |\; \mathbf D_{1:n}\right)- \mathds P\left( g_i^*(\mathbf D_{1:n} ) \neq A_{i,n+1}  \; |\; \mathbf D_{1:n}\right)\\
&= 2\left| \eta_i(\mathbf D_{1:n})-\frac12\right| \times \mathds E\left\{ \mathds{1}_{g_i(\mathbf D_{1:n}) \neq g^*_i(\mathbf D_{1:n})}\;|\; \mathbf D_{1:n} \right\},\end{align*}
which immediately implies that \[\mathcal R(g,\mathbf D_{1:n}) \geq  \mathcal R(g^*,\mathbf D_{1:n}) \text{  and therefore  } L(g) \geq L(g^*).\] 
\end{proposition}

 \subsubsection{Heuristic for Link Prediction}
 
 \label{link-pred}

One natural method to approximate the Bayes classifier from the previous section is to use the {\it plug-in approach}. This leads to the MRGG classifier introduced in Definition~\ref{def:mrgg-classifier}.  

\begin{definition} (The MRGG classifier)\\\label{def:mrgg-classifier}For any $n$ and any $i \in [n]$, we define $\hat \eta_i(\mathbf D_{1:n})$ as 
{\small \begin{align}\label{eq:link-esti}\int \widehat {\mathbf{p}}\left(  \widehat r_{i,n} r + \sqrt{1-r^2} \sqrt{1-\widehat r_{i,n}^2}u\right) \hat f_{\mathcal{L}}(r) w_{\frac{d-3}{2}}(u)\frac{\Gamma(\frac{d-1}{2})}{\Gamma(\frac{d-2}{2})\sqrt \pi}drdu,\end{align}}where $\widehat {\mathbf{p}}$ and $\hat f_{\mathcal L}$ denote respectively the estimate of the envelope function and the latitude function with our method and where $\widehat r :=n\widehat G $. The MRGG classifier is defined by 
\[ \forall i \in [n], \quad g^{MRGG}_i(\mathbf D_{1:n}) = \left\{
    \begin{array}{ll}
        1 & \mbox{if } \hat \eta_i(\mathbf D_{1:n}) \geq \frac12 \\
        0 & \mbox{otherwise.}
    \end{array}
\right.
 \] 
\end{definition}

\begin{figure}
\centering
\subfloat[Envelope $\mathbf{p}^{(1)}$, \; Latitude $f_{\mathcal L}^{(1)}$]{\includegraphics[width = 2.9in]{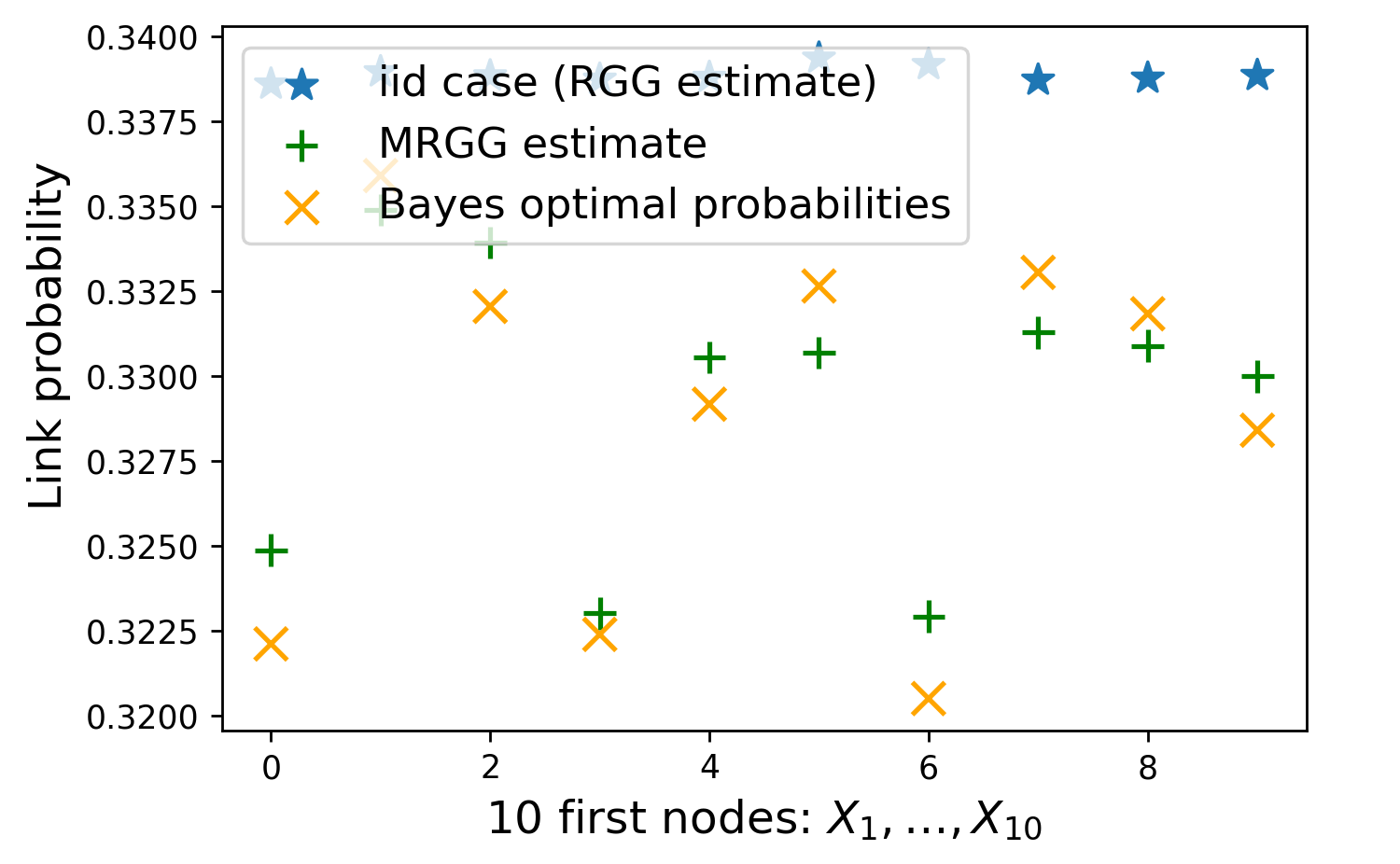}}
\subfloat[Envelope $\mathbf{p}^{(1)}$, \; Latitude $f_{\mathcal L}^{(1)}$]{\includegraphics[width = 2.9in]{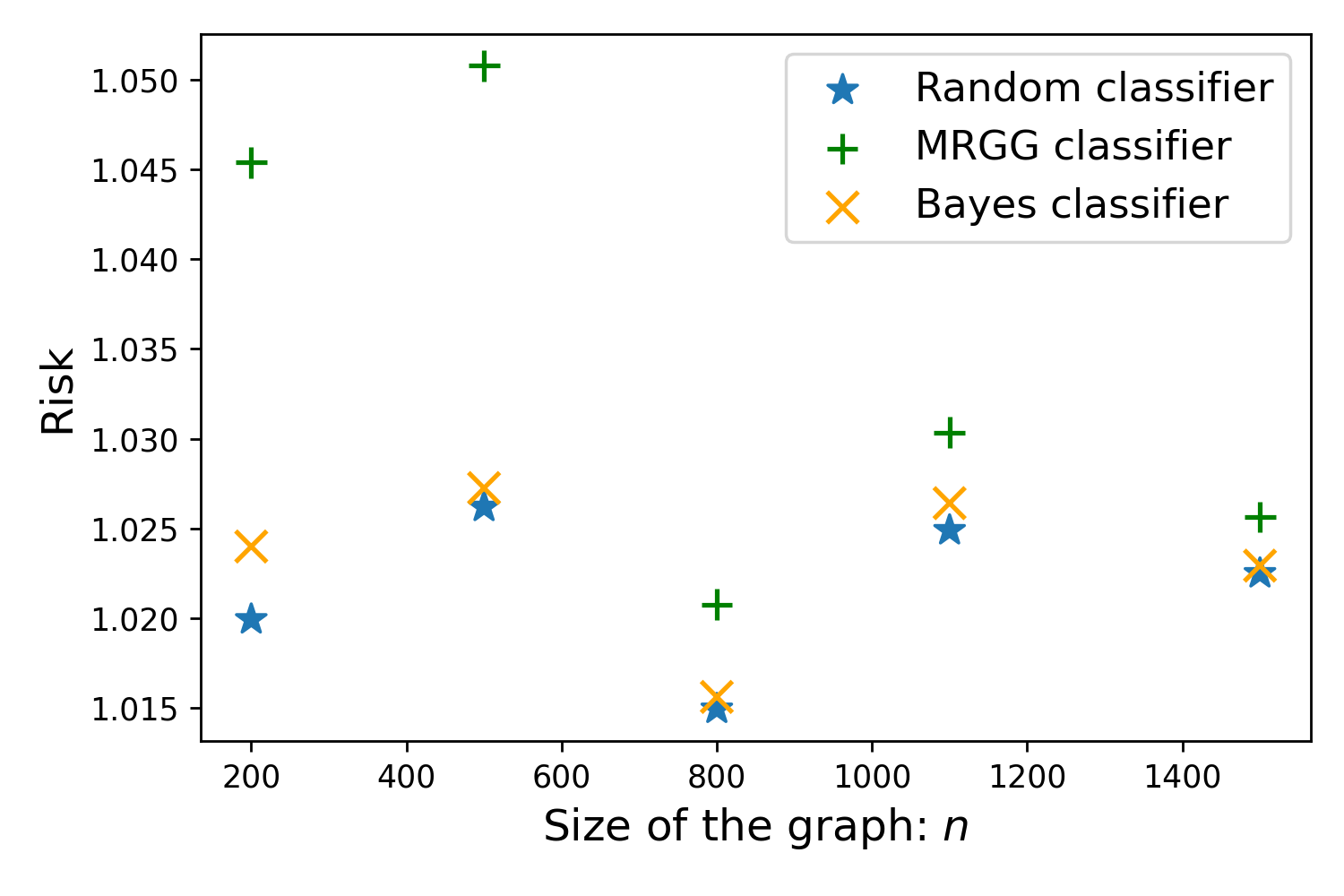} \vspace{-10pt}} \\
\subfloat[Envelope $\mathbf{p}^{(2)}$, \; Latitude $f_{\mathcal L}^{(2)}$]{\includegraphics[width = 2.9in]{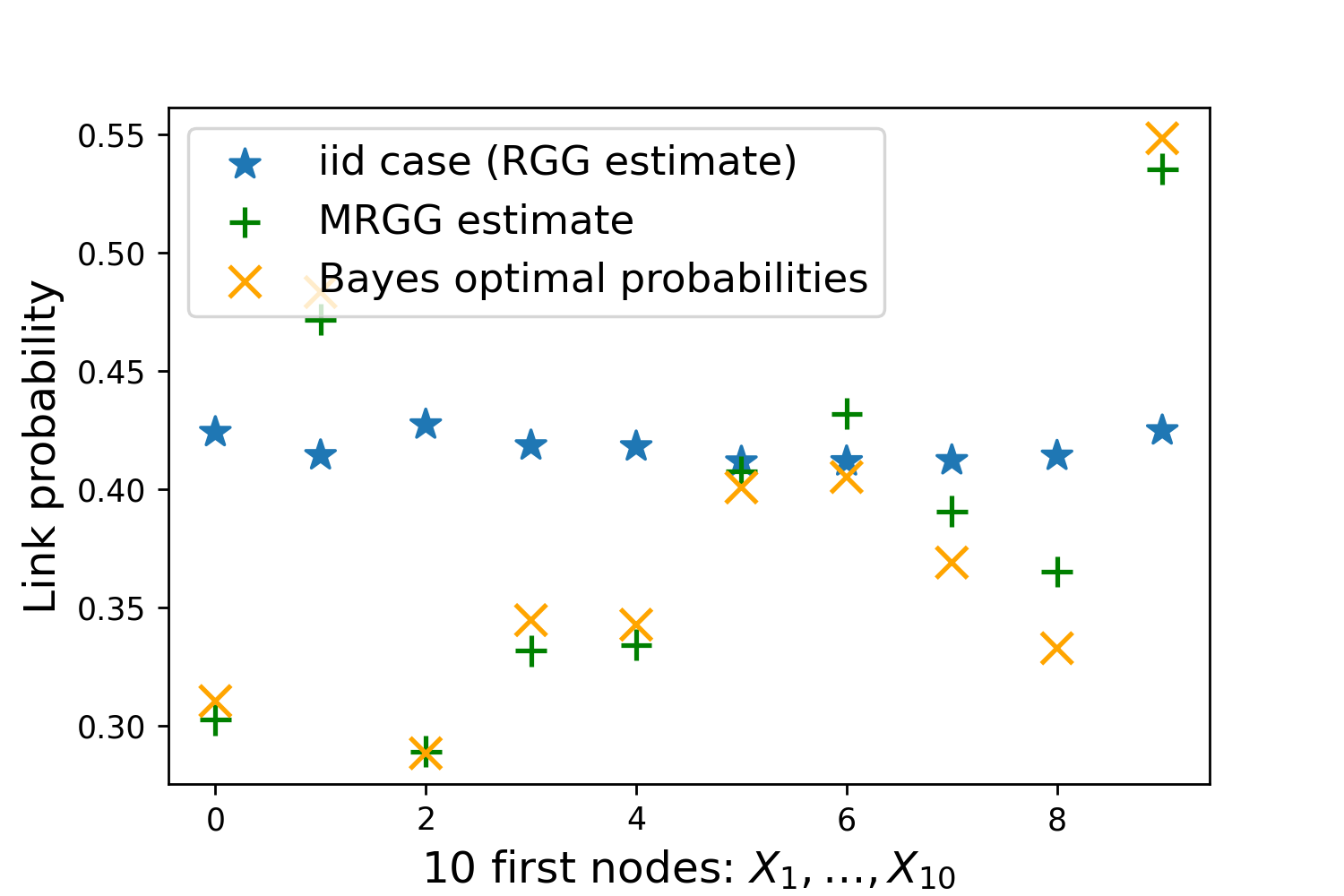}}
\subfloat[Envelope $\mathbf{p}^{(2)}$, \; Latitude $f_{\mathcal L}^{(2)}$]{\includegraphics[width = 2.9in]{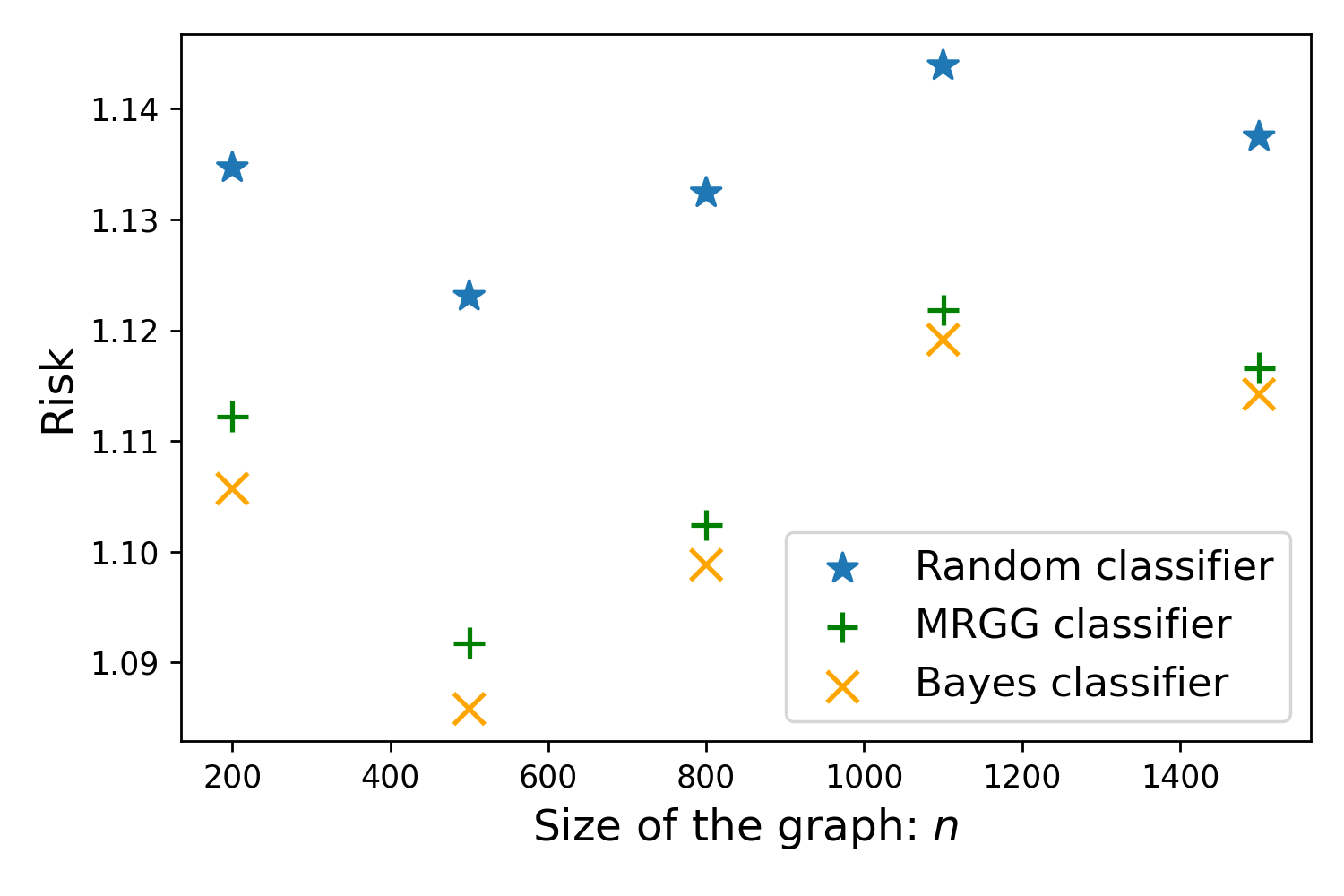} \vspace{-10pt}} \\
\subfloat[Envelope $\mathbf{p}^{(3)}$, \; Latitude $f_{\mathcal L}^{(3)}$]{\includegraphics[width = 2.9in]{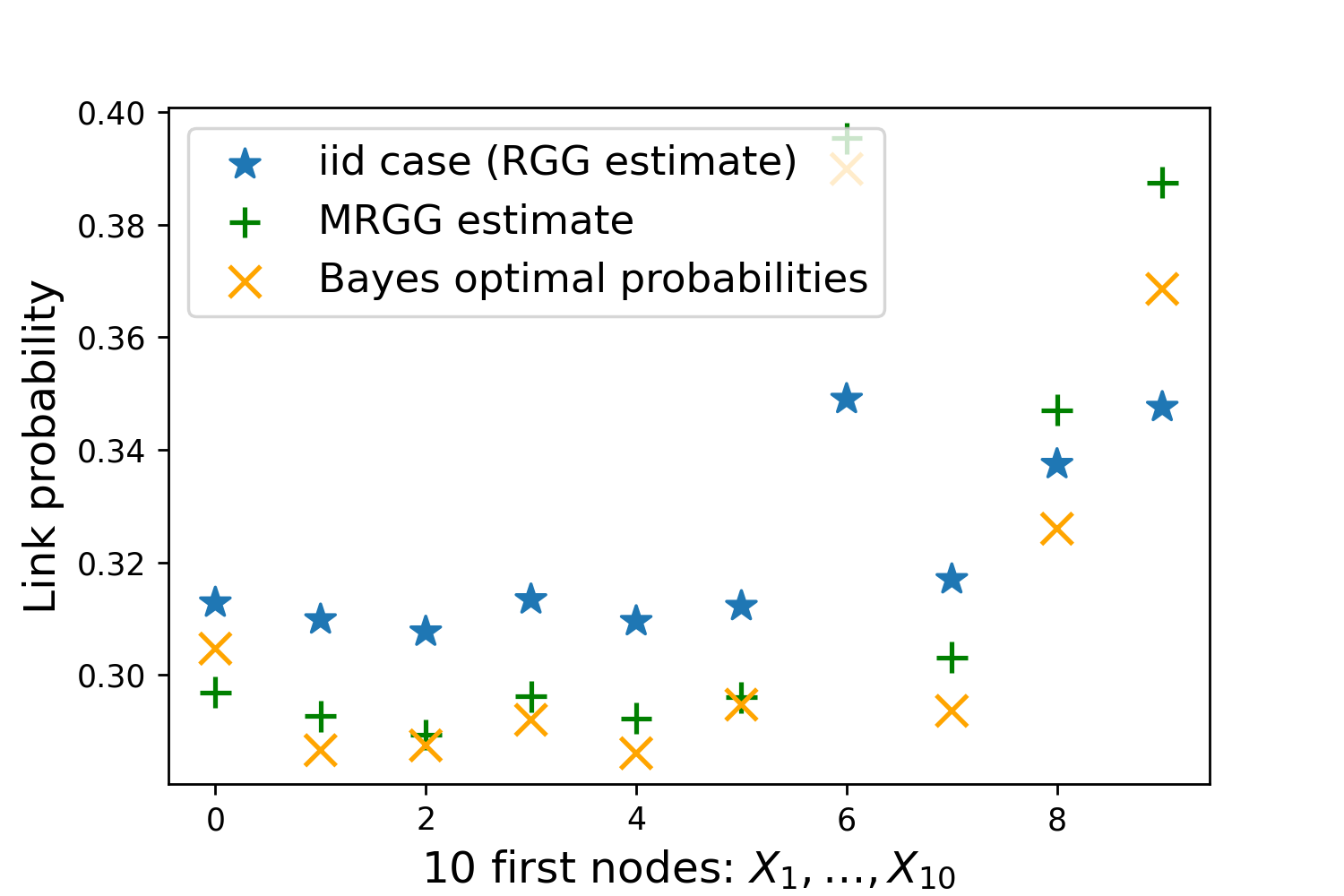}}
\subfloat[Envelope $\mathbf{p}^{(3)}$, \; Latitude $f_{\mathcal L}^{(3)}$]{\includegraphics[width = 2.9in]{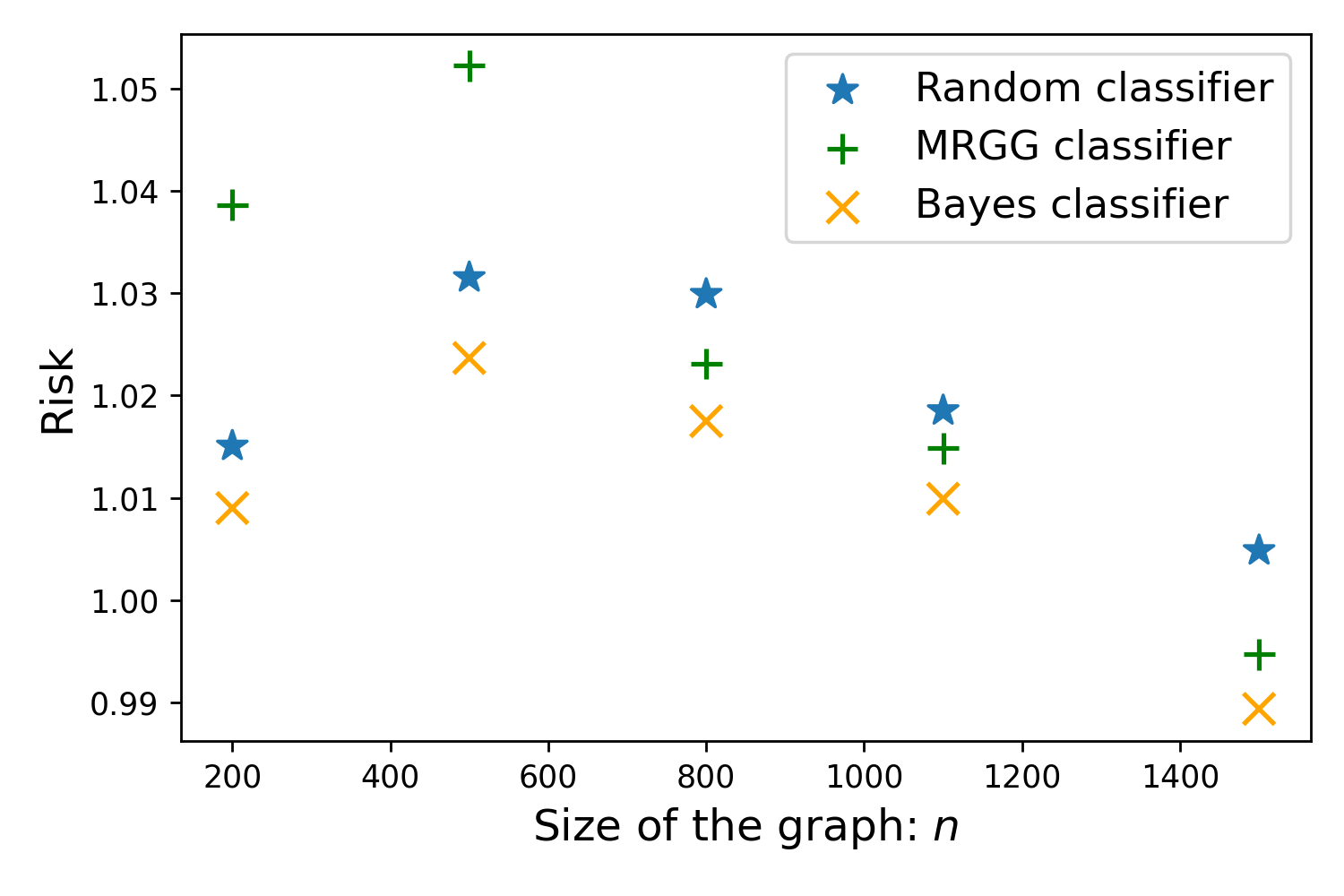} \vspace{-10pt}} 
\caption[Link predictions in the MRGG model.]{{\bf $\leftarrow$ On the left:} Link predictions between the future node $X_{n+1}$ and the $10$ first nodes $X_1, \dots, X_{10} $. {\bf $\rightarrow$ On the right:} Comparison between the risk (defined in Eq.\eqref{BLP:risk}) of the MRGG classifier, the random classifier and the risk of the optimal Bayes classifier.} 
\label{fig:link-risk}
\end{figure}

To illustrate our approach we work with a graph of $1 500$ nodes with $d=4$, and we consider the envelope and latitude functions defined in Eq.\eqref{eq:simu-env-lat}.
The plots on the left column of Figure~\ref{fig:link-risk} show that we are able to recover the probabilities of connection of the nodes already present in the graph with the coming node $X_{n+1}$. Using the decomposition of $\langle X_i,X_{n+1} \rangle$ given by
Eq.\eqref{eq:decompo}, orange crosses are computed using Eq.\eqref{eq:link}. Green plus are computed similarly replacing $\mathbf{p}$ and $f_{\mathcal L}$ by their estimations $\widehat {\mathbf{p}}$ and $\hat f_{\mathcal L}$ following Eq.\eqref{eq:link-esti}. Blue stars are computed using Eq.\eqref{eq:link} by replacing $f_{\mathcal L}$ by $\frac {w_{\beta}}{\|w_{\beta}\|_1}$ (with $\beta=\frac{d-2}{2}$) which implicitly supposes that the points are sampled uniformly on the sphere.

With the plots on the left column of Figure~\ref{fig:link-risk}, we compare the risk of the {\it random} classifier - whose guess $g_i(\mathbf D_{1:n})$ is a Bernoulli random variable with parameter given by the ratio of edges compared to complete graph - with the risk of the MRGG classifier (cf. Definition~\ref{def:mrgg-classifier}). These figures show that for a small number of nodes, the risk estimate provided by the MRGG classifier can be significantly far from the one of the Bayes classifier. However, when the number of nodes is getting larger, the MRGG classifier gives similar results compared to the optimal Bayes classifier. This risk estimate can be significantly smaller than the one of the random classifier (see for example the plots corresponding to the envelope $\mathbf{p}^{(2)}$ and the latitude $f_{\mathcal L}^{(2)}$).

\section{Discussion}
\label{sec:discussion}

In this section, we want to push the investigation of the performance of our estimation methods as far as possible. In Section~\ref{sec:mispe} we study the robustness of our methods under model mispecification before inspecting the influence of the mixing time of the Markov chain $(X_i)_{i\geq1}$ on the estimation error in Section~\ref{sec:mixing}. \\
On a more theoretical side, we show that replacing the use of the complete linkage by the Ward distance in the SCCHEi algorithm, Theorem~\ref{thm:SCCHEi} might not be true anymore. We conclude with some remarks and by highlighting future research directions.

\subsection{Robustness to model mispecification}
\label{sec:mispe}

\begin{figure*}[!ht]
    \centering
    \begin{subfigure}[b]{0.32\textwidth}
        \centering
        \includegraphics[width=\textwidth]{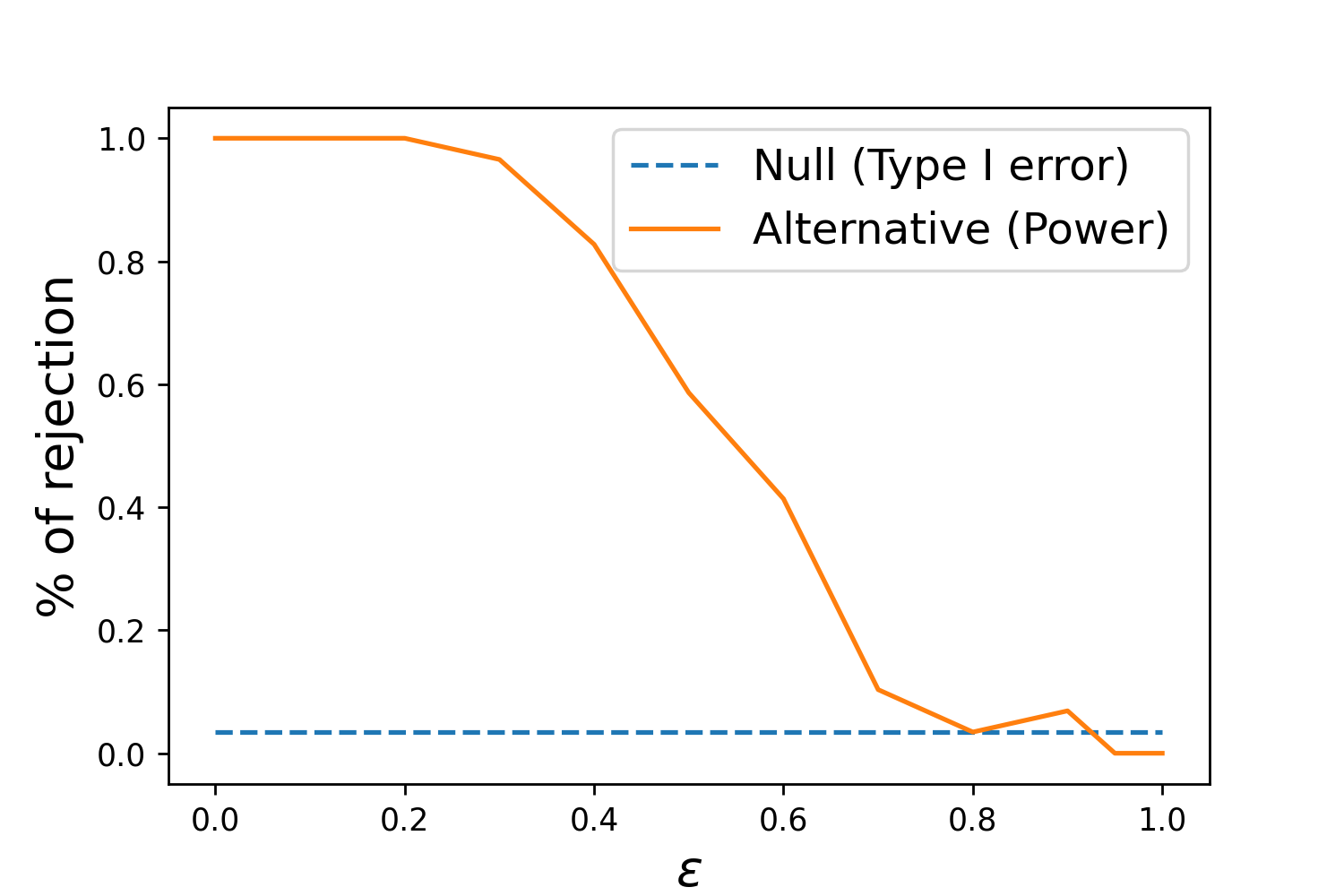}
        \caption[]%
        {{\small $n=200$}}    
    \end{subfigure}
    \hfill
    \begin{subfigure}[b]{0.32\textwidth}  
        \centering 
        \includegraphics[width=\textwidth]{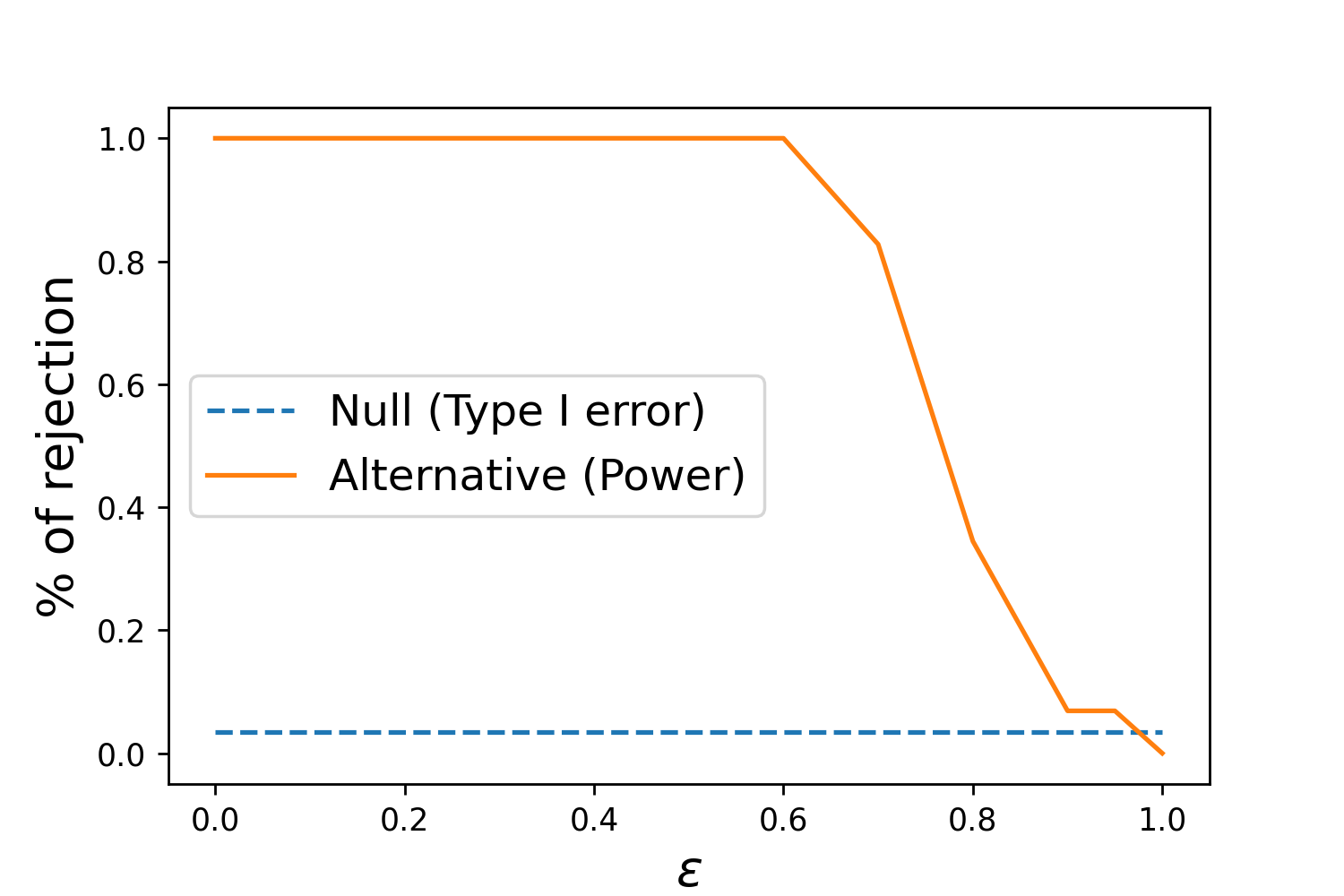}
        \caption[]%
        {{\small $n=500$}}    
    \end{subfigure}
    \hfill
    \begin{subfigure}[b]{0.32\textwidth} 
        \centering 
        \includegraphics[width=\textwidth]{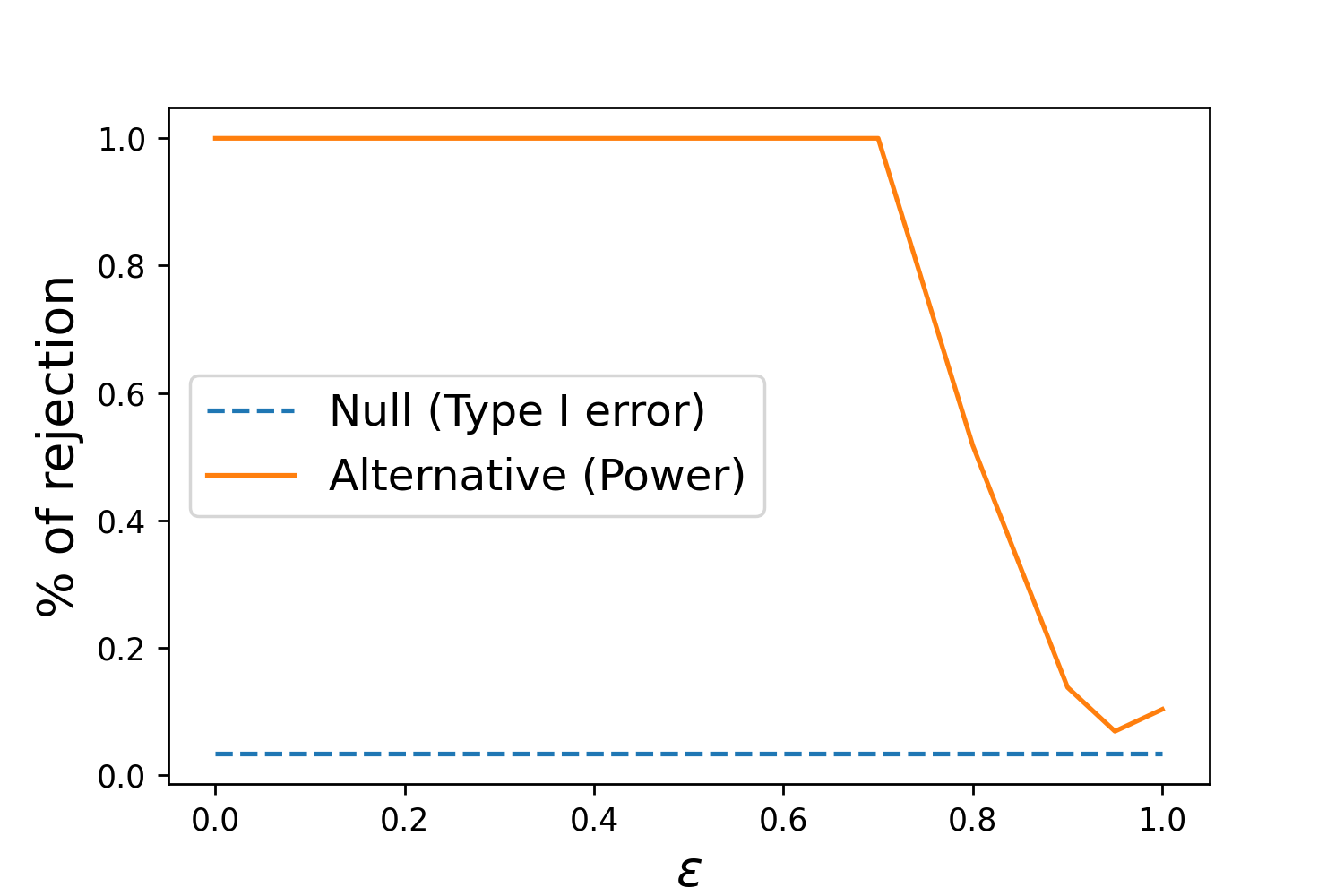}
        \caption[]%
        {{\small $n=1500$}}    
    \end{subfigure}
    \caption[Robustness of our methods to model mispecification (1/2).]{\small Studying the robustness of our method under model mispecification. We study the evolution of power for Markovian Dynamic Testing when the mixture parameter $\epsilon$ ranges $(0,1)$. We conduct this analysis for different values of $n.$  } 
    \label{fig:power_mixture}
\end{figure*}

\begin{figure*}[!ht]
    \centering
    \begin{subfigure}[b]{0.32\textwidth}
        \centering
        \includegraphics[width=\textwidth]{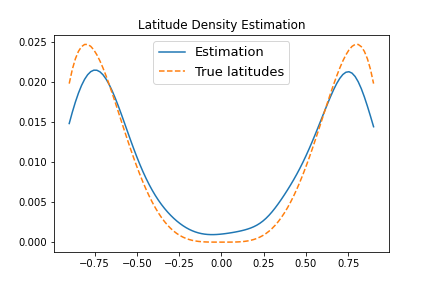}
        \caption[]%
        {{\small $\epsilon=0.1$}}    
    \end{subfigure}
    \hfill
    \begin{subfigure}[b]{0.32\textwidth}  
        \centering 
        \includegraphics[width=\textwidth]{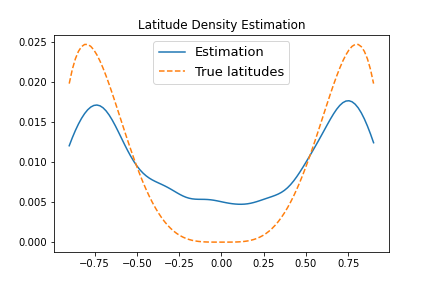}
        \caption[]%
        {{\small $\epsilon=0.5$}}    
    \end{subfigure}
    \hfill
    \begin{subfigure}[b]{0.32\textwidth} 
        \centering 
        \includegraphics[width=\textwidth]{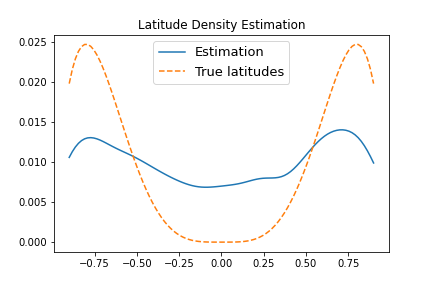}
        \caption[]%
        {{\small $\epsilon=0.7$}}    
    \end{subfigure}
    \caption[Robustness of our methods to model mispecification (2/2).]{\small Studying the robustness of the estimation of the latitude function under model mispecification. We plot our kernel density estimator of the latitude function for $n=1500$, $d=3$ and for $\epsilon \in \{0.1,0.5,0.7\}$. We use the envelope $\mathbf{p}^{(1)}$ and latitude function $f_{\mathcal L}^{(1)}$ defined in Eq.\eqref{eq:simu-env-lat}.} 
    \label{fig:latitude_mixture}
\end{figure*}

We consider a mixture model for the sampling scheme of the latent position. We fix some $\epsilon\in (0,1)$ and we draw $X_1$ randomly on the sphere. Then at time step $i\geq 2$, the point $X_i$ is sampled as follows:
\begin{itemize}
\item with probability $1-\epsilon$, $X_i$ is drawn following the Markovian dynamic described in Section~\ref{sec:dynamic} (based on $X_{i-1}$).
\item with probability $\epsilon$, $X_i$ is drawn uniformly on the sphere.
\end{itemize}
Figure~\ref{fig:power_mixture} and Figure~\ref{fig:latitude_mixture} show the numerical results obtained under this mispecified model. We consider the hypothesis testing question presented in Section~\ref{subsec:testing} with the same settings namely $d=3$ and the envelope and latitude functions $\mathbf{p}^{(1)}$ and $f_{\mathcal L}^{(1)}$ of Eq.\eqref{eq:simu-env-lat}. We can see that when $\epsilon=0$, the power of our test is 1 and we always reject the null hypothesis (uniform sampling of the latent positions) under the alternative. On the contrary, when $\epsilon=1$, the points are sampled uniformly on the sphere and we obtain a power of the order of the level of our test (i.e. $5\%$) as expected. The larger the sample size $n$ is, the greater $\epsilon$ can be chosen while keeping a large power. In the case where $n=1 500$, one can afford to sample $75\%$ of latent positions uniformly (and the rest using our Markovian sampling scheme) while keeping a power equal to 1. Figure~\ref{fig:latitude_mixture} shows that the larger $\epsilon$ is, the closer the estimated latitude function is to $\frac{w_{\beta}}{\|w_{\beta}\|_1}\equiv \frac12$ (since $d=3$) which corresponds to the density of a one-dimensional marginal of a uniform random point on $\mathds S^{d-1}.$

\subsection{Influence of mixing time on estimation error}
\label{sec:mixing}

In order to assert that the dependence of the latent variables has an influence on the estimation of the unknown functions of our model, we would require a minimax bound. The derivation of such minimax result is still an open problem, even in the independent setting (cf. \cite{CL18}). Nevertheless, by making explicit the constants involved in concentration inequalities, we can show that the mixing time of the latent Markovian dynamic affects our bound on the $\delta_2$ error between spectra. 
For any $r^*\in (-1,1)$, let us consider the following latitude function
\begin{equation*}f_{\mathcal L}^{r^*}(r) :=\frac{1}{I(r^*)} (1-r^2)^{\frac{d-3}{2}} \mathds 1_{r\in(r^*,1)}, \quad I(r^*) := \int_{r^*}^1(1-r^2)^{\frac{d-3}{2}}dr.\end{equation*}
Note that the Markov transition kernel $P$ of the chain $(X_i)_{i \geq1}$ using this latitude function is the one that starting from a point $x \in \mathds S^{d-1}$ samples uniformly a point in the set $\{ z \in \mathds S^{d-1} \; | \; \|x-z\|_2^2\leq 2 (1-r^*) \}.$ In particular, when $r^*=-1$, we recover the uniform distribution on the sphere. It is clear that the closer $r^*$ to one, the larger the mixing time of the chain. One can show that for any $r^*\in (-1,1)$, the chain is uniformly ergodic by proving that there exist an integer $m\geq 1$, a constant $\delta_m>0$ and a probability measure $\nu$ such that
\begin{equation}\label{eq:mixing-uni}\forall x \in \mathds S^{d-1}, \; \forall A \in \Sigma,\quad  P^m(x,A)\geq \delta_m \nu(A) \quad \text{(cf. Definition~\ref{apdx:uni-ergodicity}).}\end{equation}
Eq.\eqref{eq:mixing-uni} holds by considering for example $\nu=\pi$ the uniform distribution on the sphere. It is straightforward to show that the smallest integer $m(r^*)\geq1$ satisfying Eq.\eqref{eq:mixing-uni} is larger than $\frac{2}{1-r^*}$.\footnote{Indeed, the latitude function $f_{\mathcal L}^{r^*}$ allows to make a jump at each time step of size at most $1-r^*$. Since the length of the shortest arc on $\mathds S^{d-1}$ joining the north pole to the south pole is $2$, the result follows.}    Taking a closer look at the constants involved in the concentration inequality from \cite{duchemin20} (cf. \cite[Section 3.1.1]{duchemin20}), we get that
\[\mathds{E}\left[\delta_2^2(\lambda(\mathds{T}_W),\lambda(T_n))\vee \delta_2^2(\lambda(\mathds{T}_W),\lambda^{R_{opt}}(\widehat{T}_n)) \right]< C \left[ \frac{n}{\log^2(n)} \right]^{- \frac{2s}{2s+d-1}},\]
where $C > m(r^*)^2 \tau(r^*)^2 \|f_{\mathcal L}^{r^*}\|_{\infty}$ and $\tau(r^*)\geq1$ is the Orlicz norm of some regeneration time. Since for any $0<r^*<1$, 
\begingroup
\allowdisplaybreaks
\begin{align*}
I(r^*) &= \int_{r^*}^1 (1-r^2)^{\frac{d-3}{2}}dr= \int_{0}^{1-r^*} e^{\frac{d-3}{2}\ln(1-(r+r^*)^2)}dr\\
&= (1-(r^*)^2)^{\frac{d-3}{2}}\int_{0}^{1-r^*} e^{\frac{d-3}{2}\left\{\ln(1-(r+r^*)^2)-\ln(1-(r^*)^2)\right\}}dr\\
&\leq (1-(r^*)^2)^{\frac{d-3}{2}}\int_{0}^{1-r^*} e^{-\frac{d-3}{2}\left\{\frac{2r r^* +r^2}{1-(r^*)^2}\right\}}dr\\
&\leq (1-(r^*)^2)^{\frac{d-3}{2}}\int_{0}^{1-r^*} e^{-\frac{d-3}{2}\left\{2r r^* +r^2\right\}}dr\\
&\leq (1-(r^*)^2)^{\frac{d-3}{2}}\int_{0}^{1} e^{-\frac{d-3}{2}\left\{2r r^* \right\}}dr\\
&\leq  (1-(r^*)^2)^{\frac{d-3}{2}} \left(1 \wedge \frac{1}{r^* (d-3)}\right),
\end{align*}
\endgroup
we get that $ \|f_{\mathcal L}^{r^*}\|_{\infty}\geq \frac{1}{I(r^*)} (1-(r^*)^2)^{\frac{d-3}{2}}\geq r^*(d-3).$
Finally we obtain
\[C> \frac{2r^*}{1-r^*}(d-3),\]
where $r^* \mapsto \frac{2r^*}{1-r^*}(d-3)$ is increasing in $r^*$ and diverges to $+\infty$ when $r^*\to1^-$. Hence, the closer $r^*$ is to one, the slower the chain is mixing, and the poorer is our bound.

\begin{figure*}[!ht]
    \centering
    \begin{subfigure}[b]{0.49\textwidth}
        \centering
        \includegraphics[width=\textwidth]{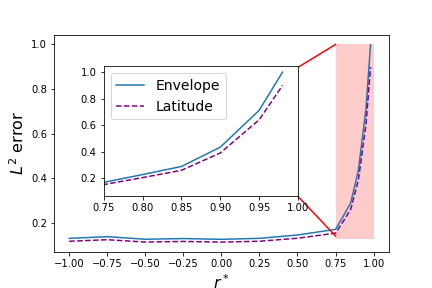}
        \caption[]%
        {{\small $n=200$}}    
    \end{subfigure}
    \hfill
    \begin{subfigure}[b]{0.49\textwidth} 
        \centering 
        \includegraphics[width=\textwidth]{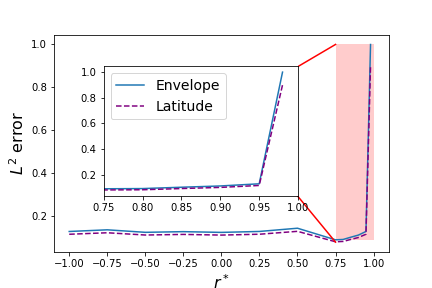}
        \caption[]%
        {{\small $n=1500$}}    
    \end{subfigure}
    \caption{\small Studying the influence of the mixing time of the chain on the $L^2$ errors between $(i)$ the envelope function and its estimate (using our adaptive procedure), and $(ii)$ the latitude function and its estimate obtained with a kernel estimator.} 
    \label{fig:dependency}
\end{figure*}

Figure~\ref{fig:dependency} presents the result of the simulations using the latitude function $f_{\mathcal L}^{r^*}$ and the envelope function $\mathbf{p}:t\mapsto \mathds 1_{t\geq 0}$. We compute the $L^2$ error between the true and the estimated envelope functions (respectively the true and the estimated latitude functions). When $r^*$ is getting closer to $1$, the chain is mixing slowly and we need to increase the sample size if we want to prevent the $L^2$ errors from blowing up. Graphs have been generated with a latent dimension $d=3$ and by sampling the latent positions using our isotropic sampling procedure with latitude function $f_{\mathcal L}^{r^*}.$

\subsection{Choice of the clustering algorithm for the SCCHEi}

The SCCHEi algorithm relies on the clustering of the eigenvalues of the adjacency matrix provided by the HAC with complete linkage. In this section, we motivate the use of the HAC algorithm with complete linkage by showing that the theoretical results from Section~\ref{sec:theo-guarantees} could be much more involved to establish by using another clustering procedure. Indeed, if one would consider for example the HAC with the Ward distance, the theoretical result obtained for the correctness of the SCCHEi algorithm (cf. Theorems~\ref{thm:SCCHEi} and~\ref{thm:poly}) is likely to be no longer true (even if the sample size $n$ is chosen arbitrarily large). Let us show this on a simple example. \\
\smallskip

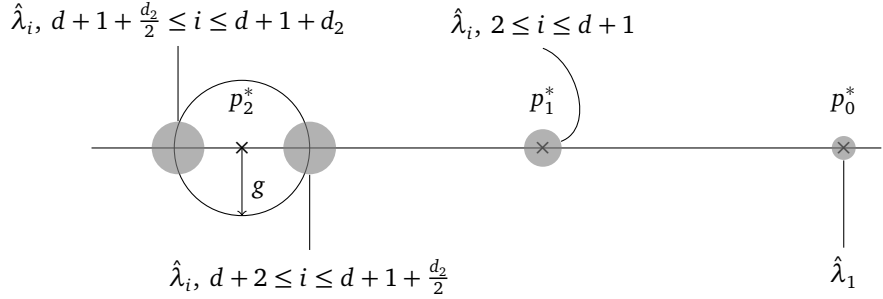
\begin{figure}[!ht]
\centering
\begin{tikzpicture}
\draw (-1,0) -- (9.5,0);
\node[circle,label=above:{$p_2^*$}] (p2) at (1,0) {$\times$};
\node[circle,label=above:{$p_1^*$}] (p1) at (5,0) {$\times$};
\node[circle,label=above:{$p_0^*$}] (p0) at (9,0) {$\times$};
\draw (1,0) circle (0.9cm) node[text=blue] {};

\node[circle,fill=gray,minimum size=7mm,,opacity=.6] (labrightp2) at (1.9,0) {};
\node[below=of labrightp2,yshift=-0pt] (labrightp2test) {$\hat \lambda_i, \; d+2\leq i \leq d+1+\frac{d_2}{2}$};
\draw  (labrightp2test.north) -- (labrightp2);
 \node[circle,fill=gray,minimum size=7mm,,opacity=.6] (lableftp2) at (0.15,0) {};
\node[above=of lableftp2,yshift=0pt] (lableftp2test) {$\hat \lambda_i, \; d+1+\frac{d_2}{2}\leq i \leq d+1+d_2$};
\draw  (lableftp2test.south) -- (lableftp2);
  
\node[circle,fill=gray,minimum size=5mm,,opacity=.6] (labp1) at (5,0) {};
\node[above=of p1,yshift=0pt] (lableftp1test) {$\hat \lambda_i, \; 2\leq i \leq d+1$};
\draw  (lableftp1test.south) to[out=-20,in=20] (labp1);

\node[circle,fill=gray,opacity=.6] (labp0) at (9,0) {};
\node[below=of p0,yshift=0pt] (lableftp0test) {$\hat \lambda_1$};
\draw  (lableftp0test.north) -- (labp0);
\draw[->] (1,0) -- (1,-0.9) node [right, sloped, yshift=10pt] (TextNode1) {$g$} ;
\end{tikzpicture}
\caption[Visualization of the counter-example proposed to stress the choice of the linkage function in our Hierarchical Clustering Algorithm.]{Visualization of the eigenvalues of the envelope function of our example.}\label{fig:cex-SCCHEi}
\end{figure}

We fix a resolution level $R=2$ and we consider some $\Delta^G>0$. We set $p^*_0=4\Delta^G$, $p^*_1=3\Delta^G$, $p^*_2=2\Delta^G$, and $p^*_k=0$ for all $k\geq 3$. Let us consider some $g \in (0,\Delta^G/4)$ that can be taken arbitrarily small. Let us denote $\lambda^{R}(\widehat{T}_n)=(\hat \lambda_1, \dots, \hat \lambda_{\widetilde R}, 0 ,0 \dots)$ and assume that it holds $\hat \lambda_1=p^*_0$, $\hat \lambda_2=\dots=\hat \lambda_{d+1}=p^*_1$ (we recall that $d_1=d$), $\hat \lambda_{d+2} =\dots=\hat \lambda_{d+1+\lfloor d_2/2\rfloor}=p^*_2+g$ and $\hat \lambda_{d+2+\lfloor d_2/2\rfloor} =\dots=\hat \lambda_{1+d+d_2}=p^*_2-g$. To simplify the presentation, we will assume in the following that $d_2=\frac{(d+1)d}{2}-1$ is even (which holds for example if $d=2k$ for any $k\geq 1$ odd). Figure~\ref{fig:cex-SCCHEi} gives a visualization of this example.

Applying the HAC algorithm (with the Ward distance) to the eigenvalues $(\hat \lambda_1, \dots , \hat \lambda_{\widetilde R})$, it is obvious that the state reached after $\widetilde R-4=1+d+d_2-4$ iterations in the HAC procedure will be 
\begin{align*}\widehat {\mathcal G}_0:=&  \{\hat \lambda_1\}\\
\widehat {\mathcal G}_{1}:=& \{\hat\lambda_{i} \; |\; 2\leq i\leq d\} \\
\widehat {\mathcal G}_{2}:=& \{\hat\lambda_{i} \; |\; d+2\leq i\leq d+1+d_2/2\}\\
\widehat {\mathcal G}_{3}:=& \{\hat\lambda_{i} \; |\;  d+2+ d_2/2 \leq i \leq 1+d+d_2\}
\end{align*}
Hence, in order to understand which clusters will be merged at the next step of the HAC algorithm, we compute the Ward distance between the different clusters.

Let us recall that for two finite and non-empty sets $S,S' \subset \mathds R$ with respective cardinality $|S|$ and $|S'|$, the Ward distance between $S$ and $S'$ is given by
\[d_W(S,S'):= \frac{|S| \times |S'|}{|S|+|S'|}\left(\frac{1}{|S|} \sum_{x_s \in S}x_s - \frac{1}{|S'|} \sum_{x'_s \in S'}x'_s\right)^2.\]

\begin{center}
\renewcommand{\arraystretch}{1.8}
\begin{tabular}{|c|c|c|c|}
\multicolumn{4}{c}{Ward distances between clusters}\\\cline{2-4}
\multicolumn{1}{c|}{ }&  \cellcolor{gray!25}$\widehat {\mathcal G}_{1}$ & \cellcolor{gray!50}$\widehat {\mathcal G}_{2}$ & \cellcolor{gray!75}$\widehat {\mathcal G}_{3}$\\ \cline{2-4}\hline
$\widehat {\mathcal G}_{0}$ &   $\frac{d}{d+1}(\Delta^G)^2$ & $\frac{d_2}{d_2+2}(2\Delta^G-g)^2$&$\frac{d_2}{d_2+2}(2\Delta^G+g)^2$\\
\cellcolor{gray!25} $\widehat {\mathcal G}_{1}$ & &$\frac{d\times d_2}{2d+d_2}(\Delta^G-g)^2$ &$\frac{d\times d_2}{2d+d_2}(\Delta^G+g)^2$\\
\cellcolor{gray!50}$\widehat {\mathcal G}_{2}$  & & & $d_2 \times g^2$\\\hline
\end{tabular}
\end{center}
We deduce that all Ward distances between pair of clusters are scaling at least linearly with $d$ except the Ward distances between $\widehat {\mathcal G}_{0}$ and the other three clusters $\widehat {\mathcal G}_{1}$, $\widehat {\mathcal G}_{2}$ and $\widehat {\mathcal G}_{3}$. Indeed, for any $i\in \{1,2,3\}$, $d_{W}(\widehat {\mathcal G}_{0},\widehat {\mathcal G}_{i}) $ remains bounded independently of the latent dimension $d$. Hence,
for any $ g\in (0,\Delta^G/4)$ which can be chosen arbitrarily small, one can take $d$ large enough to ensure that
\begin{equation}\label{eq:cex-SCCHEi}\max \left\{d_{W}(\widehat {\mathcal G}_{0},\widehat {\mathcal G}_{i}) \;,\;i\in\{1,2,3\}\right\} < d_{W}(\widehat {\mathcal G}_{2},\widehat {\mathcal G}_{3}) .\end{equation}
We deduce that for any $ g\in (0,\Delta^G/4)$, we can choose $d$ large enough to ensure that Eq.\eqref{eq:cex-SCCHEi} holds and thus the clusters merged between depths $4$ and $3$ from the root of the HAC's tree will not be  $\widehat {\mathcal G}_{2}$ and $\widehat {\mathcal G}_{3}$. This means that the state obtained at depth $3$ from the root is not of type $(\mathcal S)$ (in the sense defined in Lemma~\ref{lemma2:SCCHEi}). 

If this is not a sufficient condition to state that the SCCHEi will fail to recover the correct clusters, this example shows that the use of Ward distance can lead to some unexpected clustering of the eigenvalues. Our example proves that using the HAC algorithm with the Ward distance, the result of Lemma~\ref{lemma2:SCCHEi} does not hold anymore. Namely, regardless of how large the sample size is chosen, there are situations (in particular for a large latent dimension) where the states of type $(\mathcal S)$ (cf. Lemma~\ref{lemma2:SCCHEi}) are never reached in the HAC tree with the Ward distance. Hence obtaining a theoretical guarantee for the clustering provided by the SCCHEi in this framework may be impossible or at least much more involved.

\subsection{Concluding remarks}
\subsubsection{Estimation of the latent dimension}

The proposed methods implicitly assume that the latent dimension $d$ is known. \cite{AC19} proved that the latent dimension $d$ can be easily recovered in practice for $n$ large enough provided that the spectral gap condition~\eqref{gap-condition} holds. In the following, we briefly describe their approach.\\ Given some matrix $\widehat T_n$ as input and some set of candidates $\mathcal D$ for the dimension $d$ (typically $\mathcal D=\{2,3, \dots, d_{\max}\}$), apply the Algorithm HEiC (cf. Algorithm 3 in Section~\ref{sec:reminder_heic}) for any $d_c \in \mathcal D$ and store the returned value $gap:=gap(d_c)$. Let us recall that $gap(d_c)$ corresponds to the largest gap between a bulk of $d_c$ eigenvalues of $\widehat T_n$ and the rest of the spectrum (see the definition of $\mathrm{Gap}_1$ in Section~\ref{sec:reminder_heic} for details). Once we have computed the different gaps, we pick the candidate $d_c$ that led to the largest one.
Given the guarantees provided by Proposition~\ref{prop:HEiC}, the previously described procedure will find
the correct dimension, with high probability (on the event $\mathcal E$ with the notations of Proposition~\ref{prop:HEiC}), if the true dimension of the
latent space is in the candidate set $\mathcal D$. 

\subsubsection{Future research directions}

Our work encourages the development of growth model in random graphs and in particular the derivation of similar results in MRGGs with other latent spaces. It would be also desirable to extend our methods to the case where we consider more complex Markovian sampling of the latent positions, typically one that is not isotropic. Our work leaves open the question of getting a theoretical guarantee for the estimation of the latitude function. If we proved (with Theorem~\ref{gram}) that we can consistently estimate the Gram matrix of the latent positions in Frobenius norm, this is not sufficient to ensure that our kernel density estimator is consistent since we cannot ensure that $\frac{1}{n-1} \sqrt{\sum_{i=2}^{n} (r_i  - \hat r_i)^2}$ tends to 0 as $n$ goes to $+\infty.$ Deriving a theoretical result regarding the estimation of the latitude function seems challenging and we believe that it would require significantly different proof techniques.

\vspace*{\fill}
\vspace*{\fill}

\clearpage
\bibliography{sample} 

\clearpage

\appendix

\textbf{Guidelines for the Appendix} \smallskip\\
\underline{Sections \ref{apdx:def-MC} to \ref{apdx:Harmonic-Analysis}: Basic definitions and Complements}\smallskip\\
In Section \ref{apdx:def-MC} we recall basic definitions on Markov chains which are required for Section~\ref{apdx:properties-MC} where we describe some properties verified by the Markov chain $(X_i)_{i \geq1}$. Section~\ref{apdx:Harmonic-Analysis} provides complementary results on the Harmonic Analysis on $\mathds S^{d-1}$ which will be useful for our proofs.
\smallskip

\noindent \underline{Sections \ref{apdx:lemmas-SCCHEi} to \ref{sec:reminder_heic}: Algorithms and Experiments}\smallskip\\
In Section~\ref{apdx:lemmas-SCCHEi}, we give the proof of Lemma~\ref{lemma1:SCCHEi} and Lemma~\ref{lemma2:SCCHEi}. They are the cornerstones of the proof of Theorem~\ref{thm:SCCHEi} that provides a theoretical guarantee for the correctness of the algorithm SCCHEi. 
Section~\ref{apdx:slope-real-data} describes precisely the slope heuristic used to perform the adaptive selection of the model dimension $\hat R$. Section~\ref{sec:reminder_heic} provides a complete description of the HEiC algorithm used to extract $d$-eigenvectors of the adjacency matrix that will be used to estimate the Gram matrix of the latent positions. 
\smallskip

\noindent \underline{Sections \ref{proof-ustat} to \ref{apdx:gram}: Proofs of theoretical results}\smallskip\\
Thereafter, we dig into the most theoretical part of the Appendix. In Section~\ref{proof-ustat}, we discuss the assumptions we made on the Markov chain $(X_i)_{i \geq1}$. Section~\ref{proof-ustat} is also dedicated to the presentation of a concentration result for a particular U-statistic of the Markov chain $(X_i)_{i\geq1}$ that is an essential element of the proof of Theorem \ref{thm:delta2-proba-ope} which is provided in Section~\ref{proof-delta2-proba-ope}. Finally, the proof of Theorem~\ref{gram} can be found in Section~\ref{apdx:gram}.

\section{Definitions for general Markov chains}
\label{apdx:def-MC}

We consider a state space $E$ and a sigma-algebra $\Sigma$ on $E$ which is a standard Borel space. We denote by $(X_i)_{i\geq 1}$ a  time homogeneous Markov chain on the state space $(E,\Sigma)$ with transition kernel $P$.

\subsection{Ergodic and reversible Markov chains}

 \begin{definition}~\citep[section 3.2]{Roberts04} ($\phi$-irreducible Markov chains)\label{apdx:phi-irreducible} \\
 The Markov chain $(X_i)_{i\geq 1}$ is said $\phi$-irreducible if there exists a non-zero $\sigma$-finite measure $\phi$ on $E$ such that for all
$A \in \Sigma$ with $\phi(A)>0$, and for all $x \in E$, there exists a positive integer $n=n(x,A)$ such that $P^n(x,A)>0$ (where $P^n(x,\cdot)$ denotes the distribution of $X_{n+1}$ conditioned on $X_1=x$). 
 \end{definition}

\begin{definition}~\citep[section 3.2]{Roberts04} (Aperiodic Markov chains) \label{apdx:aperiodic}\\
The Markov chain $(X_i)_{i\geq 1}$  with  invariant  distribution $\pi$  is aperiodic if there do not exist $m\geq 2$  and  disjoint  subsets $A_1, \dots,A_m \subset E$ with $P(x,A_{i+1})  =  1 $ for  all $x\in A_i\;(1\leq i\leq m-1)$, and $P(x,A_1)  =  1$  for  all $x\in A_m$, such that $\pi(A_1)>0$ (and hence $\pi(A_i)>0$ for all $i$). 
\end{definition}

\begin{definition}~\citep[section 3.4]{Roberts04}  (Geometric ergodicity)\label{apdx:geo-ergodicity}\\
The Markov chain $(X_i)_{i\geq 1}$ is said geometrically ergodic if there exists an invariant distribution $\pi$, $\rho\in (0,1)$ and $C:E \rightarrow [1,\infty)$ such that \[ \|P^n(x,\cdot)-\pi \|_{TV} \leq C(x) \rho^n, \qquad \forall n \geq0, \; \pi\mathrm{-a.e}\;  x \in E,  \]
where $\|\mu\|_{TV}:= \sup_{A \in \Sigma}|\mu(A)|.$
\end{definition}

\begin{definition}~\citep[section 3.3]{Roberts04}  and \citep[Chapter 16]{tweedie} (Uniform ergodicity)\label{apdx:uni-ergodicity}\\
The Markov chain $(X_i)_{i\geq 1}$ is said uniformly ergodic if there exists an invariant distribution $\pi$ and constants $0<\rho<1$ and $L>0$ such that \[ \|P^n(x,\cdot)-\pi \|_{TV} \leq L \rho^n, \qquad \forall n \geq0, \; \pi\mathrm{-a.e}\;  x \in E,  \]
where $\|\mu\|_{TV}:= \sup_{A \in \Sigma}|\mu(A)|.$

\noindent 
Equivalently, the Markov chain $(X_i)_{i\geq1}$ is uniformly ergodic if the whole space $\mathds S^{d-1}$ is a small set, namely if there exist an integer $m\geq 1$, $\delta_m>0$ and a probability measure $\nu$ such that
\[\forall x \in \mathds S^{d-1}, \; \forall A \in \Sigma,\quad  P^m(x,A)\geq \delta_m \nu(A).\]
\end{definition}
\noindent {\bf Remark.} A Markov chain geometrically or uniformly ergodic admits a unique invariant distribution and is aperiodic.

\begin{definition} \label{apdx:reversibility}
A  Markov  chain is said reversible if there exists a distribution $\pi$ satisfying \[\pi(dx) P(x,dy) = \pi(dy)P(y,dx). \]
\end{definition}

 \subsection{Spectral gap}

This section is largely inspired from~\cite{FJS18}. Let us consider that the Markov chain $(X_i)_{i\geq 1}$ admits a unique invariant distribution $\pi$ on $\mathds S^{d-1}$.

For any real-valued, $\Sigma$-measurable function $h:E \rightarrow \mathds{R}$, we define $\pi(h):= \int h(x)\pi(dx)$. The set  \begin{equation}\notag \mathcal{L}_2(E,\Sigma,\pi):=\{h:\pi(h^2)<\infty\}\end{equation} is a Hilbert space endowed with the inner product \begin{equation}\notag\langle h_1,h_2\rangle_{\pi}=\int h_1(x)h_2(x)\pi(dx), \; \forall h_1,h_2 \in \mathcal{L}^2(E,\Sigma,\pi).\end{equation} The map \begin{equation}\notag\| \cdot \|_{\pi}: h\in \mathcal{L}_2(E,\Sigma,\pi) \mapsto \|h\|_{\pi}=\sqrt{\langle h,h \rangle_{\pi}},\end{equation} is a norm on $\mathcal{L}_2(E,\Sigma,\pi) $.  $\| \cdot \|_{\pi}$ naturally allows to define the norm of a linear operator $T$
 on $\mathcal{L}_2(E,\Sigma,\pi)$ as \begin{equation}\notag N_{\pi}(T)= \sup\{\|Th\|_{\pi}:\|h\|_{\pi}= 1\}.\end{equation}  To each transition probability kernel $P(x,B) $  with $x\in E$ and $B\in\Sigma$ invariant with respect to $\pi$, we can associate a bounded linear operator $h\mapsto \int h(y)P(\cdot,dy)$ on $\mathcal{L}_2(E,\Sigma,\pi)$. Denoting this operator $P$, we get \begin{equation}\notag Ph(x) =\int h(y)P(x,dy), \; \forall x \in E, \; \forall h \in \mathcal{L}_2(E,\Sigma,\pi).\end{equation}
 Let $\mathcal{L}_2^0(\pi) :=\{h\in  \mathcal{L}_2(E,\Sigma, \pi) \;: \; \pi(h) = 0 \}$. We define the absolute spectral gap of a Markov operator.
 \begin{definition} (Spectral gap) \label{apdx:spectralgap}
   A Markov operator $P$ reversible admits an absolute spectral gap $1-\lambda$ if \begin{equation}\notag\lambda := \sup\left\{\frac{\|Ph\|_{\pi}}{\|h\|_{\pi}}\;:\;h\in \mathcal{L}_2^0(\pi), \; h \neq0\right\}<1.\end{equation}
 \end{definition}
 The next result provides a connection between spectral gap and geometric ergodicity for reversible Markov chains.
 
 \begin{proposition}~\citep[section 2.3]{ferre12} \label{apdx:ergodic-gap}\\
A uniformly ergodic Markov chain admits a spectral gap.
\end{proposition}

\section{Properties of the Markov chain}
\label{apdx:properties-MC}

In the following, we denote $\lambda_{Leb} \equiv \lambda_{Leb,d}$ the Lebesgue measure on $\mathds{S}^{d-1}$ and $\lambda_{Leb,d-1}$ the Lebesgue measure on $\mathds{S}^{d-2}$. Using~\citep[Section 1.1]{Xu}, it holds $b_d:=\int_{ x\in\mathds{S}^{d-1}} \lambda_{Leb,d}(dx) = \frac{2\pi^{d/2}}{\Gamma(d/2)}$. Let $P$ be the Markov operator of the Markov chain $(X_{i})_{i \geq 1}$. By abuse of notation, we will also denote $P(x, \cdot)$ the density of the measure $P(x,dz)$ with respect to $\lambda_{Leb}(dz)$. For any $x,z \in \mathds{S}^{d-1}$, we denote $R_x^z \in \mathds{R}^{d \times d}$ a rotation matrix sending $x$ to $z$ (i.e. $R_x^z x=z$) and keeping $\mathrm{Span}(x,z)^{\perp}$ fixed. In the following, we denote $e_d:=(0,0,\dots,0,1)\in \mathds R^d$. 

\subsection{Invariant distribution and reversibility for the Markov chain}
\paragraph{Reversibility of the Markov chain $(X_i)_{i \geq 1}$.}

\begin{lemma} \label{kernel-MC}
For all $x,z \in \mathds{S}^{d-1}$, $P(x,z)=P(z,x)=P(e_d,R_z^{e_d}x).$
\end{lemma}

\begin{proof}[Proof of Lemma \ref{kernel-MC}.]
Using our model described in Section~\ref{model}, we get $X_{2} = r X_{1} + \sqrt{1-r^2}Y$ where conditionally on $X_1$, $Y$ is uniformly sampled on $\mathcal{S}(X_{1}):=\{q \in \mathds{S}^{d-1} \; : \; \langle q ,X_{1}\rangle=0\},$ and where $r$ has density $f_{\mathcal{L}}$ on $[-1,1]$. Let us consider a Gaussian vector $W \sim \mathcal{N}(0,I_d)$. Using the Cochran's theorem and Lemma~\ref{gaussian-sphere}, we know that conditionally on $X_1$, the random variable $\frac{W-\langle W,X_{1} \rangle X_{1}}{\|W-\langle W,X_{1} \rangle X_{1}\|_2}$ is distributed uniformly on $\mathcal{S}(X_{1})$. 

\begin{lemma} \label{gaussian-sphere}
Let $W \sim \mathcal{N}(0,I_d)$. Then, $\frac{W}{\|W\|_2}$ is distributed uniformly on the sphere $\mathds{S}^{d-1}.$
\end{lemma}

In the following, we denote $\overset{(d)}{=}$ the equality in distribution sense. We have conditionally on $X_1$
\[R_{X_1}^{e_d}\frac{W-\langle W,X_{1} \rangle X_{1}}{\|W-\langle W,X_{1} \rangle X_{1}\|_2}= \frac{\hat{W}-\langle \hat{W},e_d \rangle e_d}{\|\hat{W}-\langle \hat{W},e_d \rangle e_d\|_2} ,\]
where $\hat{W}=R_{X_1}^{e_d}W \sim \mathcal{N}(0,I_d)$. Using Cochran's theorem, we know that $\hat{W}-\langle \hat{W},e_d \rangle e_d$ is a centered normal vector with covariance matrix the orthographic projection matrix onto the space $\mathrm{Span}(e_d)^{\perp}$, leading to \[\hat{W}-\langle \hat{W},e_d \rangle e_d\overset{(d)}{=} \begin{bmatrix}Y \\ 0 \end{bmatrix},\] where $Y \sim \mathcal{N}(0,I_{d-1})$. Using Lemma~\ref{gaussian-sphere}, we conclude that conditionally on $X_1$, the random variable $\frac{W-\langle W,X_{1} \rangle X_{1}}{\|W-\langle W,X_{1} \rangle X_{1}\|_2}$ is distributed uniformly on $\mathcal{S}(X_{1})$ (because the distribution of $Y$ is invariant by rotation).
\medskip

We deduce that
\begin{align*}
X_{2}& \overset{(d)}{=} r X_{1} + \sqrt{1-r^2}\frac{W-\langle W,X_{1} \rangle X_{1}}{\|W-\langle W,X_{1} \rangle X_{1}\|_2} \\
& \overset{(d)}{=} r X_{1} + \sqrt{1-r^2}\frac{R_{X_2}^{X_1}W'-\langle R_{X_2}^{X_1} W',X_{1} \rangle X_{1}}{\|R_{X_2}^{X_1}W'-\langle R_{X_2}^{X_1} W',X_{1} \rangle X_{1}\|_2},
\end{align*}
where $W':=R_{X_1}^{X_2}W$. Note that $W'\in \mathds{R}^{d}$ is also a standard centered Gaussian vector because this distribution is invariant by rotation. Since $\langle R_{X_2}^{X_1} W',X_{1} \rangle = \langle W',X_2 \rangle$ and $\|R_{X_2}^{X_1}q\|_2=\|q\|_2, \; \forall q \in \mathds{S}^{d-1}$, we deduce that
\begin{equation}
X_{2} - r X_{1} \overset{(d)}{=} R_{X_2}^{X_1}\left[ \sqrt{1-r^2}\frac{W'-\langle  W',X_{2} \rangle X_{2}}{\|W'-\langle  W',X_{2} \rangle X_{2}\|_2} \right]. \label{rotation-distribution}
\end{equation}

$R_{X_1}^{X_2}$ is the rotation that sends $X_1$ to $X_2$ keeping the other dimensions fixed. Let us denote $a_1:=X_1$, $a_2:=\frac{X_{2} - r X_{1}}{\|X_{2} - r X_{1}\|_2}$ and complete the linearly independent family $(a_1,a_2)$ in an orthonormal basis of $\mathds{R}^{d}$ given by $a:=(a_1,a_2,\dots,a_d)$. Then, the matrix of $R_{X_1}^{X_2}$ in the basis $a$ is 
\[\begin{bmatrix}  r & -\sqrt{1-r^2} & 0_{d-2}^{\top}\\
\sqrt{1-r^2} & r & 0_{d-2}^{\top}\\
0_{d-2} & 0_{d-2} & I_{d-2} \end{bmatrix}.\]
We deduce that
\begingroup
\allowdisplaybreaks
\begin{align*}
\left(R_{X_2}^{X_1}\right)^{-1}\left(X_{2} - r X_{1}\right)  &= R_{X_1}^{X_2}\left(X_{2} - r X_{1}\right)  \\
&=\|X_{2} - r X_{1}\|_2 R_{X_1}^{X_2}\left(\frac{X_{2} - r X_{1}}{\|X_{2} - r X_{1}\|_2}\right)\\
&=\|X_{2} - r X_{1}\|_2 R_{X_1}^{X_2}a_2\\
&= \|X_{2} - r X_{1}\|_2 \left[ -\sqrt{1-r^2}a_1 + r a_2\right]\\
&= -\sqrt{1-r^2}\|X_{2} - r X_{1}\|_2 X_1 + r X_2-r^2X_1\\
&= -(1-r^2)X_1 +r X_2-r^2X_1\\
&= -X_1+rX_2.
\end{align*}
\endgroup
Going back to Eq.\eqref{rotation-distribution}, we deduce that \begin{equation} X_1 \overset{(d)}{=} rX_2 +\sqrt{1-r^2}\frac{\tilde{W}-\langle  \tilde{W},X_{2} \rangle X_{2}}{\|\tilde{W}-\langle  \tilde{W},X_{2} \rangle X_{2}\|_2}, \label{reverse-MC}\end{equation} where $\tilde{W}=-W'$ is also a standard centered Gaussian vector in $\mathds{R}^{d}$. Thus, we proved the first equality of Lemma \ref{kernel-MC}. Based on Eq.\eqref{reverse-MC} we have,
\begin{align*}R_{X_2}^{e_d}X_1
&\overset{(d)}{=} rR_{X_2}^{e_d}X_2 +\sqrt{1-r^2}\frac{R_{X_2}^{e_d}\tilde{W}-\langle  \tilde{W},X_{2} \rangle R_{X_2}^{e_d}X_{2}}{\|\tilde{W}-\langle  \tilde{W},X_{2} \rangle X_{2}\|_2}\\
&=re_d +\sqrt{1-r^2}\frac{R_{X_2}^{e_d}\tilde{W}-\langle  R_{X_2}^{e_d}\tilde{W},e_d \rangle e_d}{\|R_{X_2}^{e_d}\tilde{W}-\langle  R_{X_2}^{e_d}\tilde{W},e_d \rangle e_d\|_2},\end{align*}
which proves that $P(e_d,R_{x_2}^{e_d}x_1)= P(x_2,x_1)$ for any $x_1,x_1 \in \mathds S^{d-1}$ because $R_{X_2}^{e_d}\tilde{W}$ is again a standard centered Gaussian vector in $\mathds{R}^d$.
\end{proof}

\paragraph{Stationary distribution of the Markov chain.}

\begin{proposition} \label{apdx:uniform-invariant}
The uniform distribution on the sphere $\mathds{S}^{d-1}$ is a stationary distribution of the Markov chain $(X_i)_{i \geq 1}$.
\end{proposition}

\begin{proof}[Proof of Proposition~\ref{apdx:uniform-invariant}.]
Let us consider $z \in \mathds{S}^{d-1}$. We have using Lemma \ref{kernel-MC},
 \begingroup
 \allowdisplaybreaks 
\begin{align*}
&\int_{x \in \mathds{S}^{d-1}} P(x,z) \lambda_{Leb}(dx) =\int_{x \in \mathds{S}^{d-1}} P(z,x) \lambda_{Leb}(dx)=1,
\end{align*}
\endgroup
 which proves that the uniform distribution on the sphere is a stationary distribution of the Markov chain.
\end{proof}

 

\subsection{Ergodicity of the Markov chain}
\label{apdx:ergo-CS}

Our results hold under the condition that the Markov chain $(X_i)_{i\geq 1}$ is uniformly ergodic (cf. \hyperref[assumptionA]{Assumption A}). In this section, we provide a sufficient condition on the latitude function $f_{\mathcal L}$ for uniform ergodicity to hold.

\begin{lemma} \label{apdx:ergodicity}
We consider that $f_{\mathcal{L}}$ is bounded away from zero. Then, the Markov chain $(X_i)_{i \geq 1}$ is $\pi$-irreducible and aperiodic.
\end{lemma}

\begin{lemma} \label{uni-ergo-MC}
We consider that $f_{\mathcal{L}}$ is bounded away from zero. Then the Markov chain $(X_i)_{i \geq 1}$ is uniformly ergodic.
\end{lemma}

\begin{proof}[Proof of Lemmas \ref{apdx:ergodicity} and \ref{uni-ergo-MC}]

Considering for $\pi$ the uniform distribution on $\mathds{S}^{d-1}$, we get that for any $x \in \mathds{S}^{d-1}$ and any $A \subset \mathds{S}^{d-1}$ with $\pi(A)>0$,
 \begingroup
 \allowdisplaybreaks 
\begin{align}
&P(x,A)\notag\\
&= \int_{z \in A} P(x,z)\frac{\lambda_{Leb,d}(dz)}{b_d} \notag\\
&= \int_{z \in A} P(e_d,R_x^{e_d}z)\frac{\lambda_{Leb,d}(dz)}{b_d} \quad \text{(Using Lemma~\ref{kernel-MC})}\notag\\
&=  \int_{z \in R_x^{e_d}A} P(e_d,z)\frac{\lambda_{Leb,d}(dz)}{b_d} \notag\\& \text{(Using the change of variable }z \mapsto R_x^{e_d}z\text{ with }R_x^{e_d}A=\{R_x^{e_d}a : a \in A\}\text{)}\notag\\
&= \int_{r \in[-1,1]} \int_{\xi \in \mathds{S}^{d-2}} f_{\mathcal{L}}(r) 1_{(\xi^{\top}\sqrt{1-r^2},r)^{\top} \in R_x^{e_d}A}dr\frac{\lambda_{Leb,d-1}(d\xi)}{b_{d-1}b_d}  \notag\\
&\geq \inf_{s \in [-1,1]}{f_{\mathcal L}(s)} \int_{r \in[-1,1]} \int_{\xi \in \mathds{S}^{d-2}} 1_{(\xi^{\top}\sqrt{1-r^2},r)^{\top} \in R_x^{e_d}A}dr\frac{\lambda_{Leb,d-1}(d\xi)}{b_{d-1}b_d} \notag \\
&\geq \inf_{s \in [-1,1]}{f_{\mathcal L}(s)} \int_{r \in[-1,1]} \int_{\xi \in \mathds{S}^{d-2}} 1_{(\xi^{\top}\sqrt{1-r^2},r)^{\top} \in R_x^{e_d}A} \left(1-r^2\right)^{\frac{d-3}{2}}\frac{dr\lambda_{Leb,d-1}(d\xi)}{b_{d-1}b_d} \notag \\
&=  \frac{1}{b_{d-1}}\inf_{s \in [-1,1]}{f_{\mathcal L}(s)} \pi(R_x^{e_d}A)=  \frac{1}{b_{d-1}}\inf_{s \in [-1,1]}{f_{\mathcal L}(s)} \pi(A), \notag
\end{align}
\endgroup
since $\pi$ is invariant by rotation and $ f_{\mathcal{L}}$ is bounded away from zero. We also used that $\int_{-1}^1 (1-r^2)^{\frac{d-3}{2}}dr= \frac{b_d}{b_{d-1}}.$ This result means that the whole space $\mathds S^{d-1}$ is a small set. Hence, the Markov chain is uniformly ergodic (cf.~\citep[Theorem 16.0.2]{tweedie}) and thus aperiodic and $\pi$-irreducible.
\end{proof}

 \subsection{Computation of the absolute spectral gap of the Markov chain}
 \label{apdx:spectral-gap-1}

Thanks to Proposition~\ref{apdx:ergodic-gap} (in Appendix~\ref{apdx:def-MC}), we know that if $f_{\mathcal L}$ is such that $(X_i)_{i\geq 1}$ is uniformly ergodic, the Markov chain has an absolute spectral gap (cf. Definition~\ref{apdx:spectralgap}). In the following, we show that this absolute spectral gap is equal to $1.$
 
 Keeping notations of Appendix~\ref{apdx:def-MC}, let us consider $h \in L^2_0(\pi)$ such that $\|h\|_{\iffalse \varkappa \fi \pi}=1$. Then
 \begingroup
 \allowdisplaybreaks 
 \begin{align*}
 \|Ph\|_{\iffalse \varkappa \fi \pi}^2 &= \int_{x \in \mathds{S}^{d-1}} \left( \int_{y \in \mathds{S}^{d-1}}P(x,dy)h(y) \right)^2 \pi(dx)\\
 &= \int_{x \in \mathds{S}^{d-1}} \left( \int_{y \in \mathds{S}^{d-1}}P(x,y)h(y)\pi(dy) \right)^2 \pi(dx)\\
 &= \int_{x \in \mathds{S}^{d-1}} \left( \int_{y \in \mathds{S}^{d-1}}P(e_d,R_y^{e_d}x)h(y)\pi(dy) \right)^2 \pi(dx) \quad \text{(Using Lemma~\ref{kernel-MC})}\\
 &= \int_{x \in \mathds{S}^{d-1}} \left( \int_{y \in \mathds{S}^{d-1}}P(e_d,x)h(y)\pi(dy) \right)^2 \pi(dx)\\
 \qquad&\qquad \text{(Using the rotational invariance of $\iffalse \varkappa \fi \pi$)}\\
  &= \int_{x \in \mathds{S}^{d-1}}P(e_d,x)^2 \left( \int_{y \in \mathds{S}^{d-1}}h(y)\pi(dy) \right)^2 \pi(dx) \\
  &= 0 ,
 \end{align*}
 \endgroup
where the last equality comes from $h \in L^2_0(\pi)$. Hence, the Markov chain $(X_i)_{i \geq 1}$ has $1$ for absolute spectral gap.

  \section{Complement on Harmonic Analysis on the sphere}
 \label{apdx:Harmonic-Analysis}
 
 This section completes the brief introduction to Harmonic Analysis on the sphere $\mathds{S}^{d-1}$ provided in Section \ref{model}. We will need in our proof the following result which states that fixing one variable and integrating with respect to the other one with the uniform measure on $\mathds S^{d-1}$ gives $\|W-W_R\|_2^2.$

\begin{lemma}
\label{W-WR-inte}
For any $x \in \mathds S^{d-1}$, \[\mathds E_{X \sim\pi}[(W-W_R)^2(x,X)]=\|W-W_R\|_2^2,\] where $\pi$
 is the uniform measure on the $\mathds S^{d-1}$.
\end{lemma}
\begin{proof}[Proof of Lemma \ref{W-WR-inte}.]
 \begingroup
 \allowdisplaybreaks 
\begin{align*}
&\mathds E_{X\sim\pi}[(W-W_R)^2(x,X)]\\ &= \int_y (W-W_R)^2(x,y)\pi(dy)\\
&= \int_y \left(\sum_{r>R} p^*_{r}  \sum_{l=1}^{d_r} Y_{r,l}(x)Y_{r,l}(y) \right)^2\pi(dy)\\ 
&= \int_y \sum_{r_1,r_2>R} p^*_{r_1} p^*_{r_2}  \sum_{l_1=1}^{d_{r_1}}\sum_{l_2=1}^{d_{r_2}} Y_{r_1,l_1}(x)Y_{r_1,l_1}(y)Y_{r_2,l_2}(x)Y_{r_2,l_2}(y) \pi(dy)\\
&=  \sum_{r_1,r_2>R} p^*_{r_1} p^*_{r_2}  \sum_{l_1=1}^{d_{r_1}}\sum_{l_2=1}^{d_{r_2}} Y_{r_1,l_1}(x) Y_{r_2,l_2}(x) \int_y Y_{r_1,l_1}(y)Y_{r_2,l_2}(y) \pi(dy).
\end{align*}
\endgroup
Since $\int_y Y_{r,l}(y)Y_{r',l'}\pi(dy) $ is $1$ if $r=r'$ and $l=l'$ and 0 otherwise, we have that
\begin{align*}
\mathds E_{X\sim\pi}[(W-W_R)^2(x,X)]
&= \sum_{r>R} (p^*_r)^2  \sum_{l=1}^{d_r} Y_{r,l}(x)^2\\
&=  \sum_{r>R} (p^*_r)^2  d_r \quad \text{(Using~\citep[Eq.(1.2.9)]{Xu})}\\
&= \|W-W_R\|_2^2.
\end{align*}
\end{proof}

Let us consider $\beta:=\frac{d-2}{2}$ and the weight function $w_{\beta}(t) := (1-t^2)^{\beta-\frac12} $.
As highlighted in section \ref{model}, any envelope function $\mathbf{p}\in L^2([-1,1],w_{\beta})$ can be decomposed as $\mathbf{p} \equiv \sum_{k=0}^Rp^*_kc_kG_k^{\beta}$ where $G_l^{\beta}$ is the Gegenbauer polynomial of degree $l$ with parameter $\beta$ and where $c_k:= \frac{2k+d-2}{d-2}$. The Gegenbauer polynomials are orthonormal polynomials on $[-1,1]$ associated with the weight function $w_{\beta}$. The eigenvalues $(p^*_k)_{k \geq 0}$ of the envelope function can be computed numerically through the formula
\begin{equation*}\forall l \geq 0, \quad p^*_l =\left( \frac{c_l b_d}{d_l} \right) \int_{-1}^1 p(t)G_l^{\beta}(t) w_{\beta}(t)dt,\end{equation*}
 where
$ b_d:=\frac{\Gamma(\frac{d}{2})}{\Gamma(\frac12)\Gamma(\frac{d}{2}-\frac12)}$
with $\Gamma $ the Gamma function.  Hence, it is possible to recover the envelope function $\mathbf{p}$ thanks to the identity
\begin{equation}\mathbf{p} = \sum_{l\geq0}\sqrt{d_l} p_l^* \frac{G_l^{\beta}}{\| G_l^{\beta}\|_{L^2([-1,1],w_{\beta})}} = \sum_{l \geq 0} p_l^*c_l G_l^{\beta}. \label{reconstruction-p}\end{equation}

\section{Proofs of the two key lemmas for Theorem~\ref{thm:SCCHEi}}
\label{apdx:lemmas-SCCHEi}

In the proofs of Lemma~\ref{lemma1:SCCHEi} and Lemma~\ref{lemma2:SCCHEi} provided in this section, we keep the notations and the assumptions used in the proof of Theorem~\ref{thm:SCCHEi}. To ease the reading of this section, we recall here important notations.\\
We denoted \[\Delta^G = \min_{0 \leq k\neq l\leq R,\; p^*_k \neq p^*_l} \; |p^*_k - p^*_l| \wedge \min_{ 0\leq k\leq R,\;  p^*_k \neq 0} \; |p^*_k| >0.\]
For any $g \in (0, \frac{\Delta^G}{4}),$ the proof of Theorem~\ref{thm:delta2-proba-ope} (cf. Section~\ref{proof-delta2-proba-ope}) ensures that for $n$ large enough it holds
\begin{equation} \label{apdx-delta2-control}\delta_2^2(\lambda(\mathds{T}_{W_{R}}),\lambda^{R}(\widehat{T}_n))\leq g^2 . \end{equation}
Let us finally recall (cf. Section \ref{sec:intro}) that
\begin{align}\delta_2^2(\lambda(\mathds{T}_{W_{R}}),\lambda^{R}(\widehat{T}_n))&=\inf_{\sigma \in \mathfrak{S}} \sum_{i\geq 1} \left((\lambda(\mathds{T}_{W_{R}})_{\sigma(i)} -\lambda^{R}(\widehat{T}_n)_i \right)^2 \label{eq:apdx-delta2-hac}.
\end{align}

\subsection{Proof of Lemma~\ref{lemma1:SCCHEi}}

    We denote $\sigma^*$ a permutation achieving the minimum in Eq.\eqref{eq:apdx-delta2-hac}.
    
    $\bullet$ First we show that we can choose $\sigma^*$ such that $\sigma^*(\{1,\dots,\widetilde R\})=\{1,\dots,\widetilde R\}$.
    We recall that
    \begin{align*}
    \lambda(\mathds{T}_{W_{R}}) & =  \left(\underbrace{\overbrace{p^*_0}^{d_0=1},\overbrace{p^*_1,\dots,p^*_1}^{d_1=d},\dots, \overbrace{p^*_{R},\dots,p^*_{R}}^{d_{R}}}_{\widetilde R},0,0,\dots\right),\\
    \text{and } \quad \lambda^{R}(\widehat{T}_n) & = \left(\underbrace{\lambda^{R}(\widehat{T}_n)_1,\dots, \lambda^{R}(\widehat{T}_n)_{\widetilde R}}_{\widetilde R},0,0,\dots\right),
    \end{align*}
    with $\lambda^{R}(\widehat{T}_n)_1\geq \dots\geq \lambda^{R}(\widehat{T}_n)_{\widetilde R}$.

    $\rightsquigarrow$ If $p_k^* \neq 0$ for all $0\leq k \leq R$, then it is clear that $\sigma^*(\{1,\dots,\widetilde R\})=\{1,\dots,\widetilde R\}$. Otherwise, there would exist some $i \in \{1, \dots, \widetilde R\}$ such that $\sigma^*(j) \neq i $ for all $j\in \{1, \dots, \widetilde R\}$. Hence, we would obtain that $\delta_2^2(\lambda(\mathds{T}_{W_{R}}),\lambda^{R}(\widehat{T}_n))\geq |\lambda(\mathds{T}_{W_{R}})_i|^2\geq (\Delta^G)^2$, which would contradict Eq.\eqref{apdx-delta2-control}.
    
    $\rightsquigarrow$ If $p^*_k=0$ for all $0\leq k\leq R$, it is clear that we can take $\sigma^*=\mathrm{Id}$. 
    
    $\rightsquigarrow$ Otherwise, let us denote $Null$ the list of all indexes $i \in \{1,\dots,\widetilde R\}$ such that $\lambda(\mathds{T}_{W_{R}})_i=0$. It holds that $N_0=|Null| = \sum_{0\leq k \leq R \; s.t. \; p^*_k=0} d_k$. We also denote $NoNull$ the complement of $Null$ in $\{1,\dots,\widetilde R\}$ (i.e. the list of indexes in $\{1,\dots,\widetilde R\}$ that are not in $Null$). 
    
    For any $1\leq i \leq \widetilde R$ such that $\lambda(\mathds{T}_{W_{R}})_i\neq 0$, it must exist some $j \in \{1, \dots, \widetilde R\}$ such that $\sigma^*(j)=i$. Otherwise, we would have 
    \[\delta_2^2(\lambda(\mathds{T}_{W_{R}}),\lambda^{R}(\widehat{T}_n))\geq |\lambda(\mathds{T}_{W_{R}})_i|^2 \geq (\Delta^G)^2,\]
    which would contradict Eq.\eqref{apdx-delta2-control}. Hence, we get that \[(\sigma^*)^{-1}(NoNull)\subset \{1, \dots ,\widetilde R\}.\] We deduce that for any $i\in \{1, \dots ,\widetilde R\} \backslash (\sigma^*)^{-1}(NoNull)$, $\lambda(\mathds{T}_{W_{R}})_{\sigma^*(i)}= 0$. Hence, we can define $\sigma^*$ such that this permutation sends the $N_0$ indexes in $\{1, \dots ,\widetilde R\} \backslash (\sigma^*)^{-1}(NoNull)$ to the $N_0$ indexes in $Null$. Such $\sigma^*$ still achieves the minimum in Eq.\eqref{eq:apdx-delta2-hac}. In the following, we thus consider that $\sigma^*(\{1,\dots, \widetilde R\})=\{1,\dots, \widetilde R\}$.
    
    $\bullet$ Let us recall that the function $f^*$ is defined by 
    \begin{align*}
    f^*:\{1,\dots, \widetilde R\} &\rightarrow \{p^*_k,\; 0\leq k \leq R\}\\
    i &\mapsto \lambda(\mathds{T}_{W_{R}})_{\sigma^*(i)}.
    \end{align*}
    Note that for any $1\leq i\leq \widetilde R$, $\sigma^*(i)\leq \widetilde R$ thanks to the previous paragraph. We denote $p^*_{(0)} \geq \dots \geq p^*_{(R)}$ the ordered sequence of $p_0^*,\dots,p^*_{R}$ and $d_{(k)}$ is the multiplicity of the eigenvalue $p^*_{(k)}$ of the operator $\mathds T_W$. We show that $f^*$ is such that $f^*(1)=\dots =f^*(d_{(0)})=p^*_{(0)}$, $f^*(d_{(0)}+1)=\dots=f^*(d_{(0)}+d_{(1)})=p^*_{(1)}$, $f^*(d_{(0)}+d_{(1)}+1)=\dots=f^*(d_{(0)}+d_{(1)}+d_{(2)})=p^*_{(2)}$, $\dots$. This is equivalent to say that the function $f^*$ is non-increasing. If this was not true, it would mean that there exist $1\leq j<i \leq \widetilde R$ such that $f^*(j)<f^*(i)$. Since $\lambda^{R}(\widehat{T}_n)_{j}\geq\lambda^{R}(\widehat{T}_n)_{i}$ (because $j<i$), we would get that
    \begin{align*}
    \Delta^G &< f^*(i)-f^*(j)\\
    &=\underbrace{f^*(i)-\lambda^{R}(\widehat{T}_n)_{i}}_{\leq g}+\underbrace{\lambda^{R}(\widehat{T}_n)_{i}-\lambda^{R}(\widehat{T}_n)_{j}}_{\leq 0}+\underbrace{\lambda^{R}(\widehat{T}_n)_{j}-f^*(j)}_{\leq g}\\
    i.e. \quad & \Delta^G\leq2g.
    \end{align*}
    Since we chose $g$ such that $\Delta^G>4g$, this previous inequality is absurd. This concludes the proof.

\subsection{Proof of Lemma~\ref{lemma2:SCCHEi}}

    We prove our result by induction. In the following, we say that an intermediate state of the HAC algorithm is {\it valid} if it is still possible to reach state $(\mathcal S)$ in the next iterations of the algorithm. Stated otherwise, a state is {\it valid} if it does not exist $1\leq i\neq j \leq \widetilde R$ such that $f^*(i)\neq f^*(j)$ with $\lambda^{R}(\widehat{T}_n)_i$ and $\lambda^{R}(\widehat{T}_n)_j$ in the same cluster.
    It is obvious that the initial state of the HAC algorithm is {\it valid} since all eigenvalues are alone in their respective clusters.
    
    Suppose now that we are at iteration $2\leq t\leq \widetilde R -R-2$ of the HAC algorithm and that our procedure is {\it valid} until step $t$. We are sure that we did not reach a state of type $(\mathcal S)$ before step $t$ because only the state at depth $R$ from the root of the HAC's tree contains exactly $R+1$ clusters. For any cluster $S$ formed at step $t$ by the HAC algorithm, we denote by abuse of notation $f^*(S):=f^*(i)$ for any $i$ such that $\lambda^{R}(\widehat{T}_n)_i \in S$ (which is licit since step $t$ is {\it valid}). By contradiction, assume that the algorithm does not make a valid merging at step $t+1$. This means that the two merged clusters $S_a$ and $S_b$ at step $t+1$ are such that $f^*(S_a) \neq f^*(S_b)$. Since at step $t$ we did not reach a state of type $(\mathcal S)$, this means that there are two clusters $S_{i}$ and $S_{j}$ with $i\neq j$ such that $f^*(S_i)=f^*(S_j)$.\\
    For any $\lambda^{R}(\widehat{T}_n)_i\in S_i$ and $\lambda^{R}(\widehat{T}_n)_j\in S_j$,
    \[|\lambda^{R}(\widehat{T}_n)_i-\lambda^{R}(\widehat{T}_n)_j|\leq |\lambda^{R}(\widehat{T}_n)_i-f^*(S_i)| + \underbrace{|f^*(S_i)-\lambda^{R}(\widehat{T}_n)_j|}_{=|f^*(S_j)-\lambda^{R}(\widehat{T}_n)_j|}\leq 2g,\]
    and for any $\lambda^{R}(\widehat{T}_n)_a\in S_a$ and $\lambda^{R}(\widehat{T}_n)_b\in S_b$,
    \begin{align*}&|\lambda^{R}(\widehat{T}_n)_a-\lambda^{R}(\widehat{T}_n)_b|\\
    &\geq -|\lambda^{R}(\widehat{T}_n)_a-f^*(S_a)| + |f^*(S_a)-\lambda^{R}(\widehat{T}_n)_b| \\
    &\geq |f^*(S_a)-f^*(S_b)|-|\lambda^{R}(\widehat{T}_n)_a-f^*(S_a)|-|\lambda^{R}(\widehat{T}_n)_b-f^*(S_b)|\\
    &\geq \Delta^G - 2g.\end{align*}
    Since we chose $\Delta^G>4g$, we get
    \[d_c(S_a,S_b)>d_c(S_i,S_j).\]
    This is a contradiction since at step $t$, the HAC algorithm merges the two clusters with the smallest complete linkage distance. Hence, the algorithm performs a valid merging at step $t+1$.
    
    We proved that a state of type $(\mathcal S)$ is reached by the HAC algorithm with complete linkage at iteration $\widetilde R -R-1$.
    Since $d\geq3$, it holds $d_0<d_1<d_2<\dots$ and since the SCCHEi starts by selecting the cluster of size $d_0$ in the tree as close as possible to the root, we get $\mathcal C_{d_0} = \widehat {\mathcal C}_{d_0}$. Continuing the process of the "for loop" in the SCCHEi algorithm, the SCCHEi algorithm then selects the cluster of size $d_1$ in the remaining tree (where we removed all eigenvalues in $\widehat {\mathcal C}_{d_0}$ in the tree of the HAC). Hence, the SCCHEi algorithm sets $\mathcal C_{d_1} = \widehat {\mathcal C}_{d_1}$. Following this procedure, it is straightforward to see that the SCCHEi returns the partition $\mathcal C_{d_0}=\widehat {\mathcal C}_{d_0}, \dots, \mathcal C_{d_{R}}=\widehat {\mathcal C}_{d_{R}}$.

\section{Slope heuristic}
\label{apdx:slope-real-data}

\begin{figure}[!ht]
\begin{minipage}[c]{0.52\linewidth}
\includegraphics[width=\linewidth]{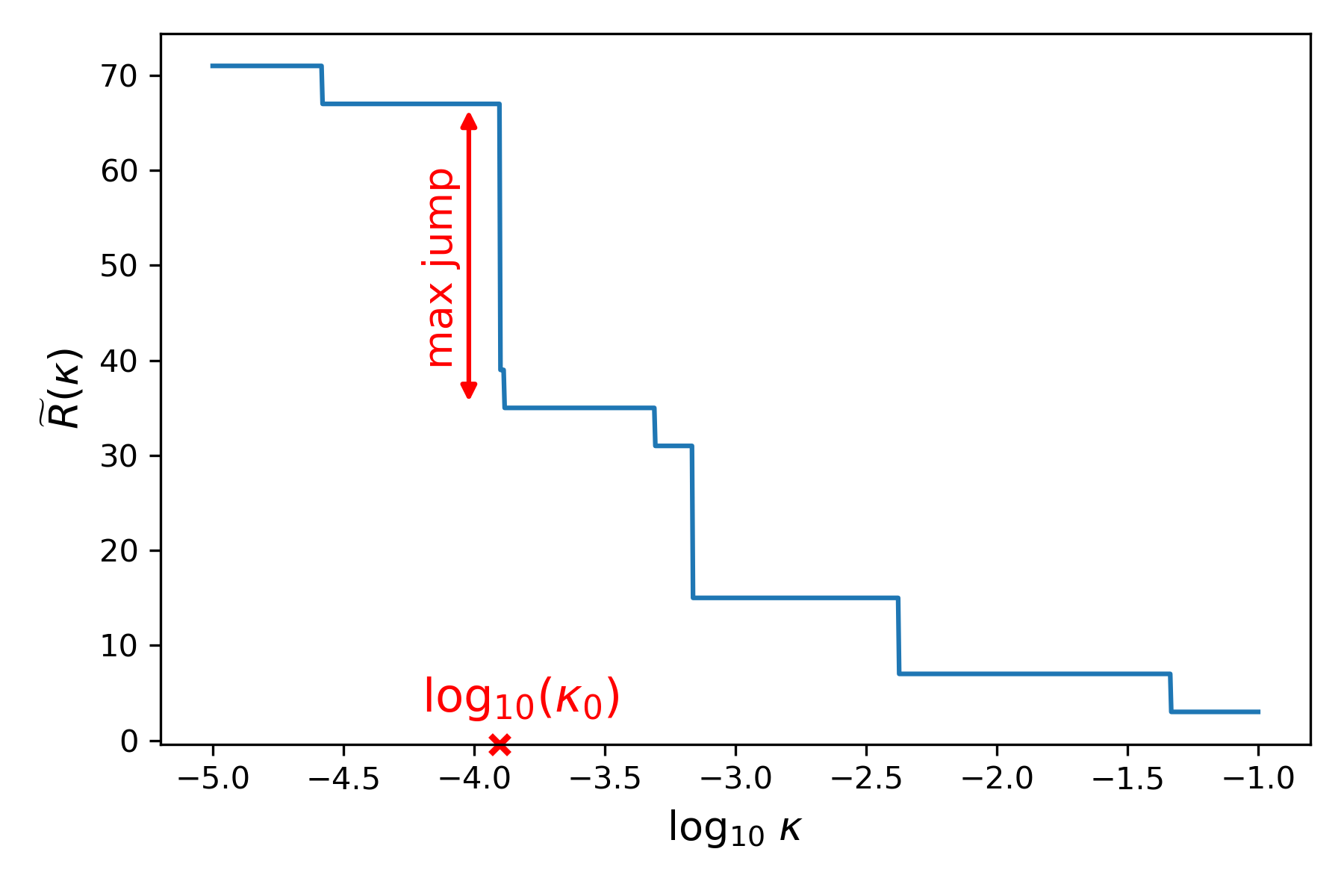}
\end{minipage} 
\begin{minipage}[c]{0.45\linewidth}
\caption{We sample the parameter $\kappa$ on a logscale between $10^{-5}$ and $10^{-1}$ and we compute the corresponding $R(\kappa)$ defined in Eq.\eqref{R-kappa}. We plot the values of $\widetilde R(\kappa)$ with respect to $\kappa$. The larger jump allows us to define $\kappa_0$.}
\label{fig:slope-jump}
\end{minipage}
\end{figure}

We propose a detailed analysis of the slope heuristic described in Section \ref{sec:slope-main} on simulated data using $d=3$, the envelope function $\mathbf{p}^{(1)}$ and the latitude function $f^{(1)}_{\mathcal L}$ presented in Eq.\eqref{eq:simu-env-lat}. We recall that $R(\kappa)$ represents the optimal value of $R$ to minimize the bias-variance decomposition defined by Eq.\eqref{R-kappa} for a given hyperparameter $\kappa$. Figure~\ref{fig:slope-jump} shows the evolution of $\widetilde R(\kappa)$ with respect to $\kappa$ which is sampled on a logscale. $\widetilde R(\kappa)$ is the dimension of the space of Spherical Harmonics with degree at most $R(\kappa)$. Our slope heuristic consists in choosing the value $\kappa_0$ leading to the larger jump of the function $\kappa \mapsto \widetilde R(\kappa)$. In our case, Figure~\ref{fig:slope-jump} shows that $\kappa_0=10^{-3.9}$. As described in Section \ref{subsec:SCCA}, the resolution level $\hat{R}$ selected to cluster the eigenvalues of the matrix $\widehat{T}_n$ is given by $R(2\kappa_0)$. 

\section{Reminder on Harmonic EigenCluster(HEiC)}
\label{sec:reminder_heic}

Before presenting the algorithm HEiC, let us define for a given set of indices $i_1,\dots,i_d \in [n]$ \begin{equation} \notag \mathrm{Gap}_1(\widehat{T}_n;i_1,\dots,i_d) := \min_{i \notin \{i_1,\dots,i_d\}} \max_{j \in \{i_1,\dots,i_d\}} |\hat{\lambda}_i-\hat{\lambda}_j|.\end{equation}

\begin{algorithm}  \caption{ Harmonic EigenCluster(HEiC) algorithm.}\label{HEiC}
\textbf{Data: } Adjacency matrix $A$. Dimension $d$.
	\begin{algorithmic}[1]
\STATE $ ( \hat{\lambda}_1^{sort}, \dots , \hat{\lambda}_{n}^{sort})\leftarrow $ eigenvalues of $\widehat{T}_n$ sorted in decreasing order.
\STATE  $\Lambda_1 \leftarrow \{ \hat{\lambda}_1^{sort}, \dots , \hat{\lambda}_{d}^{sort}\}$.
\STATE Initialize $i=2$ and gap$=\mathrm{Gap}_1(\widehat{T}_n; 1,2 , \dots,d)$.
    \WHILE{$i \leq n-d+1$} 
    \IF{$\mathrm{Gap}_1(\widehat{T}_n; i,i+1 , \dots,i+d-1)>\mathrm{gap}$} 
            \STATE $\Lambda_1 \leftarrow \{ \hat{\lambda}_i^{sort}, \dots, \hat{\lambda}_{i+d-1}^{sort}\}$
   \ENDIF
   \STATE $i = i+1$ \;
    \ENDWHILE
 	\end{algorithmic}
 	\textbf{Return:} $\Lambda_1,$ gap.
\end{algorithm}

\section{Concentration inequality for U-statistics with Markov chains}
\label{proof-ustat}

In this section, we present a recent concentration inequality for a U-statistic of the Markov chain $(X_i)_{i\geq 1}$ from~\cite{duchemin20} which is a key result to prove Theorem~\ref{thm:delta2-proba-ope}. In the first subsection, we remind the assumptions made on the Markovian dynamic, namely \hyperref[assumptionA]{Assumption A}.

\subsection{Assumptions and notations for the Markov chain}

Let us recall that \hyperref[assumptionA]{Assumption A} states that the latitude function $f_{\mathcal{L}}$ is such that $\|f_{\mathcal{L}}\|_{\infty} < \infty$ and makes the chain $(X_i)_{i\geq 1}$ uniformly ergodic. \hyperref[assumptionA]{Assumption A} guarantees in particular that there exists $\delta_M>0$ such that \[\forall x \in \mathds S^{d-1}, \forall A \in \mathcal{B}(\mathds S^{d-1}), \quad  P(x,A) \leq \delta_M  \nu(A),\] for some probability measure $\nu$ (e.g. the uniform measure on the sphere $\pi$).

\noindent In Section~\ref{apdx:ergo-CS}, we provide a sufficient condition on the latitude function $f_{\mathcal L}$ ensuring the uniform ergodicity of the chain with associated constants $L>0$ and $0<\rho<1$ (cf. Definition~\ref{apdx:uni-ergodicity}). In Section~\ref{apdx:spectral-gap-1}, we explain why \hyperref[assumptionA]{Assumption A} ensures that the Markov chain $(X_i)_{i\geq 1}$ has a spectral gap and we show that this spectral gap is equal to $1.$

\subsection{Concentration inequality of U-statistic for Markov chain}

One key result to prove Theorem \ref{thm:delta2-proba-ope} is the concentration of the following U-statistic
\[U_{stat}(n)=\frac{1}{n^2}\sum_{1\leq i<j \leq n}\left[ (W-W_R)^2(X_i,X_j)- \|W-W_R\|_2^2\right].\] 
Note that $\|W-W_R\|_2^2$ corresponds to the expectation of the kernel $(W-W_R)^2(\cdot,\cdot)$ under the uniform distribution on $\mathds S^{d-1}$ which is known to be the unique invariant distribution $\pi$ of the Markov chain $(X_i)_{i\geq 1}$ (cf. Appendix~\ref{apdx:properties-MC}). More precisely, for any $x \in \mathds S^{d-1}$, it holds \[\|W-W_R\|_2^2=\mathds E_{X \sim \pi}[(W-W_R)^2(x,X)] = \mathds E_{(X,X') \sim \pi \otimes \pi}[(W-W_R)^2(X,X')], \]
see Lemma~\ref{W-WR-inte} for a proof. Applying~\citep[Theorem~2]{duchemin20} in a our framework leads to the following result.

\begin{lemma}
\label{lemma:ustat} Let us consider $\gamma \in (0,1)$ satisfying $\log(e \log(n)/\gamma) \leq  n $. Then it holds with probability at least $1-\gamma$,
\[U_{stat}(n) \leq M \frac{\|\mathbf{p}-\mathbf{p}_R\|^2_{\infty}\log n}{n} \log (e\log (n)/\gamma) , \]
where $M>0$ only depends on constants related to the Markov chain $(X_i)_{i \geq 1}$.
\end{lemma}

\section{Proof of Theorem \ref{thm:delta2-proba-ope}}
\label{proof-delta2-proba-ope}

The proof of Theorem \ref{thm:delta2-proba-ope} mainly lies in the following result which is proved in Section~\ref{apdx-delta2-proba-ope}. Coupling the convergence of the spectrum of the matrix of probability $T_n$ with a concentration result on the spectral norm of random matrices with independent entries (cf.~\cite{bandeira2016}), we show the convergence in metric $\delta_2$ of the spectrum of $\widehat{T}_n$ towards the spectrum of the integral operator $\mathds{T}_W$.

\begin{theorem}
\label{delta2-proba-ope}
Let us consider $\gamma \in (0,1)$ satisfying $\log(e \log(n)/\gamma) \leq n/(13 \widetilde R)$. Then it holds with probability at least $1-\gamma$,
\begin{align*}
 &\delta_2\left( \lambda(\mathds{T}_W), \lambda(T_n) \right)\\
\leq \quad&  2\|\mathbf{p}-\mathbf{p}_R\|_2 + 8\sqrt{\frac{\widetilde R}{n}  \ln(e/\gamma)} + M \|\mathbf{p}-\mathbf{p}_R\|_{\infty}\sqrt{\frac{\log n}{n}} \left(\log (e\log (n)/\gamma)\right)^{1/2},
\end{align*}
 where $M>0$ only depends on constants related to the Markov chain $(X_i)_{i\geq 1}$ (cf. Lemma \ref{lemma:ustat}).
\end{theorem}

\paragraph{First part of the proof for Theorem~\ref{thm:delta2-proba-ope} }
We start by establishing the convergence rate for $\delta_2\left(\lambda(\mathds T_{ W}),\lambda(T_n)\right).$ We keep notations of Theorem \ref{delta2-proba-ope}. Let us consider $\gamma\in (0,1)$ satisfying $\log(e \log(n)/\gamma) \leq (n/(13 \widetilde R))$, and assume that $p \in Z^s_{w_{\beta}}((-1,1))$ with $s>0$.

Let us define the event 
\begin{align*}\Omega(\gamma):= &\Bigg\{ 
\delta_2\left( \lambda(\mathds{T}_W), \lambda(T_n) \right)\leq  2\|\mathbf{p}-\mathbf{p}_R\|_2 + 8\sqrt{\frac{\widetilde R}{n}  \ln(e/\gamma)}\\
&  +  M \|\mathbf{p}-\mathbf{p}_R\|_{\infty}\sqrt{\frac{\log n}{n}} \left( \log (e\log (n)/\gamma)\right)^{1/2} \Bigg\}.
\end{align*}
 
 Using Theorem \ref{delta2-proba-ope}, it holds $\mathds{P}\left(\Omega(\gamma)\right) \geq 1-\gamma$. Remarking further that 
 \[\delta_2\left( \lambda(\mathds{T}_W), \lambda(T_n) \right)\leq \delta_2\left( \lambda(\mathds{T}_W), 0 \right) +\delta_2\left( 0, \lambda(T_n) \right)\leq \|\mathbf{p}\|_2+\sqrt{n} \leq \sqrt{2}+\sqrt{n},\]
 we have
 \begin{align*}
& \mathds{E}[\delta_2^2(\lambda(\mathds{T}_W), \lambda(T_n))] \\=\quad & \mathds{E}[\delta_2^2(\lambda(\mathds{T}_W), \lambda(T_n)) \mathds{1}_{\Omega(\gamma)}]   + (1+\sqrt{2})^2n \mathds{P}(\Omega(\gamma)^c) \\
 \leq  \quad &  c \|\mathbf{p}-\mathbf{p}_R\|_2^2 +c \frac{\widetilde{R}}{n} \log(e/\gamma) 
 +c\|\mathbf{p}-\mathbf{p}_R\|_{\infty}^2\frac{\log n}{n}\log (e\log (n)/\gamma) \\ & \quad + (1+\sqrt{2})^2n\gamma,
 \end{align*}

where $c>0 $ is a constant that does not depend on $R$, $d$ nor $n$. Since for some constant $C(\mathbf p,s,d)>0$ (depending only on $\mathbf p$, $s$ and $d$)
\begin{equation}\|\mathbf{p}-\mathbf{p}_R\|_2^2 = \sum_{k>R} (p^*_k)^2d_k \frac{(1+k(k+2\beta))^s}{(1+k(k+2\beta))^s} \leq C(\mathbf p,s,d)R^{-2s}, \label{R2s}\end{equation}
and since \begin{equation}\widetilde{R} = O(R^{d-1}),\label{tildeR}\end{equation}
we have choosing $\gamma=1/n^2$
\begin{equation}
 \mathds{E}[\delta_2^2(\lambda(\mathds{T}_W), \lambda(T_n))]
  \leq D'\left[ R^{-2s} + R^{d-1} \frac{\log(n)}{n} 
 + \|\mathbf{p}-\mathbf{p}_R\|_{\infty}^2\frac{\log^2(n)}{n}\right],\label{rate-eq}
 \end{equation}
where $D'>0$ is a constant independent of $n$ and $R$. Let us show that choosing $R= \lfloor \left(n/\log^2(n)\right)^{\frac{1}{2s+d-1}}\rfloor$ concludes the proof. Since $\| G_k^{\beta}\|_{\infty}=G_k^{\beta}(1)=d_k/c_k$, we get that \[\| \mathbf{p}_R\|_{\infty} \leq \sum_{k=0}^R |p^*_k|c_kG_{k}^{\beta}(1) =\sum_{k=0}^R |p^*_k|d_k \leq \sqrt{\widetilde{R}} \| \mathbf{p}_R\|_2,\]
and using Eq.\eqref{pr}, we deduce that
\begin{equation}\| \mathbf{p}-\mathbf{p}_R\|_{\infty}\leq  \|\mathbf{p} \|_{\infty}+\| \mathbf{p}_R\|_{\infty} \leq 1+\sqrt{2\widetilde{R}}.\label{p-pR}\end{equation}
Hence, Eq.\eqref{rate-eq} becomes 
\begin{align*}
 \mathds{E}[\delta_2^2(\lambda(\mathds{T}_W), \lambda(T_n))]
 & \leq D''\left[ R^{-2s} + R^{d-1} \frac{\log(n)}{n} 
 + \widetilde{R}\frac{\log^2(n)}{n}\right],
 \end{align*}
where $D''$ is a constant that does not depend on $n$ nor $R$. Choosing $R= \lfloor \left(n/\log^2(n)\right)^{\frac{1}{2s+d-1}}\rfloor$ and using Eq.\eqref{tildeR} we get 
\begin{align*}
 &\mathds{E}[\delta_2^2(\lambda(\mathds{T}_W), \lambda(T_n))] \\
  \leq \quad & D''\left[ \left( \frac{n}{\log^2(n)}\right)^{\frac{-2s}{2s+d-1}} +  2\left( \frac{n}{\log^2(n)}\right)^{\frac{d-1}{2s+d-1}} \frac{\log^2(n)}{n} 
 \right]\\ \leq \quad & 3D'' \left( \frac{n}{\log^2(n)}\right)^{\frac{-2s}{2s+d-1}}.
 \end{align*}

\paragraph{Second part of the proof for Theorem~\ref{thm:delta2-proba-ope} }

Let us recall that in the statement of Theorem~\ref{thm:delta2-proba-ope}, $\lambda^{R_{opt}}(\widehat{T}_n)$ is the sequence of the $\widetilde{R}_{opt}$ first eigenvalues (sorted in decreasing absolute values) of the matrix $\widehat{T}_n$ where $R_{opt}$ is the value of the parameter $R$ leading to the optimal bias-variance trade off, namely
\[\lambda^{R_{opt}}(\widehat{T}_n)=(\hat{\lambda}_1, \dots,\hat{\lambda}_{\widetilde R_{opt}}, 0,0,\dots).\] From the computations of the first part of the proof, we know that $R_{opt}=\floor{\left(n/\log^2(n)\right)^{\frac{1}{2s+d-1}}}.$ That corresponds to the situation where we choose optimally $R$ and it is in practice possible to approximate this best model dimension using e.g. the slope heuristic. Therefore, $\delta_2\left(\lambda(\mathds T_W),\lambda^{R_{opt}}(\widehat{T}_n)\right)$ is the quantity of interest since it represents the distance between the eigenvalues used to built our estimates $(\hat{p}_k)_k$ and the true spectrum of the envelope function $\mathbf p$. Since $\widetilde{R} = \mathcal{O}\left( R^{d-1} \right)$ for all integer $R\geq0$, we have $\widetilde{R}_{opt} = \mathcal{O}\left( (n/\log^2(n))^{\frac{d-1}{2s+d-1}} \right)$. We deduce that for $n$ large enough $2\widetilde R_{opt} \leq n$ and using~\citep[Proposition 15]{CL18} we obtain \begin{align} &\delta_2\left( \lambda^{R_{opt}}(\widehat{T}_n),\lambda(\mathds T_{W_{R_{opt}}}) \right)\nonumber\\ \leq \quad & \delta_2\left( \lambda(T_n),\lambda(\mathds T_{W_{R_{opt}}} )\right) + \sqrt{2\widetilde R_{opt}}\|\widehat{T}_n-T_n\|\nonumber\\
\leq \quad & \delta_2\left( \lambda(T_n),\lambda(\mathds T_{W})\right)+\delta_2\left( \lambda(\mathds T_{W}),\lambda(\mathds T_{W_{R_{opt}}} )\right) + \sqrt{2\widetilde R_{opt}}\|\widehat{T}_n-T_n\|,\label{decompo-thm1part2}\end{align}
where $\lambda(\mathds T_{W_{R_{opt}}}) = (\lambda^*_1, \dots,\lambda^*_{\widetilde R_{opt}},0,0,\dots)$. Let us consider $\gamma \in (0,1)$.  Using Theorem~\ref{delta2-proba-ope}, we know that with probability at least $1-\gamma$ it holds for $n$ large enough \begin{align*}\delta_2\left( \lambda(T_n),\lambda(\mathds T_{W})\right)\leq \quad & 2\|\mathbf{p}-\mathbf{p}_{R_{opt}}\|_2 + 8\sqrt{\frac{\widetilde R_{opt}}{n}  \ln(e/\gamma)} \\
\qquad & + M \|\mathbf{p}-\mathbf{p}_{R_{opt}}\|_{\infty}\sqrt{\frac{\log n}{n}} \left(\log (e\log (n)/\gamma)\right)^{1/2}.\end{align*}
Using Eq.\eqref{R2s}, Eq.\eqref{p-pR} and the fact that $\widetilde R = \mathcal{O}(R^{d-1})$, it holds with probability at least $1-1/n^2$, \begin{align*} \delta_2^2\left( \lambda(T_n),\lambda(\mathds T_{W})\right)&\leq c\left[R_{opt}^{-2s} +R_{opt}^{d-1} \frac{\log n}{n} + M  R_{opt}^{d-1}\frac{\log^2 n}{n} \right]\\
&\leq (M')^2(n /\log^2 n)^{\frac{-2s}{2s+d-1}},\end{align*}
where $c>0$ is a numerical constant and $M'>0$ depends on constants related to the Markov chain $(X_i)_{i\geq1}$ (see Theorem~\ref{delta2-proba-ope} for details). 
Moreover, \begin{align}\delta_2^2\left( \lambda(\mathds T_{W}),\lambda(\mathds T_{W_{R_{opt}}} )\right)&=\|\mathbf{p}-\mathbf{p}_{R_{opt}}\|_2^2\notag\\
&\leq C(\mathbf p,s,d) R_{opt}^{-2s} = \mathcal{O}\left( (n /\log^2 n)^{\frac{-2s}{2s+d-1}} \right),\label{W-WRopt}\end{align}where we used Eq.\eqref{R2s}. Finally, using the concentration of spectral norm for random matrices with independent entries from~\cite{bandeira2016}, there exists a universal constant $C_0>0$ such that conditionally on $(X_i)_{i\geq1}$, it holds
with probability at least $1-1/n^2,$
\[\|T_n-\widehat{T}_n\| \leq \frac{3}{\sqrt{2n}}+C_0 \frac{\sqrt{\log (n^3)}}{n}.\] Using again $\widetilde R = \mathcal{O}(R^{d-1})$, this implies that for $n$ large enough, it holds conditionally on $(X_i)_{i\geq1}$ with probability at least $1-1/n^2$, \[ \sqrt{2\widetilde R_{opt}}\|T_n-\widehat{T}_n\| \leq D(n /\log^2 n)^{\frac{-s}{2s+d-1}},\]
where $D>0$ is a numerical constant. From Eq.\eqref{decompo-thm1part2}, we deduce that $\mathds P(\Omega) \geq 1-2/n^2$ where the event $\Omega$ is defined by \begin{equation*}  \Omega = \left\{\delta_2^2\left( \lambda^{R_{opt}}(\widehat{T}_n),\lambda(\mathds T_{W_{R_{opt}}} )\right) \leq \left(C(\mathbf p,s,d)^{1/2}+D+M'\right)^2(n /\log^2 n)^{\frac{-2s}{2s+d-1}}\right\}.\end{equation*}

Remarking finally that
 \begin{align*}\delta_2\left( \lambda^{R_{opt}}(\widehat{T}_n),\lambda(\mathds T_{W_{R_{opt}}} )\right)&\leq \delta_2\left( \lambda(\mathds T_{W_{R_{opt}}}), 0 \right) +\delta_2\left( 0, \lambda(\widehat{T}_n) \right)\notag\\
 &\leq \|\mathbf{p}\|_2+\sqrt{n} \leq \sqrt{2}+\sqrt{n},\end{align*}
 we obtain
 \begin{align}
&\mathds E\left[\delta_2^2\left( \lambda^{R_{opt}}(\widehat{T}_n),\lambda(\mathds T_{W_{R_{opt}}} )\right)\right]\notag\\
\leq \quad &  \mathds E\left[\delta_2^2\left( \lambda^{R_{opt}}(\widehat{T}_n),\lambda(\mathds T_{W_{R_{opt}}} )\right) \;|\; \Omega \right] + \mathds P (\Omega^c)(\sqrt2+\sqrt n)^2\nonumber\\
\leq \quad & \left(C(\mathbf p,s,d)^{1/2}+D+M'\right)^2(n /\log^2 n)^{\frac{-2s}{2s+d-1}}+2 \frac{(\sqrt{2}+\sqrt{n})^2}{n^2}\nonumber\\
= \quad & \mathcal{O}\left((n /\log^2 n)^{\frac{-2s}{2s+d-1}}\right). \label{error-truncated}
\end{align}

Using the triangle inequality, Eq.\eqref{W-WRopt} and Eq.\eqref{error-truncated} lead to \begin{align*}&\mathds E\left[\delta_2^2\left( \lambda^{R_{opt}}(\widehat{T}_n),\lambda(\mathds T_{W} )\right)\right]\\
&\quad \leq 3 \mathds E\left[\delta_2^2\left( \lambda^{R_{opt}}(\widehat{T}_n),\lambda(T_{W_{R_{opt}}}  )\right)\right]+3\delta_2^2\left( \lambda(T_{W_{R_{opt}}} ),\lambda(T_{W}  )\right)\\
&\quad =\mathcal{O}\left((n /\log^2 n)^{\frac{-2s}{2s+d-1}}\right),\end{align*} which concludes the proof of Theorem~\ref{thm:delta2-proba-ope}.

\subsection{Proof of Theorem \ref{delta2-proba-ope}}
\label{apdx-delta2-proba-ope}

We follow the same sketch of proof as in~\cite{CL18}. Let $R\geq 1$ and define,
\begingroup
 \allowdisplaybreaks 
\begin{align*}
\Phi_{k,l} &= \frac{1}{\sqrt {n}}\left[ Y_{k,l}(X_1), \dots , Y_{k,l}(X_n) \right] \in \mathds{R}^{ n} ,\\
E_{R,n} &= \left( \langle \Phi_{k,l},\Phi_{k',l'} \rangle- \delta_{(k,l),(k',l')}\right)_{(k,k') \in [R], \; l \in \{1,\dots,d_k \},\;l'\in\{1\dots,d_{k'} \}} \in \mathds{R}^{\widetilde{R}\times \widetilde{R}},\\
X_{R,n} &=\left[ \Phi_{0,1}, \Phi_{1,1}, \Phi_{1,2}, \dots , \Phi_{R,d_R} \right] \in \mathds{R}^{ n \times \widetilde{R}}, \\
A_{R,n} &= \left( X_{R,n}^{\top}X_{R,n} \right)^{1/2}\text{ with } A_{R,n}^2= \mathrm{Id}_{\widetilde{R}} + E_{R,n},\\
K_R & = \mathrm{Diag}(\lambda_1(\mathds{T}_W), \dots, \lambda_{\widetilde{R}}(\mathds{T}_W)) ,\\
T_{R,n} &= \sum_{k=0}^R p_k^* \sum_{l=1}^{d_k}\Phi_{k,l}(\Phi_{k,l})^{\top} = X_{R,n}K_R X_{R,n}^{\top} \in \mathds{R}^{n \times n} \\
\tilde{T}_{R,n} &= ((1-\delta_{i,j})T_{R,n})_{i,j  \in [n]} \in \mathds{R}^{n \times n}, \\
T^*_{R,n} &= A_{R,n}K_R A_{R,n}^{\top} \in \mathds{R}^{\widetilde{R} \times \widetilde{R}}, \\
W_R(x,y) &=  \sum_{k= 0}^R p^*_k \sum_{l=1}^{d_k} Y_{k,l}(x) Y_{k,l}(y).
\end{align*}
\endgroup

 It holds
 \[\delta_2(\lambda(\mathds{T}_W),\lambda(\mathds{T}_{W_R})) =\left(\sum_{k>R}d_k (p_k^*)^2\right)^{1/2}.\]
We point out the equality between spectra of the operator $\mathds{T}_{W_R}$ and the matrix $K_R$. Using the SVD decomposition of $X_{R,n}$, one can also easily prove that $\lambda(T_{R,n})=  \lambda(T_{R,n}^*)$. We deduce that 
\begin{align*}\delta_2\left( \lambda(\mathds{T}_{W_R}),\lambda(T_{R,n}) \right) &= \delta_2\left( \lambda(K_R),\lambda(T_{R,n}^*) \right) \\
&\leq \|T^*_{R,n} - K_R \|_F \\
&= \|A_{R,n}K_RA_{R,n} - K_R \|_F, \end{align*}
with the Hoffman-Wielandt inequality. Using equation (4.8) at (\cite{Gine} p.127) gives
\[\delta_2\left(\lambda(\mathds{T}_{W_R}),\lambda(T_{R,n})\right) \leq \sqrt{2}\|K_R\|_F \|E_{R,n}\|=\sqrt{2}\|W_R\|_2 \|E_{R,n}\|.\]
Using again the Hoffman-Wielandt inequality we get
\[\delta_2(\lambda(T_{R,n}),\lambda( \tilde{T}_{R,n}))\leq \| \tilde{T}_{R,n}-T_{R,n} \|_F =  \left[ \frac{1}{n^2} \sum_{i=1}^n W_R(X_i,X_i)^2 \right]^{1/2},\]
and
\[\delta_2\left(\lambda(\tilde{T}_{R,n}), \lambda(T_n)\right) \leq \| \tilde{T}_{R,n}-T_{n} \|_F = \left[ \frac{1}{n^2} \sum_{i\neq j}(W- W_R)^2(X_i,X_j) \right]^{1/2}.\]
Now, we invoke Lemmas  \ref{lemma:ustat}, \ref{ERN} and \ref{lemma:diag} to conclude the proof. The proofs of these last two lemmas are provided in Section~\ref{covering} and Section~\ref{square} respectively.

\begin{lemma}
Let us consider $\gamma >0$ and assume that $13 \widetilde R\ln(e/\gamma)\leq n$. Then it holds with probability at least $1-\gamma$
\[\| E_{R,n} \| \leq 4\sqrt{\frac{\widetilde R}{n}  \ln(2/\gamma)}.\]
\label{ERN}
\end{lemma}

\begin{lemma} \label{lemma:diag}
Let $R\geq 1$. We have
\begin{align*}
\frac{1}{n^2} \sum_{i=1}^n W_R(X_i,X_i)^2 =\frac{1}{n} \left( \sum_{k=0}^R p^*_k d_k\right)^2.
\end{align*}
\end{lemma}

For any $\gamma \in (0,1)$ with $\log(e \log(n)/\gamma) \leq  (n/(13 \widetilde R))$, it holds with probability at least $1-\gamma$,
\begin{align*}
&\delta_2\left( \lambda(\mathds{T}_W), \lambda(T_n) \right) \\
 \leq \quad & \delta_2\left( \lambda(\mathds{T}_W), \lambda(\mathds{T}_{W_R}) \right) +\delta_2\left( \lambda(\mathds{T}_{W_R}), \lambda(T_{R,n}) \right) +\delta_2\left( \lambda(T_{R,n}), \lambda(\tilde{T}_{R,n}) \right) \\ & \quad + \delta_2\left(  \lambda(\tilde{T}_{R,n}),\lambda(T_{n}) \right) \\
\leq \quad  &  4\sqrt{\frac{\widetilde R}{n}  \ln(2/\gamma)}+ \sqrt{2} \left( \sum_{k=0}^R d_k(p^*_k)^2 \right)^{1/2}
 + \frac{1}{\sqrt{n}}\left|\sum_{k=0}^R p^*_kd_k\right| + 2\|\mathbf{p}-\mathbf{p}_R\|_2 \\ & \quad +M \|\mathbf{p}-\mathbf{p}_R\|_{\infty}\sqrt{\frac{\log n}{n}} \left(\log (e\log (n)/\gamma)\right)^{1/2} ,
\end{align*}
where $M>0$ depends only on constants related to the Markov chain $(X_i)_{i\geq1}$. Now remark that
\[\left|\sum_{k=0}^R p^*_kd_k\right| \leq \left(\sum_{k=0}^R d_k\right)^{1/2}\left(\sum_{k=0}^R d_k(p^*_k)^2\right)^{1/2} = \sqrt{\widetilde{R}} \| \mathbf{p}_R\|_2 , \]
and that
 \begin{equation}\| \mathbf{p}_R\|_2^2 \leq \| \mathbf{p}\|_2^2 \leq 2 , \label{pr}\end{equation}
because $\mathbf{p}_R$ is the orthogonal projection of $\mathbf{p}$, and $|\mathbf{p}|\leq 1$. We deduce that 
\begingroup
 \allowdisplaybreaks 
 \begin{align*}
&\delta_2\left( \lambda(\mathds{T}_W), \lambda(T_n) \right)\nonumber\\ 
\leq \quad  & 2\|\mathbf{p}-\mathbf{p}_R\|_2 +4\sqrt{\frac{\widetilde R}{n}  \ln(2/\gamma)}+ \sqrt{\frac{2\widetilde{R}}{n}} \nonumber \\
\quad & +M \|\mathbf{p}-\mathbf{p}_R\|_{\infty}\sqrt{\frac{\log n}{n}} \left( \log (e\log (n)/\gamma)\right)^{1/2} \nonumber\\ 
\leq \quad  & 2\|\mathbf{p}-\mathbf{p}_R\|_2 + 8\sqrt{\frac{\widetilde R}{n}  \ln(e/\gamma)}\nonumber
\\\quad & + M \|\mathbf{p}-\mathbf{p}_R\|_{\infty}\sqrt{\frac{\log n}{n}} \left(\log (e\log (n)/\gamma)\right)^{1/2} .\end{align*}
\endgroup

\subsection{Proof of Lemma \ref{ERN}}
\label{covering}

Observe that $n E_{R,n}=\sum_{i=1}^n \left(Z_iZ_i^{\top}-\mathrm{Id}_{\widetilde{R}}\right)$ where for all $i \in [n]$, $Z_i \in \mathds{R}^{\widetilde{R}}$ is defined by \begin{align*}Z_i:=Z(X_i):= \big( & Y_{0,1}(X_i),Y_{1,1}(X_i), Y_{1,2}(X_i), \dots, Y_{1,d_1}(X_i) , \dots,\\
&Y_{R,1}(X_i), \dots, Y_{R,d_R}(X_i)\big).\end{align*}
By definition of the spectral norm for a Hermitian matrix,
\begin{align*}
\| \frac{1}{n}\sum_{i=1}^n Z_i Z_i^{\top} - \mathrm{Id}_{\widetilde{R}}\|& = \underset{x,\; \|x\|_2=1}{\max} \left| x^{\top}\left(\frac{1}{n}\sum_{i=1}^n Z_i Z_i^{\top}\right)x - 1 \right|.
\end{align*}

We use a covering set argument based on the following Lemma.
\begin{lemma}\label{coversetS-tropp} (cf.~\cite[Lemma 4.10]{P89})\\
Let us consider an integer $D \geq 2$. For any $\epsilon_0>0$, there exists a set $Q \subset \mathds{S}^{D-1}$ of cardinality at most $(1+2/\epsilon_0)^D$ such that \begin{equation}
\notag\forall \alpha \in \mathds{S}^{D-1}, \quad \exists q \in Q, \quad \| \alpha-q\|_2 \leq \epsilon_0.\end{equation}
\end{lemma}
We consider $Q$ the set given by Lemma \ref{coversetS-tropp} with $D=d$ and $\epsilon_0 \in (0,1/2).$
Let us define $x_0 \in \mathds S^{d-1}$ such that $|x_0^{\top}E_{R,n}x_0| = \|E_{R,n}\|$ and $q_0\in Q$ such that $\|x_0-q_0 \|_2\leq \epsilon_0$. Then,
\begin{align*}
|x_0^{\top}E_{R,n}x_0|-|q_0^{\top}E_{R,n}q_0|& \leq
|x_0^{\top}E_{R,n}x_0-q_0^{\top}E_{R,n}q_0| \text{ (by triangle inequality)} \\
&= |x_0^{\top}E_{R,n}(x_0-q_0)-(q_0-x_0)^{\top}E_{R,n}q_0|\\
&\leq \|x_0\|_2\|E_{R,n}\|\|x_0-q_0\|_2 +  \|q_0-x_0\|_2\|E_{R,n}\|\|q_0\|_2\\
&\leq 2 \epsilon_0\|E_{R,n}\|.
\end{align*}
which leads to \[|x_0^{\top}E_{R,n}x_0|=\|E_{R,n}\| \leq |q_0^{\top}E_{R,n}q_0| + 2 \epsilon_0\|E_{R,n}\|.\]
Hence,
\[ \| E_{R,n} \| \leq \frac{1}{1-2\epsilon_0}\max_{q\in Q} |q^{\top}E_{R,n}q|.\]
We introduce for any $q \in Q$ the function \[F_q:x=(x_1, \dots, x_n) \mapsto \frac{1}{n}  \sum_{i=1}^n q^{\top}\left(Z_iZ_i^{\top}-1\right)q : =\frac{1}{n}\sum_{i=1}^n f_q(x_i) , \]
where $f_q(x) = q^{\top}\left(Z(x)Z(x)^{\top}-1\right)q$.

Let us consider $t>0$. We want to apply Bernstein's inequality for Markov chains from~\citep[Theorem 1.1]{jiang2018bernsteins}. In the following, we denote $\mathds E_{\pi}[\cdot]$ the expectation with respect to the measure $\pi$. We remark that $\mathds{E}_{\pi}[f_q(X)]=0$ and that $\|f_q\|_{\infty}\leq \widetilde R -1$. For all $m \in [\widetilde R]$, we denote $\phi_m=Y_{r,l}$ with $r\in \{0,\dots, R\}$ and $l \in [d_r]$ such that
$m=l+\sum_{i=0}^rd_i-1$.  Then, for any $x \in \mathds S^{d-1},$ and for all $k,l \in [\widetilde R]$, $\left( (Z(x)^{\top}Z(x))^2\right)_{k,l} =  \sum_{m=1}^{\widetilde R} \phi_l(x) \phi_m(x)^2 \phi_k(x) = \widetilde R \phi_l(x) \phi_k(x) = \widetilde R \left( Z(x)Z(x)^{\top}\right)_{k,l}$ where we used~\citep[Eq.(1.2.9)]{Xu}. We deduce that 
\begin{align*}
\mathds E_{\pi}[f_q(X)^2] &= \mathds E_{\pi}[q^{\top}Z(X)Z(X)^{\top}qq^{\top}Z(x)Z(x)^{\top}q]-2\mathds E_{\pi}[q^{\top}Z(X)Z(X)^{\top}q]+1\\
&= \mathds E_{\pi}[q^{\top}\underbrace{(Z(X)Z(X)^{\top})^2}_{=\widetilde R \cdot Z(X)Z(X)^{\top}}q]-2q^{\top}\underbrace{\mathds E_{\pi}[Z(X)Z(X)^{\top}]}_{=\mathrm{Id}}q+1\\
&= \widetilde R \cdot q^{\top}\mathds E_{\pi}[Z(X)Z(X)^{\top}]q-1\\
&= \widetilde R -1.
\end{align*}

Using that the Markov chain $(X_i)_{i\geq 1}$ has an absolute spectral gap equals to $1$ (cf. Section~\ref{apdx:spectral-gap-1}), we get from~\citep[Eq. (1.6)]{jiang2018bernsteins} that
\[\mathds{P}\left(|F_q(X)|\geq t\right) = \mathds{P}\left(|q^{\top}E_{R,n}q|\geq t\right) \leq 2 \exp \left( \frac{- nt^2 }{ 4(\widetilde R-1)+10 (\widetilde R-1)t} \right),\]
which leads to\begin{align*}
\mathds{P}\left( \max_{q\in Q} |q^{\top}E_{R,n}q|\geq t\right)&\leq \mathds{P}\left( \bigcup_{q\in Q} |q^{\top}E_{R,n}q|\geq t\right) \\
&\leq 2\exp \left( \frac{- nt^2/(\widetilde R-1) }{ 4+10t} \right) \left( 1+2/\epsilon_0 \right)^{\widetilde R}.\end{align*}
Choosing $\epsilon_0 = 2 \left( \exp\left( \frac{nt^2/2}{(\widetilde R-1)\widetilde R (4+10t)}\right)-1\right)^{-1}$ in order to satisfy $(1+2/\epsilon_0)^{\widetilde R} = \exp(nt^2(\widetilde R-1)^{-1}(4+10t)^{-1}/2)$, we get
\[\mathds{P}\left( \max_{q\in Q} |q^{\top}E_{R,n}q|\geq t\right)\leq 2 \exp \left( \frac{- n t^2 }{(\widetilde R-1)(8+20t)} \right).\]

We deduce that if $\frac{25}{2}\ln(2/\alpha) \widetilde R \leq n$, it holds with probability at least $1-\alpha$, \[\max_{q\in Q} |q^{\top}E_{R,n}q| \leq 16 \sqrt{\frac{\widetilde R}{n}  \ln(2/\alpha)}.\]

Assuming that $200\ln(7) \widetilde R^3\ln(2/\alpha)\leq n^3$ in order to have $1/(1-2\epsilon_0)\leq 4$, it holds with probability at least $1-\alpha$
\[\| E_{R,n} \| \leq \frac{1}{1-2\epsilon_0}\max_{q\in Q} |q^{\top}E_{R,n}q| \leq 4\sqrt{\frac{\widetilde R}{n}  \ln(2/\alpha)}.\]

\subsection{Proof of Lemma \ref{lemma:diag}}
\label{square}

Reminding that for all $x \in \mathds{S}^{d-1}$ and for all $k \geq0$, $\sum_{l=1}^{d_k}Y_{k,l}(x)^2=d_k$ (cf. Corollary 1.2.7 from~\cite{Xu}), we get
\begin{align*}
\frac{1}{n^2} \sum_{i=1}^n W_R(X_i,X_i)^2 &= \frac{1}{n^2} \sum_{i=1}^n \left( \sum_{k=0}^R p^*_k \sum_{l=1}^{d_k} Y_{k,l}(X_i)^2\right)^2 \\
& =\frac{1}{n^2} \sum_{i=1}^n \left( \sum_{k=0}^R p^*_k d_k\right)^2 \\
&= \frac{1}{n} \left( \sum_{k=0}^R p^*_k d_k\right)^2.
\end{align*}

\section{Proof of Theorem \ref{gram}}
\label{apdx:gram}

Proposition \ref{prop:HEiC} is the counterpart of Proposition 1 in~\cite{AC19} in our dependent framework. 
This result is the cornerstone of Theorem \ref{gram} and is proved in Section~\ref{apdx:eventE}.

\begin{proposition} \label{prop:HEiC}
We assume that $\Delta^*>0$. Let us consider $\gamma>0$ and define the event \[\mathcal{E}:=\left\{\delta_2(\lambda(T_n),\lambda(\mathds{T}_W))\vee \frac{ 2^{\frac{9}{2}}\sqrt{d}}{\Delta^*}\|T_n-\widehat{T}_n\|\leq \frac{\Delta^*}{4}\right\}.\]
Then for $n$ large enough, \[\mathds{P}(\mathcal{E})\geq 1-\gamma/2.\]
Moreover, on  the  event $\mathcal{E}$,  there  exists  one  and  only  one  set $\Lambda_1$,  consisting  of $d$ eigenvalues of $\widehat{T}_n$, whose diameter is smaller that $\Delta^*/2$ and whose distance to the rest of the spectrum of $\widehat{T}_n$ is at least $\Delta^*/2$. Furthermore, on the event $\mathcal{E}$, the algorithm HEiC returns the matrix $\hat{G}= \frac{1}{d}\hat{V}\hat{V}^{\top}$, where $\hat{V}$ has by columns the eigenvectors corresponding to the eigenvalues in $\Lambda_1$.
\end{proposition}

In the following, we work on the event $\mathcal{E}$.
Let us consider $\gamma \in (0,1).$

 We choose $R =  (n/ \log^2 n)^{\frac{1}{2s+d-1}}.$ Reminding that $W_R$ is the rank $R$ approximation of $W$, the Gram matrix associated with the kernel $W_R$ is
\[T_{R,n} = \sum_{k=0}^R p_k^* \sum_{l=1}^{d_k}\Phi_{k,l}(\Phi_{k,l})^{\top} = X_{R,n}K_R X_{R,n}^{\top} \in \mathds{R}^{n \times n} \]
where 
\begingroup
 \allowdisplaybreaks 
\begin{align*}
\Phi_{k,l} &= \frac{1}{\sqrt {n}}\left[ Y_{k,l}(X_1), \dots , Y_{k,l}(X_n) \right] \in \mathds{R}^{ n} ,\\
X_{R,n} &=\left[ \Phi_{0,1}, \Phi_{1,1}, \Phi_{1,2}, \dots , \Phi_{R,d_R} \right] \in \mathds{R}^{ n \times \widetilde{R}} \text{ and }\\
K_R & = \mathrm{Diag}(\lambda_1(\mathds{T}_W), \dots, \lambda_{\widetilde{R}}(\mathds{T}_W)) .
\end{align*}
\endgroup

Let us denote now $\tilde{V}$ (resp. $\tilde{V}_R$) the orthonormal matrix formed by the eigenvectors of the matrix $T_n$ (resp. $T_{R,n}$). We have the following eigenvalue decompositions
\[T_n = \tilde{V} \Lambda \tilde{V}^{\top} \text{ and } T_{R,n} = \tilde{V}_R \Lambda_R \tilde{V}_R^{\top} ,\]
where $\Lambda=\mathrm{Diag}(\lambda_1, \dots, \lambda_{n})$ are the eigenvalues of the matrix $T_n$ and where \\$\Lambda_R = (p^*_0,p^*_1, \dots,p^*_1, \dots, p^*_{R}, \dots,p^*_{R},0,\dots,0) \in \mathds{R}^{n}$ where each $p^*_k$ has multiplicity $d_k$. Then, we note by $V \in \mathds{R}^{n\times d}$ (resp. $V_R$) the matrix formed by the columns $1, \dots, d$ of the matrix $\tilde{V}$ (resp. $\tilde{V}_R$). The matrix $V^* \in \mathds{R}^{n\times d}$ is the orthonormal matrix with $i-$th column $\frac{1}{\sqrt{n}}\left(Y_{1,1}(X_i), \dots, Y_{1,d}(X_i)\right)$.  The matrices $G^*, G, G_R$ and $G_{proj}^*$ are defined as follows
\begin{align*}
G^*&:= \frac{1}{c_1} V^*(V^*)^{\top}, &&\quad G:= \frac{1}{c_1} VV^{\top} \\
G_R&:= \frac{1}{c_1} V_RV_R^{\top}, &&\quad
G_{proj}^*:=V^*((V^*)^{\top}V^*)^{-1}(V^*)^{\top}.
\end{align*}

$G_{proj}^*$ is the projection matrix for the columns span of the matrix $V^*$. Using the triangle inequality we have \[\| G^*-G\|_F \leq \|G^*-G_{proj}^*\|_F + \|G_{proj}^*-G_R \|_F+\|G_R -G\|_F.\]

\paragraph{Step 1: Bounding $\|G-G_R\|_F.$}
 Since the columns of the matrices $V$ and $V_R$ correspond respectively to the eigenvectors of the matrices $T_n$ and $T_{R,n}$, applying the Davis Kahan sinus Theta Theorem (cf. Theorem \ref{davis-kahan}) gives that there exists $O \in \mathds{R}^{d \times d}$ such that
\[ \|VO-V_R\|_F\leq \frac{2^{3/2}\|T_n-T_{R,n}\|_F}{\Delta},\]

where $\Delta := \min_{k \in \{0,2,3,\dots,R\}} | p^*_1-p^*_{k}| \geq \Delta^* = \min_{k \in \mathds{N}, \; k \neq 1 } | p^*_1-p^*_{k}|$.  Using Lemma \ref{lemma:frobe2frobe-ope} and $c_1=\frac{d}{d-2}$, we get that \[\| G-G_R\|_F =\frac{d-2}{d}\| VO(VO)^{\top}-V_RV_R^{\top}\|_F  \leq 2\|VO-V_R \|_F.\] Hence, using the proof of Theorem \ref{thm:delta2-proba-ope}, we get that with probability at least $1-1/n^2$,
\[\| G-G_R\|_F \leq 2 \|VO-V_R\|_F\leq \frac{C}{\Delta^*} \left( \frac{n}{\log^2 n} \right)^{-\frac{s}{2s+d-1}},\]
where $C>0$ is a constant.
\paragraph{Step 2: Bounding~$\|G^*-G^*_{proj}\|_F$.}
To bound $\|G^*-G_{proj}^*\|_F$, we apply first Lemma \ref{lemma:cycle-frobe} with $B=V^*$. This leads to
\[\|G^*-G_{proj}^*\|_F \leq \|\mathrm{Id}_d - (V^*)^{\top}V^*\|_F \leq \sqrt{d}\|\mathrm{Id}_d - (V^*)^{\top}V^*\|.\]
Using a proof rigorously analogous to the proof of Lemma \ref{ERN}, it holds with probability at least $1-\gamma$ and for $n$ large enough,
\begin{align*}
\|\mathrm{Id}_d - (V^*)^{\top}V^*\| \leq  4\sqrt{\frac{d\log(e/\gamma)}{n}}.
\end{align*}
We get by choosing  $\gamma=1/n^2$ that it holds with probability at least $1-1/n^2$,
\begin{align*}
\|\mathrm{Id}_d - (V^*)^{\top}V^*\| \leq C' \sqrt{\frac{d\log(n)}{n}},
\end{align*}
where $C'>0$ is a universal constant.

\paragraph{Step 3: Bounding  $\|G^*_{proj}-G_R \|_F$.} We proceed exactly like in~\cite{AC19} but we provide here the proof for completeness.
Since $G^*_{proj}$ and $G_R$ are projectors we have, using for example~\citep[p.202]{bathia},
\begin{equation}\|G^*_{proj}-G_R\|_F =2 \|G^*_{proj}G_R^{\perp}\|_F. \label{diffG}\end{equation}
We use Theorem \ref{sintheta-stats} with $E = G^*_{proj}$, $F=G_R^{\perp}$, $B=T_{R,n}$ and $A=T_{R,n}+H$ where \[H=\tilde{X}_{R,n}K_R\tilde{X}_{R,n}^{\top} - X_{R,n} K_R X_{R,n},\]
where the columns of the matrix $\tilde{X}_{R,n}$ are obtained using a Gram-Schmidt orthonormalization process on the columns of $X_{R,n}$. Hence there exists a matrix $L$ such that $\tilde{X}_{R,n}=X_{R,n}(L^{-1})^{\top}.$ This matrix $L$ is such that a Cholesky decomposition of $X_{R,n}^{\top}X_{R,n}$ reads as $LL^{\top}$.

$A$ and $B$ are symmetric matrices thus we can apply Theorem \ref{sintheta-stats}. On the event $\mathcal{E}$, we can take $S_1 = (\lambda_1-\frac{\Delta^*}{8}, \lambda_1+\frac{\Delta^*}{8})$ and $S_2=\mathds{R} \backslash (\lambda_1-\frac{7\Delta^*}{8}, \lambda_1+\frac{7\Delta^*}{8})$. By Theorem \ref{sintheta-stats} we get
\begin{equation} \|G^*_{proj}G_R^{\perp}\|_F \leq \frac{\|A-B\|_F}{\Delta^*} = \frac{\|H\|_F}{\Delta^*}. \label{GprojGR}\end{equation}
We only need to bound $\|H\|_F$.
\begin{align}
\| H\|_F& \leq \|L^{-\top}K_RL^{-1} - K_R\|_F \|X_{R,n}^{\top}X_{R,n} \| \nonumber\\
& \leq \| K_R\|_F \|L^{-1} L^{-\top} - \mathrm{Id}_{\widetilde R} \| \| \label{H} X_{R,n}^{\top}X_{R,n}\|,\end{align}
where the last inequality comes from Lemma \ref{lemma:ostrowski}.
From the previous remarks on the matrix $L$, we directly get 
\[\|L^{-1} L^{-\top} - \mathrm{Id}_{\widetilde{R}}\| = \|\left( X_{R,n}^{\top}X_{R,n}\right)^{-1} - \mathrm{Id}_{\widetilde{R}}\|.\]
Using the notations of the proof of Theorem \ref{delta2-proba-ope} which is provided in Section~\ref{apdx-delta2-proba-ope}, we get
\[\|L^{-1} L^{-\top} - \mathrm{Id}_{\widetilde{R}}\|  \| X_{R,n}^{\top}X_{R,n}\| = \| X_{R,n}^{\top}X_{R,n} - \mathrm{Id}_{\widetilde{R}}\|=\|E_{R,n}\|.\]
Noticing further that $\|K_R\|_F^2 \leq \sum_{k\geq0}(p_k^*)^2d_k =\|\mathbf p\|_2^2\leq 2$ (because $|\mathbf p|\leq 1$), Eq.\eqref{H} becomes \begin{equation}\| H\|_F \leq \sqrt{2}\|E_{R,n}\| . \label{HERn}\end{equation}
Using Lemma \ref{ERN},  it holds with probability at least $1-\gamma$ and for $n$ large enough,
\begin{equation}
\|E_{R,n}\|   \leq 4\sqrt{\frac{\widetilde R}{n}  \ln(2/\gamma)}.\label{majoERn}
\end{equation}
 Since $\widetilde{R} =\mathcal{O}\left(R^{d-1}\right)$ and $R =\mathcal{O}\left(\left(  n/\log^2 n \right)^{\frac{1}{2s+d-1}}\right)$, we obtain using Eqs.\eqref{diffG}, \eqref{GprojGR}, \eqref{HERn} and \eqref{majoERn} that with probability at least $1-1/n^2$ it holds
\[\|G^*_{proj}-G_R\|_F = 2 \| G^*_{proj}G_R^{\perp}\|_F \leq \frac{C_{d}}{\Delta^*} \left(  \frac{n}{\log^2(n)} \right)^{\frac{-s}{2s+d-1}},\]
where $C_{d}>0$ is a constant that may depend on $d$ and on constants related to the Markov chain $(X_i)_{i \geq 1}$.
\paragraph{Conclusion.}We proved that on the event $\mathcal{E}$, it holds with probability at least $1-3/n^2$, \[\|G^*-G\|_F \leq D_1 \left(  \frac{n}{\log^2(n)} \right)^{\frac{-s}{2s+d-1}},\]
where $D_1>0$ is a constant that depends on $\Delta^*$, $d$ and on constants related to the Markov chain $(X_i)_{i \geq 1}$. Moreover, Eq.\eqref{G-Ghat} from the proof of Proposition \ref{prop:HEiC} gives that on the event $\mathcal{E}$, we have 
\[\|G-\hat{G}\|_F = \frac{d-2}{d}\|VV^{\top}-\hat{V}\hat{V}^{\top}\|_F \leq \frac{2^{\frac{9}{2}}\sqrt{d}\|T_n-\widehat{T}_n\|}{3 \Delta^*} .\]
Using the concentration result from~\cite{bandeira2016} on spectral norm of centered random matrix with independent entries we get that there exists some constant $D_2>0$ such that with probability at least $1-1/n^2$ it holds
\[\|G-\hat{G}\|_F  \leq D_2 \frac{\sqrt{\log n}}{n} .\]

Using again Proposition \ref{prop:HEiC}, we know that for $n$ large enough, $\mathds{P}(\mathcal{E}) \geq 1-1/n^2$. We conclude that for $n$ large enough, it holds with probability at least $1-5/n^2$, \[\|G^*-\hat{G}\|_F \leq D_3 \left(  \frac{n}{\log^2(n)} \right)^{\frac{-s}{2s+d-1}},\]
for some constant $D_3>0$ that depends on $\Delta^*$, $d$ and on constants related to the Markov chain $(X_i)_{i \geq 1}$ (see Theorem~\ref{delta2-proba-ope} for details).

\subsection{Proof of Proposition \ref{prop:HEiC}}

\label{apdx:eventE}
\paragraph{First part of the proof}

Let us consider $\gamma >0.$

Using the concentration of spectral norm for random matrices with independent entries from~\cite{bandeira2016}, there exists a universal constant $C_0$ such that
\[\mathds{P}\left(\|T_n-\widehat{T}_n\| \leq \frac{3\sqrt{2D_0}}{n}+C_0 \frac{\sqrt{\log n/\gamma}}{n}\right) \leq \gamma,\]
where denoting $Y = T_n-\widehat{T}_n$, we define $D_0:=\max_{1 \leq i\leq n} \sum_{j=1}^n Y_{i,j} \left(1-Y_{i,j}\right).$
We deduce that for $n$ large enough, it holds with probability at least $1-\gamma/4$,
\begin{equation}\|T_n-\widehat{T}_n\| \leq \frac{(\Delta^*)^2}{2^{\frac{13}{2}}\sqrt{d}} \label{bandeira-event}.\end{equation}
 Using now Theorem \ref{thm:delta2-proba-ope}, it holds with probability at least $1-\gamma/4$ for $n$ large enough
 \begin{equation}
 \delta_2\left( \lambda(T_n), \lambda(\mathds{T}_W)\right) \leq C \left( \frac{\log^2 n}{n} \right)^{\frac{s}{2s+d-1}} \leq \frac{\Delta^*}{8}.
 \label{delta2-event}
 \end{equation}
 Putting together Eq.\eqref{bandeira-event} and Eq.\eqref{delta2-event}, we deduce that for $n$ large enough, \[\mathds{P}\left( \mathcal{E} \right) \geq 1-\gamma/2.\]
 
 \paragraph{Second part of the proof}
In the following, we work on the event $\mathcal{E}$.
 Since $\Delta^*>0$ by assumption, we get that $p^*_1=\lambda_1^*= \dots = \lambda^*_d$ is the only eigenvalue of $\mathds{T}_W$ with multiplicity $d$. Indeed, all eigenvalue $p^*_k$ with $k >d$ has multiplicity $d_k>d$ and $p^*_0$ has multiplicity 1. Moreover, from Eq.\eqref{delta2-event}, we have that there exists a unique set of $d$ eigenvalues of $T_n$, denoted   $\lambda_{i_1},\lambda_{i_2},\dots,\lambda_{i_d}$, such that they are at a distance least $3\Delta^*/4$ away from the other eigenvalues, i.e.
\begin{equation}\Delta:=\min_{\nu_1 \in \lambda(T_n) \backslash \{\lambda_{i_1},\lambda_{i_2},\dots,\lambda_{i_d} \} } \max_{ \nu_2 \in \{ \lambda_{i_1},\lambda_{i_2},\dots,\lambda_{i_d} \}} |\nu_1-\nu_2|\geq \frac{3 \Delta^*}{4} . \label{Delta-minoration}\end{equation}
Let us form the matrix $V \in \mathds{R}^{n \times d}$ where the $k$-th column is the eigenvector of $T_n$ associated with the eigenvalue $\lambda_{i_k}$. We denote further $G:= VV^{\top}/d$.
Let  $\hat{V} \in \mathds{R}^{n\times d}$ be the matrix with columns corresponding to the eigenvectors associated to eigenvalues $\hat{\lambda}_{i_1},\hat{\lambda}_{i_2},\dots,\hat{\lambda}_{i_d}$ of $\widehat{T}_n$ and $\hat{G}:=\hat{V}\hat{V}^{\top}/d$. Using Theorem \ref{davis-kahan} there exists some orthonormal matrix $O \in \mathds{R}^{d \times d}$ such that \[\|VO-\hat{V}\|_F \leq \frac{2^{\frac{3}{2}}\min\{\sqrt{d}\|T_n-\widehat{T}_n\|,\|T_n-\widehat{T}_n\|_F\}}{\Delta}.\]
Denoting $\lambda_{i_1}^{sort}\geq \lambda_{i_2}^{sort}\geq \dots\geq \lambda_{i_d}^{sort}$ (resp. $\hat{\lambda}_{i_1}^{sort}\geq \hat{\lambda}_{i_2}^{sort}\geq \dots\geq \hat{\lambda}_{i_d}^{sort}$) the sorted version of the eigenvalues $\lambda_{i_1},\lambda_{i_2},\dots,\lambda_{i_d}$ (resp. $\hat{\lambda}_{i_1},\hat{\lambda}_{i_2},\dots,\hat{\lambda}_{i_d}$), we have
\begin{align}
&\left[\sum_{k=1}^d \left( \lambda_{i_k}^{sort}- \hat{\lambda}_{i_k}^{sort} \right)^2\right]^{1/2}
\nonumber\\ \leq \quad & \|VV^{\top}-\hat{V}\hat{V}^{\top}\|_F \quad \text{ (Hoffman-Wielandt inequality~\citep[Thm VI.4.1]{bathia})}\nonumber\\
\leq \quad & 2\|VO-\hat{V}\|_F \quad \quad \text{ (using Lemma \ref{lemma:frobe2frobe-ope})}\nonumber\\
\leq \quad & \frac{2^{\frac{5}{2}}\min\{\sqrt{d}\|T_n-\widehat{T}_n\|,\|T_n-\widehat{T}_n\|_F\}}{\Delta}\nonumber\\
\leq \quad & \frac{2^{\frac{9}{2}}\min\{\sqrt{d}\|T_n-\widehat{T}_n\|,\|T_n-\widehat{T}_n\|_F\}}{3\Delta^*} \quad \text{ (using Eq.\eqref{Delta-minoration})} \label{G-Ghat}\\
\leq \quad & \Delta^*/8. \quad \text{ (using Eq.\eqref{bandeira-event})}\nonumber
\end{align}

 Using the triangle inequality, we get that \begin{equation}\hat{\Delta}:=\min_{\nu_1 \in \lambda(\widehat{T}_n) \backslash \{\hat{\lambda}_{i_1},\hat{\lambda}_{i_2},\dots,\hat{\lambda}_{i_d} \} } \max_{ \nu_2 \in \{ \hat{\lambda}_{i_1},\hat{\lambda}_{i_2},\dots,\hat{\lambda}_{i_d} \}} |\nu_1-\nu_2|\geq \frac{\Delta^*}{2}.\label{Deltahat}\end{equation}
We proved that on the event $\mathcal{E}$, the eigenvalues in $\Lambda_1:= \{ \hat{\lambda}_{i_1}, \dots, \hat{\lambda}_{i_d} \}$ are at distance at least $\Delta^*/2$ from the other eigenvalues of $\widehat{T}_n$ (cf. Eq.\eqref{Deltahat}) and are at distance at most $\Delta^*/8$ of the eigenvalues $ \lambda_{i_1}, \dots, \lambda_{i_d} $ of $T_n$. We could have done this analysis for different eigenvalues. Let us consider some $k \geq 0$. Eq.\eqref{delta2-event} shows that on the event $\mathcal{E}$, there exists a set of $d_k$ eigenvalues of $T_n$ which concentrate around $p^*_k$ and such that it has diameter at most $\Delta^*/4$. Weyl's inequality (cf.~\citep[p.63]{bathia}) proves that there exist $d_k$ eigenvalues of $\widehat{T}_n$ that are at distance at most $\Delta^*/4$ from $p^*_k$. If we consider now a subset $L\neq \Lambda_1$ of $d$ eigenvalues of $\widehat{T}_n$, then the previous analysis shows that there exists some eigenvalue $\hat{\lambda}$ of $\widehat{T}_n$ which is not in $L$ and that is at distance at most $\Delta^*/4$ from one eigenvalue in $L$. Using Eq.\eqref{Delta-minoration}, we deduce that Algorithm (HEiC) returns $\hat{G}=\hat{V}\hat{V}^{\top}/d$ where the columns of $\hat{V}$ correspond to the eigenvectors of $\widehat{T}_n$ associated to the eigenvalues in $\Lambda_1$.

\subsection{Useful results}
\begin{lemma}\label{lemma:frobe2frobe-ope}
Let $A,B$ be two matrices in $\mathds{R}^{n\times d}$ then
\[\|AA^{\top}-BB^{\top}\|_F \leq (\|A\|+\|B\|)\|A-B\|_F.\]
If $A^{\top}A=B^{\top}B=\mathrm{Id}$ then \[\|AA^{\top}-BB^{\top}\|_F\leq 2\|A-B\|_F.\]
\end{lemma}

\begin{proof}[Proof of Lemma \ref{lemma:frobe2frobe-ope}.]
\begin{align*}
\|AA^{\top}-BB^{\top}\|_F &= \|(A-B)A^{\top}+B(A^{\top}-B^{\top})\|_F \\
&\leq \|A(A-B)^{\top}\|_F+\|(B-A)B^{\top}\|_F \\
& \leq  \|(A \otimes \mathrm{Id}_n)vec(A-B)\|_2+\|(\mathrm{Id}_d \otimes B)vec(A-B)^{\top}\|_2\\
&\leq \left( \|A \otimes \mathrm{Id}_n\|+\|\mathrm{Id}_d \otimes B\|\right) \|A-B\|_F\\
&= (\|A\|+\|B\|)\|A-B\|_F,
\end{align*}
where $vec(\cdot)$ represents the vectorization of a matrix that is its transformation into a column vector and $\otimes$ is the notation for the Kronecker product between two matrices.
\end{proof}

\begin{theorem} (Davis-Kahan Theorem, cf.~\citep{daviskahan}) \label{davis-kahan}
Let $\Sigma$ and $\hat{\Sigma}$ be two  symmetric $\mathds{R}^{n\times n}$ matrices with  eigenvalues $\lambda_1 \geq \lambda_2\geq \dots \geq \lambda_n$ and $\hat{\lambda}_1 \geq \hat{\lambda}_2\geq \dots \geq \hat{\lambda}_n$ respectively.  For $1\leq r \leq s \leq n$ fixed,  we  assume  that $\min\{\lambda_{r-1}-\lambda_r,\lambda_s-\lambda_{s+1}\}>0$ where $\lambda_0:=\infty$ and $\lambda_{n+1}=-\infty$.  Let $d=s-r+ 1$ and $V$ and $\hat{V}$ two matrices in $\mathds{R}^{n\times d}$ with columns $(v_r,v_{r+1},\dots,v_s)$ and $(\hat{v}_r,\hat{v}_{r+1},\dots,\hat{v}_s)$ respectively, such that $\Sigma v_j=\lambda_jv_j$ and $\hat{\Sigma} \hat{v}_j = \lambda_j \hat{v}_j$. Then there exists an orthogonal matrix $\hat{O}$ in $\mathds{R}^{d\times d}$ such that \[\|\hat{V}\hat{O}-V\|_F\leq \frac{2^{3/2} \min\{\sqrt{d} \|\Sigma-\hat{\Sigma}\|,\|\Sigma-\hat{\Sigma}\|_F\}}{\min\{\lambda_{r-1}-\lambda_r,\lambda_s-\lambda_{s+1}\}}.\]
\end{theorem}

\begin{lemma}\label{lemma:cycle-frobe}
Let $B$ be a $n \times d$ matrix with full column rank. Then we have
\[ \| BB^{\top} - B(B^{\top}B)^{-1}B^{\top}\|_F = \| \mathrm{Id}_d - B^{\top}B\|_F.\]
\end{lemma}

\begin{proof}[Proof of Lemma \ref{lemma:cycle-frobe}.]
Using the cyclic property of the trace, we have
\begin{align*}
&\| BB^{\top} - B(B^{\top}B)^{-1}B^{\top}\|_F^2\\
&\; \; =\| B\left(\mathrm{Id}_d - (B^{\top}B)^{-1}\right)B^{\top}\|_F^2 \\
&\; \;  = \mathrm{Tr}\left(B\left(\mathrm{Id}_d - (B^{\top}B)^{-1}\right)B^{\top}B\left(\mathrm{Id}_d - (B^{\top}B)^{-1}\right)B^{\top} \right)\\
&\; \;  = \mathrm{Tr}\left(B^{\top}B\left(\mathrm{Id}_d- (B^{\top}B)^{-1}\right)B^{\top}B\left(\mathrm{Id}_d - (B^{\top}B)^{-1}\right) \right)\\
&\; \;  = \mathrm{Tr}\left( \left(B^{\top}B- \mathrm{Id}_d\right)\left(B^{\top}B-\mathrm{Id}_d\right) \right)\\
&\; \; =\| \mathrm{Id}_d - B^{\top}B\|_F^2.
\end{align*}
\end{proof}

\begin{theorem} (cf.~\citep[ThmVII.3.4]{bathia}) \label{sintheta-stats}
Let $A$ and $B$ be two normal operators and $S_1$ and $S_2$ two sets separated by a strip  of  size $\delta$.  Let $E$ be  the  orthogonal  projection  matrix  of  the  eigenspaces  of $A$ with eigenvalues inside $S_1$ and $F$ be the orthogonal projection matrix of the eigenspaces of $B$ with eigenvalues inside $S_2$. Then
\[\|EF\|_F \leq \frac{1}{\delta}\|E(A-B)F \|_F \leq \frac{1}{\delta}\|A- B\|_F .\]
\end{theorem}

\begin{lemma} (Ostrowski's inequality)\label{lemma:ostrowski}
Let $A\in \mathds{R}^{n\times n} $ be a Hermitian matrix and $S\in \mathds{R}^{d\times n}$ be a general matrix then
\[ \|SAS^{\top}-A\|_F \leq \|A\|_F \times \|S^{\top}S-\mathrm{Id}_n\|.\]
\end{lemma}

\section{Proof of Proposition~\ref{prop:bayes-optimal}}

Notice that for any $i \in [n]$,
\begin{align*}
\mathds P\left( g_i(\mathbf D_{1:n}) \neq A_{i,n+1} \right) = \mathds E\left[ \mathds 1_{g_i(\mathbf D_{1:n}) \neq A_{i,n+1} }\right]= \mathds E \mathds E \left[ \mathds 1_{g_i(\mathbf D_{1:n}) \neq A_{i,n+1} } \; | \; \mathbf D_{1:n}\right],
\end{align*}
and that
\begin{align*}
&\mathds E \left[ \mathds 1_{g_i(\mathbf D_{1:n}) \neq A_{i,n+1} } \; | \; \mathbf D_{1:n}\right] \\
&= \mathds E\left[ \mathds 1_{g_i(\mathbf D_{1:n})=1} \mathds 1_{A_{i,n+1}=0} \; | \;\mathbf D_{1:n}\right] + \mathds E\left[ \mathds 1_{g_i(\mathbf D_{1:n})=0} \mathds 1_{A_{i,n+1}=1} \; | \;\mathbf D_{1:n}\right] \\
&= \eta_i(\mathbf D_{1:n}) \mathds 1_{g_i(\mathbf D_{1:n})=0} +  \left(1-\eta_i(\mathbf D_{1:n})\right)\mathds 1_{g_i(\mathbf D_{1:n})=1},
\end{align*}
which leads to 
\begin{equation*}  
\mathds P\left( g_i(\mathbf D_{1:n}) \neq A_{i,n+1} \right) = \mathds E\left[   \eta_i(\mathbf D_{1:n}) \mathds 1_{ g_i(\mathbf D_{1:n})=0} +  \left(1-\eta_i(\mathbf D_{1:n})\right)\mathds 1_{ g_i(\mathbf D_{1:n})=1} \right].
\end{equation*}

By definition of the Bayes classifier $g^*$, we have for any $i \in [n]$, 
\begin{align*}
&\mathds P\left( g_i^*(\mathbf D_{1:n}) \neq A_{i,n+1} \right)\\
&= \mathds E\left[   \eta_i(\mathbf D_{1:n}) \mathds 1_{ \eta_i(\mathbf D_{1:n})<\frac12} +  \left(1-\eta_i(\mathbf D_{1:n})\right)\mathds 1_{ \eta_i(\mathbf D_{1:n})\geq \frac12} \right]\\
&= \mathds E\left[   \min \left\{\eta_i(\mathbf D_{1:n}),1-\eta_i(\mathbf D_{1:n}) \right\} \left(  \mathds 1_{ \eta_i(\mathbf D_{1:n})\geq\frac12} +\mathds 1_{ \eta_i(\mathbf D_{1:n})< \frac12} \right) \right]\\
&= \mathds E\left[   \min \left\{\eta_i(\mathbf D_{1:n}),1-\eta_i(\mathbf D_{1:n}) \right\}  \right]\\
\end{align*}

Given another classifier $g$, we have for any $i \in [n]$,
\begin{align*}
&\mathds P\left( g_i(\mathbf D_{1:n}) \neq A_{i,n+1}  \right)
- \mathds P\left( g_i^*(\mathbf D_{1:n}) \neq A_{i,n+1} \right)\\ =\quad & \mathds E\Bigg[ \eta_i(\mathbf D_{1:n}) \mathds 1_{ g_i(\mathbf D_{1:n})=0} +  \left(1-\eta_i(\mathbf D_{1:n})\right)\mathds 1_{ g_i(\mathbf D_{1:n})=1} \\
& \qquad -  \left(\eta_i(\mathbf D_{1:n}) \mathds 1_{ g_i^*(\mathbf D_{1:n})=0} +  \left(1-\eta_i(\mathbf D_{1:n})\right)\mathds 1_{ g_i^*(\mathbf D_{1:n})=1} \right)\Bigg] \\
= \quad & \mathds E\big[ \eta_i(\mathbf D_{1:n}) \left( \mathds 1_{ g_i(\mathbf D_{1:n})=0}-\mathds 1_{ g_i^*(\mathbf D_{1:n})=0}\right)\\
\qquad \qquad & +  \left(1-\eta_i(\mathbf D_{1:n})\right) \left( \mathds 1_{ g_i(\mathbf D_{1:n})=1}-\mathds 1_{ g_i^*(\mathbf D_{1:n})=1}\right) \big]\\
= \quad & \mathds E\left[ \left(2 \eta_i(\mathbf D_{1:n})-1\right) \left(\mathds 1_{ g_i^*(\mathbf D_{1:n})=1}-\mathds 1_{ g_i(\mathbf D_{1:n})=1}\right)\right],
\end{align*}
where we used that $g(\mathbf D_{1:n})$ takes only the values $0$ and $1$, so that 
\[\mathds 1_{ g_i(\mathbf D_{1:n})=0}-\mathds 1_{ g_i^*(\mathbf D_{1:n})=0}  = \left(\mathds 1_{ g_i^*(\mathbf D_{1:n})=1}-\mathds 1_{ g_i(\mathbf D_{1:n})=1}\right).\]
Since
\begin{align*}
\mathds 1_{ g_i^*(\mathbf D_{1:n})=1}-\mathds 1_{ g_i(\mathbf D_{1:n})=1} &= \left\{
    \begin{array}{ll}
        1 & \mbox{if } g_i^*(\mathbf D_{1:n})=1 \text{ and } g_i(\mathbf D_{1:n})=0 \\
        0 & \mbox{if } g_i^*(\mathbf D_{1:n})=g_i(\mathbf D_{1:n}) \\
        -1 & \mbox{if } g_i^*(\mathbf D_{1:n})=0 \text{ and } g_i(\mathbf D_{1:n})=1
    \end{array} 
\right.\\
&= \mathds 1_{g_i^*(\mathbf D_{1:n}) \neq g_i(\mathbf D_{1:n})}\; \mathrm{sgn}(\eta_i(\mathbf D_{1:n})-1/2),
\end{align*}
we deduce that \begin{align*}&\mathds P\left( g_i(\mathbf D_{1:n}) \neq A_{i,n+1} \right)
- \mathds P\left( g_i^*(\mathbf D_{1:n}) \neq A_{i,n+1} \right)\\
&= 2\mathds E\left[ \left| \eta_i(\mathbf D_{1:n})-\frac12\right| \times \mathds{1}_{g_i(\mathbf D_{1:n}) \neq g_i^*(\mathbf D_{1:n})}\right],\end{align*}
which concludes the proof.

\end{document}